# Understanding Critical Thinking in Generative Artificial Intelligence Use: Development, Validation, and Correlates of the Critical Thinking in AI Use Scale

Gabriel R. Lau[1*], Wei Yan Low[2], Louis Tay[3], Ysabel A. Guevarra[4], Dragan Gašević[5], Andree Hartanto[4*]

[1] School of Social Sciences, Nanyang Technological University, Singapore

[2] Interdisciplinary Graduate Programme, Nanyang Technological University, Singapore

[3] Department of Psychological Sciences, Purdue University, Indiana, United States

[4] School of Social Sciences, Singapore Management University, Singapore

[5] Faculty of Information Technology, Monash University, Victoria, Australia

## Author Note

[*] Correspondence concerning this article should be addressed to Gabriel R. Lau (gabriel.laury@ntu.edu.sg) and Andree Hartanto (andreeh@smu.edu.sg), Singapore Management University, School of Social Sciences, 10 Canning Rise, Level 4, Singapore 179873. Phone: (65) 6828 1901.

**Abstract**

Generative AI tools are increasingly embedded in everyday work and learning, yet their fluency, opacity, and propensity to hallucinate mean that users must critically evaluate AI outputs rather than accept them at face value. The present research conceptualises critical thinking in AI use as a dispositional tendency to verify the source and content of AI-generated information, to understand how models work and where they fail, and to reflect on the broader implications of relying on AI. Across six studies (*N* = 1341), we developed and validated the 13-item critical thinking in AI use scale and mapped its nomological network. Study 1 generated and content-validated scale items. Study 2 supported a three-factor structure (Verification, Motivation, and Reflection). Studies 3 and 4 confirmed the higher-order model, demonstrated strong factor loadings, internal consistency, sex invariance, convergent and discriminant evidence for validity, and showed that critical thinking in AI use was positively associated with openness, extraversion, positive trait affect, and frequency of AI use. Study 5 supported the scale's test-retest reliability. Lastly, Study 6 demonstrated criterion evidence of validity for the scale, with higher critical thinking in AI use scores predicting more frequent and diverse verification strategies, greater veracity judgement accuracy in a novel and naturalistic GPT-powered AI chatbot fact-checking task, and deeper reflection about responsible AI. The current work clarifies why and how people exercise oversight over generative AI outputs and provides a validated scale and ecologically grounded paradigm to support theory testing, cross-group, and longitudinal research on critical thinking in AI use.

## 1. Introduction

Generative artificial intelligence (AI), such as large language models (LLMs), has reshaped how people access, process, and use information, and it is increasingly embedded in everyday tasks such as information search and decision-making. By analysing complex problems and generating customisable content, generative AI provides round-the-clock, highly versatile support across diverse applications (C. K. Y. Chan & Hu, 2023; Doshi et al., 2025; Montazeri et al., 2024). Through reducing cognitive load in memory storage or information processing, generative AI tools enable users to produce higher-quality outputs in less time than was typically feasible without AI support (Hong et al., 2025). Empirical evidence links generative AI use to gains in productivity and work quality (Noy & Zhang, 2023), task performance (Hemmer et al., 2023), and, in some contexts, creativity (Doshi & Hauser, 2024; B. C. Lee & Chung, 2024). Users also report greater perceived usefulness, enjoyment, and satisfaction with generative AI than with conventional web search for information-seeking tasks (Karunaratne & Adesina, 2023; Xu et al., 2023). These affordances and benefits have driven rapid adoption across educational and workplace settings (Bouteraa et al., 2024; Brynjolfsson et al., 2025; Gruenhagen et al., 2024; Humlum & Vestergaard, 2025; D. Lee et al., 2024).

While the benefits of using generative AI are well-documented, there have been growing concerns about overreliance on AI tools and their potential threats to human cognitive capabilities (Gerlich, 2025a). Overreliance refers to the tendency to depend on AI outputs beyond what is warranted by their accuracy, in which users defer to AI too readily, accept outputs at face value, or reduce independent judgement and verification (Klingbeil et al., 2024; J. D. Lee & See, 2004). Two related mechanisms may explain why such overreliance develops. First, generative AI can produce highly fluent, seemingly reliable content that may function as a heuristic cue, fostering trust and reducing the felt need for further scrutiny (Klingbeil et al., 2024; Kosmyna et al., 2025; Stadler et al., 2024). Second, AI tools free working-memory resources by offloading storage and computation (Risko & Gilbert, 2016), which may further diminish active cognitive engagement and limit opportunities to practise analytic reasoning (Bastani et al., 2025; Fan et al., 2025; Singh et al., 2025). Consistent with these mechanisms, empirical evidence suggests that when users accept AI outputs with limited scrutiny, they engage in less deep analytical processing, show reduced motivation for independent analysis, and exhibit greater automation bias (Alexander et al., 2018; Gerlich, 2025a; Grinschgl & Neubauer, 2022; Gsenger & Strle, 2021; Kosmyna et al., 2025; Stadler et al., 2024). Beyond these cognitive consequences, algorithmic recommendations may also displace personal autonomy, constraining individual judgement and creative exploration (Topol, 2019). These concerns are particularly acute given that generative AI outputs can appear legitimate even when inaccurate, meaning that users who forgo verification risk accepting and acting on erroneous information (Fernandes et al., 2026; Huang et al., 2025).

Despite growing concerns, there is a limited understanding of how individuals differ in their tendency and ability to critically evaluate AI outputs. The present research anchors this idea of critical evaluation for generative AI use within the broader critical thinking literature. This construct of critical thinking in AI use remains empirically underexplored and conceptually underdefined, as existing critical thinking research has mainly relied on domain-general assessments developed for pre-AI contexts and has failed to capture the unique cognitive demands of generative AI use (e.g., opaque algorithms, fluent but inaccurate outputs; Gerlich, 2025b; C. Lee et al., 2025; Suriano et al., 2025). To address this gap, the current research aims to articulate a construct definition aligned with the broader critical-thinking literature, present a theory-driven account of underlying mechanisms to guide item generation and structure, and investigate the correlates of critical thinking in AI use through the development and validation of the critical thinking in AI use scale.

### 1.1 Critical Thinking in AI Use

Critical thinking has been hailed as one of the 21st-century skills in educational and workplace settings (Ananiadou & Claro, 2009; Thornhill-Miller et al., 2023), and in an era increasingly shaped by generative AI, critical thinking in the context of using such tools has become essential for contemporary education and work (Gerlich, 2025a). The present research proposes that critical thinking in AI use refers to reflective thinking about what to believe or do when interacting with AI systems and AI-generated outputs, adapted from Ennis' (1985, 2011) foundational definition of critical thinking. Aligned with the broader critical thinking literature (Ennis, 1985, 2011; Facione, 1990a; Sosu, 2013), it involves deliberately evaluating AI-generated content to make informed and responsible decisions.

#### 1.1.1 *Heuristic-Systematic Model*

The Heuristic-Systematic Model (HSM; Chaiken, 1980; S. Chen & Chaiken, 1999) serves as a possible explanatory framework for understanding critical thinking in AI use, with dual-process theory (Kahneman, 2011; Stanovich & West, 2008) providing the underlying cognitive architecture that distinguishes fast and intuitive heuristic processing from slower and analytic systematic processing. The HSM proposes that people either rely on heuristic processing, guided by simple cues such as fluency, apparent expertise, or consensus, or engage in systematic processing that involves effortful evaluation of the content when they have sufficient motivation and ability (Chaiken, 1980; S. Chen & Chaiken, 1999). Within this framework, features of AI systems and their outputs, such as fluent language, authoritative tone, and interface credibility, function as antecedent heuristic cues that activate heuristic processing and thereby promote uncritical acceptance of AI-generated content. When such heuristic processing is sustained over repeated interactions, it can lead users to favour AI outputs over contradictory information from non-automated sources (Alexander et al., 2018).

**1.1.2 *Distinctives and Overlaps with Traditional Critical Thinking***

The epistemic features of generative AI outputs differ fundamentally from traditional information sources in ways that heighten the need for a domain-specific construct. Unlike traditional information evaluation, which relies on examining explicit arguments, verifiable evidence, and transparent inferential steps (Facione, 1990b; Halpern, 2014), the reasoning processes of AI are largely opaque, probabilistic, and non-explanatory (Lebovitz et al., 2022; Liao & Vaughan, 2023). This opacity behind AI outputs prevents users from inspecting the premises, evidence, or inferential steps that are central to traditional critical thinking. At the same time, generative AI systems routinely produce fluent, logical-sounding outputs that may contain omissions, hallucinations, and overconfident claims (Huang et al., 2025; Lee et al., 2025; Savage et al., 2025). Beyond inaccuracies at the level of individual outputs, contemporary AI systems can perpetuate existing biases, amplify economic inequalities, and impose environmental costs (Hagerty & Rubinov, 2019; Ho et al., 2025), and can be further strategically weaponised for manipulation, disinformation, and other malicious ends (Ibrar et al., 2025). These unique features underscore the need for a domain-specific application of critical thinking that focuses on the evaluation of opaque and probabilistic AI-generated information rather than human arguments.

Critical thinking in AI use overlaps with traditional critical thinking by preserving its core aim of forming reasonable, well-grounded judgements and by drawing on reflective judgement and epistemic motivation (Ennis, 1985, 2011; Facione, 1990b; Sosu, 2013). However, because critical thinking is context- and domain-sensitive (Lilienfeld et al., 2020; McPeck, 1990), its application to AI must address the distinctive epistemic features of generative systems. Whereas traditional critical thinking typically focuses on analysing explicit arguments supported by accessible evidence and transparent reasoning, critical thinking in AI use involves evaluating opaque, probabilistic outputs that may appear coherent despite being inaccurate, biased, or misleading (Lebovitz et al., 2022; C. Lee et al., 2025; Liao & Vaughan, 2023; Savage et al., 2025). Accordingly, it entails not only assessing plausibility, but also verifying source and veracity and judging when AI outputs should be trusted, given their potential ethical and societal consequences. In other words, critical thinking in AI use also requires verification.

In summary, critical thinking in AI use comprises three dimensions: reflective judgement, verification, and epistemic motivation. In the following, we describe each in turn.

### 1.1.3 *Reflective Judgement*

Reflective judgement is defined as a form of epistemic cognition in which individuals recognise uncertainty in knowledge claims and justify conclusions by weighing evidence, alternatives, and context, particularly for complex problems with no clear single answer (King & Kitchener, 2004; Kitchener & King, 1981). In line with Sosu's (2013) notion of reflective scepticism, it involves a dispositional tendency to reassess one's own views, anticipate wider implications, and resist taking AI outputs at face value. Rather than merely asking whether a response is factually correct, users must consider the context in which it is produced, the potential consequences of accepting it, and whether their judgements should be revised. This conception also aligns with classic accounts that treat critical thinking as reflective judgement, whereby individuals evaluate the context and likely consequences of knowledge claims in deciding what to believe and what to do (Ennis, 1985, 2011; Facione, 1990a; King & Kitchener, 1994, 2004).

Given that contemporary AI systems can perpetuate existing biases, amplify economic inequalities, and impose environmental costs (Hagerty & Rubinov, 2019), critical thinking in AI use therefore requires users to reflect explicitly on the broader ethical and societal consequences of relying on AI-generated content, not only on the correctness of any single output. Such reflection on the broader risks and benefits of AI-generated content, beyond its surface accuracy, is especially vital in an era where powerful generative tools can be strategically weaponised for manipulation, disinformation, and other malicious ends (Ibrar et al., 2025). Within the HSM framework, making reflective judgements maps onto systematic processing, as it involves weighing reasons and counterarguments, anticipating downstream consequences, and scrutinising the broader ethical and societal implications of relying on AI-generated content.

#### 1.1.4 *Verification*

Second, complementing the reflective judgements, critical thinking in AI use also requires verification, because even fluent, logical-sounding outputs can contain omissions, hallucinations, and overconfident claims (Huang et al., 2025; Lee et al., 2025; Savage et al., 2025). These cues can easily be read as signals of validity and can reduce the felt need to question what the model says (Quelle & Bovet, 2024; Sun et al., 2025)[1]. Verification is defined as the specific set of evaluative behaviours through which users assess the accuracy, credibility, and source quality of AI-generated content, such as cross-referencing external sources, consulting domain experts, or checking cited references (Chu-Ke & Dong, 2024; C. Lee et al., 2025; Metzger et al., 2003). Verification involves pausing before endorsing an output, checking external sources, considering alternative interpretations and disconfirming evidence, and revisiting conclusions as new information becomes available.

Unlike reflective judgements, verification should not be understood as a purely systematic process. In practice, verification can be initiated and supported by automatic processes, such as increasingly routinised verification habits and the use of search engines and even AI tools themselves for cross-checking information. Training and frequent practice can gradually automate effortful and systematic information processing (Boissin et al., 2021; Raoelison et al., 2021). Using AI tools for verification purposes can further support this shift through cognitive offloading, defined as the delegation of cognitive tasks (e.g., memory storage, computation, information retrieval) to external tools, including AI systems, thereby reducing the demands on internal cognitive resources (Risko & Gilbert, 2016). By reducing the effort required for verification (Gerlich, 2025c), such tools can help ensure that belief aligns with the actual strength of the evidence underlying AI-generated content.

[1] Recent experimental evidence suggests that the cognitive consequences of generative AI use depend on how the interaction is structured, such that unguided AI use may foster cognitive offloading without improving reasoning quality, whereas structured prompting can reduce offloading and enhance critical reasoning and reflective engagement (Gerlich, 2025b). This suggests that AI-related cognitive outcomes are shaped not only by whether AI is used, but also by how AI is embedded within the task and decision process.

#### **1.1.5** ***Epistemic Motivation***

Lastly, understanding how the AI systems operate is foundational to critical thinking in AI because this understanding shapes one's propensity to engage in evaluative behaviours, such as scrutinising AI-generated outputs, cross-verifying sources, and revising their judgements. Critical thinking in AI use involves not only reflective judgements, and verification, but also being motivated to learn how generative models are trained, what information they draw on, and which tasks or domains are likely to exceed their capabilities.

In this context, epistemic motivation is defined as the willingness to invest cognitive effort in forming rich, accurate, and well-informed beliefs rather than relying on superficial or premature judgements (Amit & Sagiv, 2013; De Dreu et al., 2008). Because generative AI systems are opaque and rapidly evolving, understanding how they operate requires sustained cognitive investment as users must actively seek out information about training processes, model limitations, and common failure modes rather than passively acquiring it through use alone. Individuals differ in their willingness to make this investment. Those higher in epistemic motivation might be more inclined to move beyond surface-level interaction with AI, treating it as a system to be understood rather than a black-box shortcut (Pinski & Benlian, 2024; C. Zhang & Magerko, 2025). This drive to learn, in turn, equips users to better judge the plausibility and limits of AI outputs (Bećirović et al., 2025). In that sense, critical thinking in AI use includes an ongoing learning process, in which users progressively refine both their motivation and their ability to understand AI systems, so that their verification practices, reflective judgements, and reliance on AI become increasingly calibrated and responsible over time.

**1.2 Correlates of Critical Thinking in AI Use**

Beyond conceptual definition and theoretical underpinnings, the current work also aims to identify the possible correlates that are associated with critical thinking in AI use. Because AI-specific evidence on these correlates remains limited, the following hypotheses draw primarily on the broader critical-thinking literature and are extended here to the AI context. Critical thinking skills have been linked to factors such as individual personality differences, specifically, openness to experience consistently emerges as the only Big Five trait that reliably predicts critical thinking skills, with higher openness associated with better performance on standard critical thinking tests, while extraversion, agreeableness, conscientiousness, and neuroticism do not consistently show meaningful associations with critical thinking skills (Acevedo & Hess, 2022; Clifford et al., 2004; Ku & Ho, 2010). This pattern suggests that an individual's intellectual curiosity and willingness to reconsider alternative ideas are important characteristics of critical thinking. Given that critical thinking in AI use likewise requires users to seek out new information, interrogate AI outputs, and revise their beliefs in light of emerging evidence, a dispositional openness to experience is expected to be positively associated with critical thinking in AI use.

Besides personality, emotions also have a significant influence on cognition (Dolan, 2002), and this influence is evident in critical thinking performance as well (Lun et al., 2023). Recent work suggests that both positive and negative emotions are beneficial to critical thinking (Lun et al., 2023). Prior work has shown that positive mood enhances the tendency to use critical thinking in both subjective self-reports (Q. Zhang & Zhang, 2013), and objective essay exams (Lewine et al., 2015). Negative mood can reduce one's susceptibility to misleading information (Forgas, 2019) and boost performance on critical thinking assessment (Lun et al., 2023). Drawing on the assimilative–accommodative framework and feelings-as-information theory (Bless & Fiedler, 2006; Forgas, 2017; Schwarz, 2012), mood functions as a signal that tunes information-processing strategies. Negative mood signals potential problems in the environment, motivating vigilant, analytic scrutiny of evidence and reducing susceptibility to misleading claims (Forgas, 2019). From a dual-process theoretical perspective (Chaiken, 1980; S. Chen & Chaiken, 1999; Kahneman, 2011), negative mood tends to shift processing toward systematic mode, mapping directly onto the evaluative, effortful engagement that characterises critical thinking. Accordingly, negative mood is expected to be positively associated with critical thinking in AI use. The role of positive mood is less straightforward. Although positive mood is traditionally associated with heuristic processing (Schwarz, 2012), some evidence suggests that positive mood can also enhance critical thinking performance (Lewine et al., 2015; Lun et al., 2023; Q. Zhang & Zhang, 2013). One possibility is that this effect reflects arousal rather than positive valence per se, as the heightened activation accompanying certain positive states may sustain effortful engagement with the task (see Clore & Huntsinger, 2007). Given these mixed findings, it is unclear whether positive mood would also be positively associated with critical thinking in AI use.

Having outlined the construct definition, theoretical mechanisms, and expected correlates of critical thinking in AI use, we next consider its expected downstream consequences. Drawing on the HSM, critical thinking in AI use is expected to have important downstream consequences for how individuals navigate an AI-saturated information environment. When people habitually verify AI-generated content, override intuitive acceptance, and remain vigilant to deception, they should be able to better filter misinformation and resist cognitive biases that exploit the fluency and authority cues of generative AI systems (Frederick, 2005; C. Lee et al., 2025; Savage et al., 2025). One such bias is automation bias, the downstream behavioural consequence of sustained heuristic processing in AI-mediated contexts, whereby users systematically favour AI outputs over contradictory information from non-automated sources (Alexander et al., 2018). By fostering systematic evaluation of AI outputs, critical thinking in AI use should help counteract this tendency. This protective role may be especially important when AI systems are perceived as authority-like sources of direction, as obedience to AI has been linked to reduced critical thinking in AI use and greater compliance with unethical AI instructions (Lau, Tong, et al., 2026). Drawing on Facione's (1990a) view of critical thinking as purposeful, self-regulatory judgement, critical thinking in AI use should enhance the quality of everyday decisions by promoting systematic fact-checking, cross-referencing sources, and explaining why a given AI response is or is not trustworthy. Users who are able to recognise uncertainty, identify algorithmic biases, and detect subtle manipulation tactics are less likely to be nudged into unwarranted confidence, polarised beliefs, or discriminatory choices reflected on AI outputs (Almulla, 2023; Machete & Turpin, 2020; Puig et al., 2021). Over time, critical thinking in AI use should not only reduce individual susceptibility to misinformation but also support healthier information ecologies, as users are more selective about what they rely on and believe in. In this way, critical thinking in AI use functions as a protective cognitive resource which helps maintain human oversight over opaque, persuasive AI systems and fosters more accurate, accountable, and ethically informed decision-making in an increasingly complex, AI-mediated world.

**1.3 The Current Research**

The present research pursued two overarching aims. The first aim was descriptive, which was to define critical thinking in AI use as a construct, develop and validate a psychometrically sound measure of it, and identify its correlates within a broader nomological network. The second aim was evaluative, which was to examine whether this construct has meaningful behavioural implications by testing its criterion validity in AI-mediated tasks. To these ends, the present research developed and validated the critical thinking in AI use scale, an instrument to assess individuals' tendencies to verify AI sources and content, to reflect on the responsible use of AI-generated outputs, and to understand how generative models operate.

Across six studies, the present research established the theoretical coherence, psychometric robustness, and practical relevance of the scale. This staged, multi-study design follows established best practice for scale development and validation (Boateng et al., 2018; DeVellis & Thorpe, 2021), separating each phase of the validation process into independent samples to prevent capitalisation on chance, to progressively accumulate distinct forms of validity evidence (structural, convergent, discriminant, criterion-related), and to allow replication of key findings across independent datasets. Study 1 generated, refined, and content-validated an initial item pool, using expert review and naïve judges to ensure strong content adequacy and accessibility, yielding a 27-item scale with excellent substantive agreement and readability. Study 2 employed exploratory factor analysis (EFA) to identify a clear three-factor structure, and reduced the measure to 13 items with a simple and interpretable structure. Study 3 confirmed this structure via confirmatory factor analysis (CFA), demonstrated high internal consistencies, supported a higher-order critical thinking in AI use factor, and provided evidence of convergent and discriminant validity, alongside measurement invariance across sex. Study 4 conducted a second psychometric validation, replicating the factor structure and extending the nomological network. Study 5 examined both 1-week and 2-week test-retest reliability, providing evidence for the temporal stability of critical thinking in AI use. Finally, Study 6 evaluated criterion validity by linking critical thinking in AI use scores to performance and behaviour in AI-mediated tasks that represent the core processes of critical thinking in AI use.

## 2. Study 1: Item Generation and Reduction

### 2.1 Method

#### 2.1.1 *Item Generation & Refinement*

In Study 1, we aimed to create a measure to capture different aspects of critical thinking in AI use. In line with our conceptualisation of critical thinking in AI use as involving reflective judgement, epistemic motivation, and a propensity to verify AI-generated information, we operationalised three core dimensions: reflection, motivation, and verification. During item generation, we added two possible dimensions, scepticism and openness, to ensure comprehensive coverage of the critical thinking in AI use construct. In total, we generated an initial pool of ninety items adapted from three validated critical thinking measures, Critical Thinking Toolkit (Stupple et al., 2017), Critical Motivational Scale (Valenzuela et al., 2017), Critical Thinking Disposition Scale (Sosu, 2013). Each initial dimension was represented by eighteen items to ensure adequate conceptual breadth within that dimension. All items were refined to capture core characteristics of generative AI systems, including opaque reasoning processes, probabilistic and occasionally hallucinated outputs, and their susceptibility to data and societal biases, while retaining the underlying critical thinking constructs of the source scales. For example, the Critical Thinking Disposition Scale items "I usually check the credibility of the source of information before making judgements" and "I usually think about the wider implications of a decision before taking action" were adapted to "I tend to check the sources of AI-generated information before fully relying on it" and "I sometimes consider the broader consequences of using AI outputs when making decisions," respectively, to represent the verification and reflection dimensions. The full list of ninety items, the source measure for each dimension, and an example of how an original item was adapted to the AI context, is available in Supplementary Materials Table S1. Three Subject Matter Experts (SMEs), the senior authors, with expertise in scale development and generative AI research, then reviewed the items for dimensional relevance, face validity, and redundancy, selecting thirty items, approximately six per dimension, that best captured the construct.

#### 2.1.2 *Analytic Plan*

We adopted the content validation process of Anderson and Gerbing (1991) and applied the evaluation criteria of Colquitt et al. (2019) to refine the scale. Consistent with Anderson and Gerbing's recommendation to use naïve judges, we recruited a total of 24 naïve judges to independently evaluate all items and mapped them onto the five dimensions across four rounds of review administered via Qualtrics. Demographic information was not formally collected; all evaluators were recruited from local universities in Singapore and were university students (comprising both undergraduates and graduates) or recent graduates. Content adequacy was assessed using two indices: $p_{sa}$ (proportion of substantive agreement), defined as the proportion of judges who correctly mapped an item to its intended construct, and $c_{sv}$ (substantive validity coefficient), calculated as the difference between the number of correct mappings and the highest number of incorrect mappings to any other construct, divided by the total number of judges (Colquitt et al., 2019).

We evaluated the readability of the items using the Flesch–Kincaid tests, which estimate reading difficulty based on average syllables per word and words per sentence (Ravens-Sieberer et al., 2014). Prior research revealed that the average U.S. adult reads at approximately a 7th-grade level, corresponding to a Flesch Reading Ease score of 60–70 (Stockmeyer, 2008), where lower grade levels and higher ease scores indicate greater readability.

## 2.2 Results and Discussion

Across four rounds of content validation, items were iteratively revised for readability, clarity, and interpretability, before deletion. Of the initial thirty items, sixteen items, seven items, and five items failed to meet Colquitt et al.'s (2019) threshold criteria of .72 for $p_{sa}$ and .51 for $c_{sv}$ in rounds 1 to 3, respectively. After item refinement and the addition of dimensional descriptions in the survey, only three items failed in the fourth round, which were completed by five RAs, and these items were removed, yielding the 27-item critical thinking in AI use scale. The final scale demonstrated excellent content validity ($p_{sa}$ = 0.96; $c_{sv}$ = 0.93), indicating that nearly all items were correctly mapped to their intended dimensions by the majority of judges.

In terms of scale readability, the 27-item critical thinking in AI use scale achieved a Flesch–Kincaid Grade Level of 6.22 and a Reading Ease score of 74, suggesting that the items were easier to read than the average adult reading level. These results indicate that the scale is not only comprehensive in its coverage of the key dimensions of critical thinking in AI use, but also accessible and easy to understand for the broader audience. The 27-item version of the scale is available in Supplementary Materials Table S2.

# 3. Study 2: Exploratory Factor Analysis

Following the findings from Study 1, Study 2 was conducted to examine the factor structure and refine the critical thinking in AI use scale using exploratory factor analysis (EFA).

## 3.1 Method

### 3.1.1 *Participants and Procedure*

Data collection for all studies in the current work was approved by the corresponding authors' university institutional review board (IRB-25-141-A112-M3(925)) prior to data collection. To meet the recommended guideline of at least 10 respondents per item (Boateng et al., 2018), we recruited 270 adults from Prolific to complete the 27-item critical thinking in AI use scale. Participants were at least 18 years old, U.S. residents, and ranged in age from 21 to 83 years ($M_{age}$ = 44.25, $SD_{age}$ = 13.30); 52.96% were female. Informed consent was obtained electronically prior to participation. The study was administered via Qualtrics, which included the scale and demographic questions, and it took approximately 3 min to complete. Attention to instructions was verified with an attention check item ("Please select 'Disagree' for this statement"), which all participants passed. Most participants identified as American by nationality (95.9%), Caucasian/White (68.9%), and held at least a bachelor's degree (Bachelor's = 35.2%, or higher = 14.0%). Household income was in the mid-range, with 58.2% earning $25,000–$100,000. The full demographic data are available in Supplementary Materials Table S3.

**3.1.2 *Measure***

The 27-item critical thinking in AI use scale was developed in Study 1 to capture five possible theoretical dimensions of critical thinking: verification, scepticism, openness, motivation, and reflection. Each item was rated on a 5-point Likert scale ranging from 1 (Strongly disagree) to 5 (Strongly agree), with higher scores reflecting greater endorsement of critical thinking in the context of using generative AI.

**3.1.3 *Analytic Plan***

All data analyses were conducted in R version 4.5.2 (R Core Team, 2025). All data cleaning and manipulation were performed using *dplyr* version 1.1.4 (Wickham et al., 2023) and *tidyr* version 1.3.1 (Wickham et al., 2024). All descriptives and internal consistency reliability were calculated using *psych* version 2.5.6 (Revelle, 2025). Descriptive statistics indicated that the items demonstrated adequate variability for factor analysis. Item means ranged from 2.40 to 4.35 (*SD*s = 0.69–1.39), with no evidence of floor or ceiling effects. Most items were negatively skewed, reflecting stronger endorsement, and kurtosis values ranged from –1.31 to 3.24, suggesting that item distributions were acceptable but varied in shape. Prior to conducting factor analyses, we evaluated the suitability of the data using Bartlett's Test of Sphericity and the Kaiser–Meyer–Olkin (KMO) Measure of Sampling Adequacy. EFA, including the KMO test and Bartlett's test of sphericity, was also conducted using *psych* version 2.5.6 (Revelle, 2025).

Although the scale was designed a priori based on a conceptual framework, we conducted an EFA to empirically examine its underlying latent structure. This analytic approach enabled us to evaluate how items clustered in practice without imposing a fixed structure and to identify potential cross-loadings, weak items, or unexpected factor patterns before proceeding to confirmatory testing in Study 3. To determine the appropriate number of factors, we conducted a parallel analysis using Pearson correlation matrices with principal axis factoring as the extraction method (Horn, 1965). Parallel analysis compares eigenvalues from the observed data with those generated from random data of the same size and dimensionality, retaining only those factors with eigenvalues greater than expected by chance (Hayton et al., 2004). We supplemented this procedure with visual inspection of the scree plot, in which factors are retained up to the point of inflection (Shrestha, 2021), as well as consideration of the Bayesian Information Criterion (BIC; Preacher et al., 2013), with lower values indicating superior fit, and the theoretical interpretability of the resulting factor solutions.

EFA was performed using principal axis factoring with oblimin rotation, which permits correlated factors. We adopted an iterative item-trimming procedure. At each iteration, items were evaluated against predefined thresholds, with communalities $< .35$, weak primary loadings $< .40$, or substantial cross-loadings $\geq .35$, particularly when the difference between the primary and secondary loading was < .20, were flagged for review as problematic (Hair et al., 2019). Communalities were obtained from the factor solution output; primary and cross-loadings were evaluated using the pattern matrix. Flagged items were removed iteratively while ensuring no factor was reduced below three items (Costello & Osborne, 2005). After each item removal, the model was re-estimated and fit indices (RMSEA with 90% CI, TLI, RMSR, BIC, and total variance explained) were recalculated, with iterations continuing until all items met criteria or further deletions violated the minimum item rule. We compiled the iteration history and selected the solution with the lowest RMSEA as the optimal model, after which items lacking conceptual alignment with the intended construct were removed (Hair et al., 2019).

**3.2 Results and Discussion**

Bartlett's test was statistically significant, $\chi^2(351) = 3714.58$, $p < .001$, indicating that the data were appropriate for factor analysis. The overall KMO value was excellent (KMO = .92), with all individual item KMOs exceeding .82. Parallel analysis showed that the first four observed eigenvalues exceeded simulated values, with subsequent eigenvalues falling below the random data threshold. The scree plot likewise displayed a clear inflection after the fourth factor (refer to Supplementary Materials Figure S1 for the parallel analysis scree plot). Although a four-factor model was initially estimated, closer inspection revealed that the fourth factor accounted for relatively little variance (4%) and only contained two items. A factor with fewer than three items is generally weak and unstable (Costello & Osborne, 2005), and the two items in question (items 15 and 16) showed weak primary loadings of .37 and .36, respectively. Furthermore, the Bayesian Information Criterion favoured the three-factor model (BIC = –973) over the four-factor model (BIC = –958). Taken together, these results supported a more parsimonious three-factor solution, which demonstrated stronger interpretability and adequate fit (RMSEA = .06, TLI = .89, CFI = .92, RMSR = .04). A three-factor structure was therefore retained for subsequent analyses.

**Table 1**

*13-Item Critical Thinking in AI Use Scale*

Preamble:
Think about how you usually use AI tools (e.g., chatbots, image generators, voice assistants).
For each statement, indicate how true it is of your typical behaviour using 1 (strongly disagree) to 5 (strongly agree).

| | 1 = Strongly disagree | 2 = Disagree | 3 = Neither agree nor disagree | 4 = Agree | 5 = Strongly agree |
|---|---|---|---|---|---|
| *Verification of source and content* | | | | | |
| 1. I often look at the sources of AI content before I rely on it | ☐ | ☐ | ☐ | ☐ | ☐ |
| 2. I sometimes check other sources or expert opinions to help judge what AI says | ☐ | ☐ | ☐ | ☐ | ☐ |
| 3. I try to check whether AI content is reliable | ☐ | ☐ | ☐ | ☐ | ☐ |
| 4. I think it is important to check the accuracy of AI information | ☐ | ☐ | ☐ | ☐ | ☐ |
| 5. I often check AI content to make sure it is correct and clear | ☐ | ☐ | ☐ | ☐ | ☐ |
| *Motivation to understand AI* | | | | | |
| 6. I try to understand how AI creates its answers whenever I can | ☐ | ☐ | ☐ | ☐ | ☐ |
| 7. Sometimes I feel motivated to learn how AI works so I can judge its answers better | ☐ | ☐ | ☐ | ☐ | ☐ |
| 8. I try to understand why the AI makes certain recommendations | ☐ | ☐ | ☐ | ☐ | ☐ |
| 9. I try to understand why AI gives different answers | ☐ | ☐ | ☐ | ☐ | ☐ |
| *Reflection on responsible AI* | | | | | |
| 10. I sometimes think about how using AI might affect the environment | ☐ | ☐ | ☐ | ☐ | ☐ |
| 11. I often think about the ethical problems that AI content might cause | ☐ | ☐ | ☐ | ☐ | ☐ |
| 12. I sometimes think about how the growing use of AI might change society | ☐ | ☐ | ☐ | ☐ | ☐ |
| 13. I sometimes think about who benefits or gets hurt by the things AI says | ☐ | ☐ | ☐ | ☐ | ☐ |

*Note*. All scores range from 1 to 5. Subscales can be analysed separately.

For scoring and computation:

Verification: (Items: 1 + 2 + 3 + 4 + 5)/5……………………………………….…………score:___

Motivation: (Items: 6 + 7 + 8 + 9)/4…………………………………...….…………..score:___

Reflection: (Items: 10 + 11 + 12 + 13)/4……………………..……..……..………score:___
Critical thinking in AI use: (Verification + Motivation + Reflection)/3….……..…...score:___

Results from the iterative refinement procedure indicated that a three-factor model with 17 items, most of which met the communality, primary loading, and cross-loading thresholds, provided the best overall fit, demonstrating excellent model performance (RMSEA = .05, TLI = .96, CFI = .98, RMSR = .03). Closer inspection, however, revealed that four items lacked adequate conceptual or psychometric support. Two items ("I try to check if AI information is still up to date before using it" and "I try to stay sceptical of AI information until I am sure it is accurate") were removed because they showed substantial cross-loadings and high complexity, making their factor membership ambiguous. Two items ("I often wonder if AI responses are biased in any way" and "I try to look at things in more than one way before deciding if I agree with what AI says") were removed because they did not correspond to their original theoretical dimensions. Removing these four items produced a more parsimonious 13-item solution characterised by a clearer simple structure, stronger construct validity, and excellent overall model fit (RMSEA = .04, TLI = .98, CFI = .99, RMSR = .02). The final 13-item version of the scale is presented in Table 1, while factor loadings are presented in Supplementary Materials Table S11.

## 4. Study 3: First Psychometric Validation

Following the EFA findings in Study 2, Study 3 was conducted with an independent sample to validate the psychometric properties of the 13-item critical thinking in AI use scale with evidence on scale reliability, structural validity, construct validity, and sex invariance. Methodological guidance in scale development recommends that factor structures identified through exploratory analyses should be confirmed in a separate sample rather than derived and confirmed within the same dataset, as this reduces capitalisation on chance and provides a more rigorous test of structural stability (Cabrera-Nguyen, 2010; Morgado et al., 2017; Worthington & Whittaker, 2006). Consistent with broader principles of cross-validation, the use of independent samples across all subsequent validation stages (Studies 4, 5, and 6) was intended to strengthen confidence that the resulting evidence is not sample-specific but replicable across different datasets and study contexts (Boateng et al., 2018).

## 4.1 Method

### 4.1.1 *Participants and Procedure*

We estimated the required sample size for the confirmatory factor analysis (CFA) a priori using the RMSEA-based power analysis framework for covariance structure modeling (MacCallum et al., 1996). This approach evaluates statistical power in terms of the RMSEA, contrasting hypotheses of close fit (e.g., RMSEA $\leq .05$) against those of not-close fit (e.g., RMSEA $\geq .08$). By specifying the model's degrees of freedom, the significance level ($\alpha = .05$), and desired power (1 – the probability of making a Type II error), the method yields the minimum sample size needed to detect whether a model exhibits adequate versus inadequate fit. For our model ($df = 62$), the analysis showed that 183 participants would be sufficient to achieve 80% power. To ensure greater precision and robustness in model estimation, we targeted 99% power, which required a minimum of 376 participants. Based on this estimation, we recruited 376 participants from Prolific to complete the scale. Participants were at least 18 years old, U.S. residents, and ranged in age from 19 to 79 years ($M_{age} = 44.97$, $SD_{age} = 14.06$); 53.46% were female. The demographic distribution of this sample (available in Supplementary Materials Table S4) is similar to the sample in Study 2. Informed consent was obtained electronically prior to participation. The study was administered via Qualtrics and took approximately 11 min to complete. The study included the critical thinking in AI use scale, psychometric scales, and demographic questions. Attention to instructions was verified via three attention check questions spread evenly across the study, which all participants passed.

**4.1.2** ***Measures***

**Critical Thinking in AI Use.** Critical thinking in AI use ($\alpha$ = .88) was assessed using the 13-item version of the scale. Critical thinking in AI use is a multidimensional construct comprising three dimensions: verification of source and content ($\alpha$ = .89; e.g., "I often check AI content to make sure it is correct and clear"), motivation to understand AI ($\alpha$ = .90; e.g., "I try to understand why the AI makes certain recommendations"), and reflection on responsible AI ($\alpha$ = .80; e.g., "I sometimes think about who benefits or gets hurt by the things AI says"). Participants rated each statement on a 5-point Likert scale (1 = *strongly disagree* to 5 = *strongly agree*).

**Critical Thinking Disposition.** Critical thinking disposition ($\alpha$ = .87) was assessed using the 11-item Critical Thinking Disposition Scale (Sosu, 2013). This scale tests the dispositional tendency to approach problems, decisions, and information with critical reflection, open-mindedness, and justification, i.e., whether someone is motivated and inclined to think critically in practice. It includes items such as "I often re-evaluate my experiences so that I can learn from them" and "I usually think about the wider implications of a decision before taking action". Responses were recorded on a 5-point Likert scale (1 = *strongly disagree* to 5 = *strongly agree*).

**Need for Cognition.** Need for cognition ($\alpha$ = .94) was assessed using the 18-item Need for Cognition Scale (Cacioppo et al., 1984). This scale includes items, such as "I would prefer complex to simple problems" and "I find satisfaction in deliberating hard and for long hours", that test an individual's motivational disposition to seek out, engage in, and derive enjoyment from cognitively effortful activities (i.e., how much people like thinking for its own sake). Responses were recorded on a 5-point Likert scale (1 = *Extremely uncharacteristic of me* to 5 = *Extremely characteristic of me*).

**Actively Open-minded Thinking.** Actively open-minded thinking ($\alpha$ = .85) was assessed using the 13-item Actively Open-minded Thinking Scale (Stanovich & Toplak, 2023). This scale includes items related to openness to counterevidence (e.g., "People should always take into consideration evidence that goes against their opinions") and willingness to revise beliefs (e.g., "People should revise their conclusions in response to relevant new information"). Responses were recorded on a 5-point Likert scale (1 = *strongly disagree* to 5 = *strongly agree*).

**Cognitive Reflection.** We administered the 4-item Cognitive Reflection Test−2 (Thomson & Oppenheimer, 2016), which measures cognitive reflection, the tendency and ability to inhibit an initial, intuitive (often wrong) response and engage in deliberate, analytic reasoning before answering. Participants provided either numerical or textual answers to each problem. For example, 83.51% correctly solved "A farmer had 15 sheep and all but 8 died. How many are left?" by responding "8," whereas only 34.84% correctly solved "How many cubic feet of dirt are there in a hole that is 3' deep × 3' wide × 3' long?" by responding "0." In total, 83.24% of participants answered two or more questions correctly.

**Attitudes Toward AI.** Perceived negative attitudes ($\alpha$ = .82) and positive attitudes ($r$SB = .83) toward AI were assessed using the 5-item Attitudes toward Artificial Intelligence Scale (Sindermann et al., 2021). The scale explores attitudes toward AI as a social and technological phenomenon, comprising two positive items related to trust and acceptance (e.g., "I trust artificial intelligence"), and three negative items related to fear of AI and concerns about its potential consequences for humanity (e.g., "I fear artificial intelligence"). Responses were recorded on an 11-point Likert scale (0 = *Strongly disagree* to 10 = *Strongly agree*).

**AI Dependency.** AI Dependency ($\alpha$ = .90) was assessed using the 11-item

Generative AI Dependency Scale (Goh et al., 2025). The scale measures the extent to which individuals develop a psychological and behavioural reliance on generative AI tools through three dimensions: (1) cognitive preoccupation ($\alpha$ = .81; e.g., "I feel an urge to use generative AI, even when it may not be necessary"), (2) negative consequences ($\alpha$ = .77; e.g., "I have trouble completing work or other responsibilities without generative AI"), and (3) withdrawal ($\alpha$ = .96; e.g., "I get frustrated or irritable when I am unable to use generative AI"). Responses were recorded on a 5-point Likert scale (1 = *strongly disagree* to 5 = *strongly agree*).

**Big Five Personality.** Participants' personality traits were assessed using the 15-item Big Five Inventory−2−XS (Soto & John, 2017). Each subscale consisted of three items: agreeableness ($\alpha$ = .65; e.g., "Is compassionate, has a soft heart"), conscientiousness ($\alpha$ = .75; e.g., "Is reliable, can always be counted on"), extraversion ($\alpha$ = .59; e.g., "Is full of energy"), neuroticism ($\alpha$ = .78; e.g., "Worries a lot"), openness ($\alpha$ = .64; e.g., "Has little interest in abstract ideas" [reverse-scored]). Five administered items differed from the intended 15-item BFI-2-XS but were domain-consistent substitutes from the 30-item BFI-2-S (e.g., agreeableness: "tends to find fault with others" [reverse-scored] instead of "assumes the best about people"). Responses were recorded on a 5-point Likert scale (1 = *Disagree strongly*, 5 = *Agree strongly*).

#### 4.1.3 *Analytic Plan*

**Confirmatory Factor Analysis.** The 13-item scale was specified to load on three correlated first-order factors derived a priori. No cross-loadings or correlated residuals were allowed. To evaluate whether the three dimensions reflect a broader concept, we also estimated a higher-order model in which a second-order factor, critical thinking in AI use, loaded on Verification, Motivation, and Reflection. In this model, correlations among first-order factors are explained by critical thinking in AI use. Indicators (5-point Likert) were treated as continuous and estimated with robust maximum likelihood (MLR). Model adequacy was judged against conventional thresholds: CFI/TLI ≥ .90 (acceptable) with preference for ≥ .95 (good) (Hopwood & Donnellan, 2010; Perry et al., 2015), RMSEA ≤ .06 and SRMR ≤ .08 (Shi & Maydeu-Olivares, 2020). The first-order and second-order structures were compared using the robust $\chi^2$ difference test, AIC/BIC, and practical-equivalence criteria (ΔCFI ≤ .01, ΔRMSEA ≤ .015, ΔSRMR ≤ .01; Cao & Liang, 2022; Cheung & Rensvold, 2002). Standardised loadings ≥ .50 with narrow CIs will be considered salient (Kline, 2015).

Evidence based on internal structure was quantified using Average Variance Extracted (AVE). For each first-order factor, AVE was computed from the standardised CFA solution as the proportion of indicator variance accounted for by the factor relative to error; for the second-order model, AVE for the critical thinking in AI use was calculated from the standardised second-order loadings on the three first-order dimensions and the residual variances of those dimensions, all estimated within the same model. We used AVE because it provides an error-adjusted index of how well a latent factor explains its indicators, a direct criterion of measurement quality in Structural Equation Models (SEMs). We treated AVE ≥ .50 as adequate and interpreted values just below .50 in conjunction with standardised loadings and composite reliability (Fornell & Larcker, 1981; Hair et al., 2019). CFA was conducted and measurement invariance was tested using *lavaan* package version 0.6-20 (Rosseel et al., 2025). Additional CFA diagnostics and AVE values were computed using *semTools* version 0.5-7 (Jorgensen et al., 2025).

**Criterion Validity.** We explored how critical thinking in AI use translates into real-world outcomes by using the critical thinking in AI use to predict CRT scores with a simple OLS regression. Given that the CRT was designed to index the capacity to inhibit compelling but incorrect intuitive responses and engage effortful analytic reasoning under conflict (Thomson & Oppenheimer, 2016), we expected individuals higher in critical thinking in AI use, particularly on the reflection facet that captures deliberation about implications and alternative viewpoints, to score higher on the CRT. OLS regressions were estimated using base R's *stats* package (R Core Team, 2025).

**Convergent & Discriminant Validity**. CFA was conducted using the higher-order model, with model fits evaluated similarly. To assess the construct validity of the critical thinking in AI use scale, evidence for both convergent and discriminant validity were tested within a single measurement model. The model specified a second-order latent factor representing overall critical thinking in AI use loading onto the first-order latent dimensions identified during CFA. All comparator constructs were included as observed composites. The magnitude of associations was interpreted using effect size guidelines proposed by Funder & Ozer (2019): $|r| < .05$ (no validity), .05−.09 (very small), .10−.19 (small), .20−.29 (medium), .30−.39 (large), and $\geq .40$ (very large). Convergent validity was assessed by examining critical thinking in AI use correlation with critical thinking disposition, need for cognition, and actively open-minded thinking. Given that both traditional critical thinking and critical thinking in AI use are conceptually aligned with critical thinking disposition as a tendency to engage in careful, truth-seeking evaluation of information (Sosu, 2013), need for cognition as a motivation to invest effort in complex thinking (Cacioppo et al., 1984; Stedman et al., 2009; Taube, 1997), and actively open-minded thinking as the willingness to consider alternative possibilities and revise beliefs in light of new evidence (Stanovich & Toplak, 2023), we expected medium to large positive associations due to their shared emphasis on reflective, effortful, and open-minded processing.

Discriminant validity evidence was assessed by examining the correlation between critical thinking in AI use and attitudes toward AI (including both positive and negative), and AI dependency. To establish evidence for discriminant validity, we evaluated the Fornell–Larcker criterion (Fornell & Larcker, 1981). Specifically, the square root of the AVE for each discriminant validity construct was compared against their standardised correlations with the second-order critical thinking in AI use latent factor. Evidence for discriminant validity is demonstrated when $\sqrt{\text{AVE}}$ exceeds the corresponding correlation ($\sqrt{\text{AVE}} > |r|$), indicating that each construct shares more variance with its own indicators than with the critical thinking in AI use construct. These constructs were selected as discriminant comparators because they represent distinct functional domains within the broader AI user nomological network. Accordingly, we expected only small associations, because critical thinking in AI use reflects effortful evaluative processing of AI-generated information, whereas AI attitudes and AI dependency primarily capture affective evaluations that can arise with relatively little analytic scrutiny.

**Correlates**. The Big Five traits were included to situate critical thinking in AI use within a nomological network that reflects broad, stable personality dimensions (Soto & John, 2017). Prior work on generic critical thinking consistently identifies openness to experience as the only Big Five trait that reliably predicts critical thinking skills, whereas extraversion, agreeableness, conscientiousness, and neuroticism show little to no systematic association (Acevedo & Hess, 2022; Clifford et al., 2004; Ku & Ho, 2010). On this basis, we expected critical thinking in AI use to correlate positively with openness to experience, reflecting the shared emphasis on intellectual curiosity, willingness to entertain alternative ideas, and readiness to revise beliefs in light of new evidence. In contrast, we anticipated near-zero or very small associations with the remaining Big Five traits, which index broad socioemotional and behavioural tendencies that are less directly tied to the evaluative, verification-focused processes captured by the construct of critical thinking in AI use.

**Sex Invariance.** We tested the measurement invariance of the higher-order critical thinking in AI use scale across sex (male vs. female) using multi-group CFA. Establishing invariance by sex is essential to ensure that the scale assesses the same underlying construct in men and women, thereby allowing meaningful comparisons of latent means. This is important given that previous studies generally report little to no overall sex differences in critical thinking (Boonsathirakul & Kerdsomboon, 2021; Fabio et al., 2025; Facione, 1990b; Walsh & Hardy, 1999), so any sex differences observed in critical thinking in AI use scale scores should reflect genuine variation in critical thinking in AI use rather than measurement artefacts.

The higher-order model specified three first-order factors (verification, motivation, reflection), which loaded onto a second-order latent factor (critical thinking in AI use). Following recommended practice (F. F. Chen, 2007; Cheung & Rensvold, 2002), we sequentially evaluated four levels of invariance: (1) configural invariance (no cross-group equality constraints), (2) metric invariance (equal first-order and second-order factor loadings), (3) scalar invariance (equal loadings and intercepts), and (4) residual invariance (equal loadings, intercepts, and residual variances). Models were estimated using the robust maximum likelihood estimator (MLR), which provides robust standard errors and scaled test statistics and allows for comparison of AIC and BIC values. To identify the model, we set the latent variances of each factor to 1 (std.lv = TRUE). We used $\Delta$CFI $\leq .01$ as our primary criterion for establishing invariance (Cheung & Rensvold, 2002).

### 4.2 Results and Discussion

#### 4.2.1 *CFA*

We tested a three-factor CFA for the 13-item scale. The model fit the data well, $\chi^2(62) = 131.75$, $p < .001$; CFI = 1.00; TLI = 1.00; RMSEA = .06, 90% CI [.04, .07]; SRMR = .05. All standardised factor loadings were large and statistically significant (all *p*s < .001), $\lambda_{\text{verification}} = .80–.93$, $\lambda_{\text{motivation}} = .81–.95$, and $\lambda_{\text{reflection}} = .68–.89$. The three first-order factors were moderately correlated with $r_{\text{verification-motivation}} = .48$, $r_{\text{verification-reflection}} = .49$, $r_{\text{motivation-reflection}}$

= .45. Reliability estimates were high with $\omega_{verification}$ = .90, $\omega_{motivation}$ = .90, $\omega_{reflection}$ = .81. Average variance extracted further supported convergent validity with $AVE_{verification}$ = .73, $AVE_{motivation}$ = .77, $AVE_{reflection}$ = .59.

We next estimated a higher-order CFA in which critical thinking in AI use loaded on verification, motivation, and reflection. As expected for this reparameterisation, global fit was identical to the correlated three-factor model with $\chi^2(62)$ = 131.75, CFI = 1.00, TLI = 1.00, RMSEA = .06 with a 90% CI from .04 to .07, and SRMR = .05. Higher-order loadings were all substantial and significant with *p*s < .001. The loadings of critical thinking in AI use on verification, motivation, and reflection were .73 ($R^2$ = .53), .66 ($R^2$ = .44), and .68 ($R^2$ = .46), respectively. These findings indicate that a sizeable proportion of each first-order factor was explained by a common higher-order factor, while meaningful specificity remained at the first-order level. Taken together, the critical thinking in AI use scale demonstrated strong structural validity, high reliability, and evidence of a coherent higher-order critical thinking in AI use factor alongside three interpretable first-order dimensions. The higher-order structure (illustrated in Figure 1) was applied in all subsequent analyses to align with the intended theoretical framework and future use of the scale.

**Figure 1**

*CFA Model of the Critical Thinking in AI Use Scale*

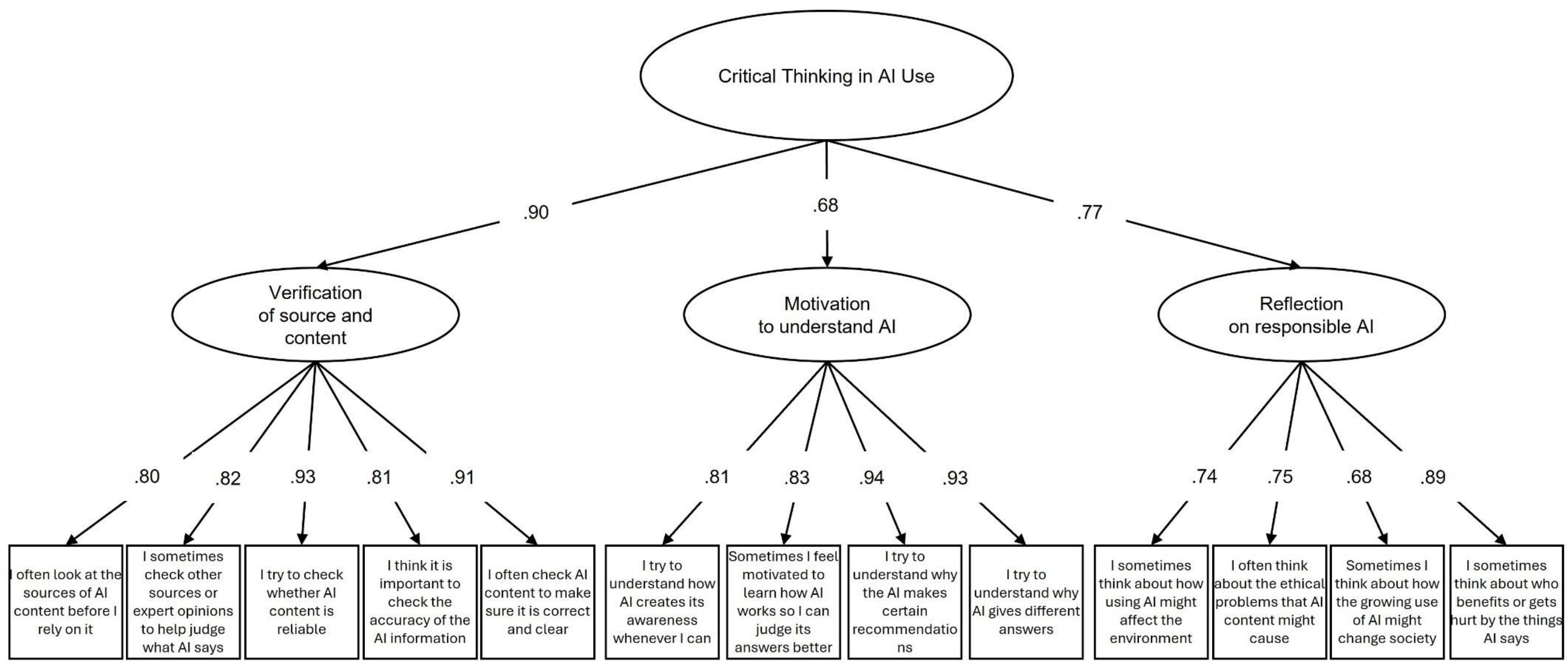

*Note*. $N = 376$. Standardised factor loadings from the higher-order confirmatory factor analysis (CFA) of the critical thinking in AI use scale. The three first-order factors were Verification (items 1–5), Motivation (items 6–9), and Reflection (items 10–13). The higher-order factor, critical thinking in AI use, loaded significantly on all three dimensions. All loadings shown are standardised estimates ($p$s < .001)

**4.2.2 *Criterion Validity***

In a simple OLS model, critical thinking in AI use did not reliably predict CRT performance, $b = -0.16$ ($SE = 0.09$), $t(374) = -1.75$, $p = .08$; 95% CI [−0.34, 0.02], with minimal explained variance, $R^2 = .01$. Consistent with this pattern, using the reflection subscale as the sole predictor yielded a small, marginally negative association, $b = -0.13$ ($SE = 0.07$), $t(374) = -1.95$, $p = .05$; 95% CI [−0.27, 0.00], $R^2 = .01$. Across both analyses, confidence intervals included zero and were compatible with effects ranging from trivially negative to null. According to Funder and Ozer's (2019) guidelines, these effects are very small, indicating that self-reported critical thinking in AI use, and in its reflective facet, shared little variance with CRT.

Although this result diverged from our a priori prediction, it is consistent with a mismatch between the cognitive processes targeted by critical thinking in AI use and those tapped by the CRT. Recent evidence indicates that CRT scores are strongly saturated with general intelligence, numeracy (Otero et al., 2022; Sinayev & Peters, 2015), working-memory efficiency (Engin, 2021), rather than a distinct reflective-thinking disposition (refer to Otero et al., 2022 for meta-analysis) that was meant to be captured by the Reflection facet of critical thinking in AI use. Multiple studies document substantial prior exposure to CRT items, typically 40–50% or more of participants in online and student samples, which inflates scores and makes prior experience and familiarity one of the strongest predictors of high CRT performance (Haigh, 2016; Stieger & Reips, 2016; Thomson & Oppenheimer, 2016). Consistent with this pattern, our CRT data showed 83.24% of participants answering at least two of the four items correctly. Woike (2019) argued that repeated exposure has turned CRT performance into an exercise in "mindless memorisation," rather than a genuine test of the intuitive–analytic conflict it was designed to capture.

**4.2.3 *Convergent & Discriminant Validity***

We estimated a nomological CFA in which critical thinking in AI use was modelled as a second-order factor and was allowed to correlate with convergent and discriminant composites. All standardised covariances between critical thinking in AI use and comparators included in the validity model are indicated in Figure 2. Model fit ($N = 376$) showed strong incremental fit with some residual misfit: $\chi^2(134) = 569.29$, $p < .001$; CFI = .98, TLI = .98; RMSEA = .09, 90% CI [.09, .11]; SRMR = .08. First-order item loadings were strong within each dimension, and the second-order factor loaded robustly on Verification, Motivation, and Reflection (standardised loadings ≈ .66–.73; all $p < .001$), supporting the hierarchical structure.

For convergent validity, critical thinking in AI use correlated positively with cognition-related dispositions as predicted. The largest association was with critical thinking disposition, $r = .70$, $p < .001$ (very large; Funder & Ozer, 2019), followed by need for cognition, $r = .36$, $p < .001$ (large), and actively open-minded thinking, $r = .20$, $p < .01$ (small, approaching medium). This finding is theoretically coherent. Critical thinking in AI use captures applied evaluative engagement with AI outputs, which overlaps most strongly with general critical thinking, substantially with enjoyment of effortful thinking, and more modestly with an openness style that is necessary but not sufficient for verification-oriented behaviour.

For discriminant validity evidence, associations with constructs from distinct functional domains were small to negligible: AI dependency, $r = −.01$, $p = .82$ (none); positive attitudes toward AI, $r = −.13$, $p = .02$ (small, negative); and negative attitudes toward AI, $r = .11$, $p = .06$ (small). Discriminant validity was further supported by Fornell–Larcker tests: in every case, the square root of AVE exceeded the absolute critical thinking in AI use correlation (AI dependency $\sqrt{\text{AVE}} = .77 > |r| = .01$; positive attitudes .84 > .13; negative attitudes .79 > .110). Overall, the pattern showed strong convergence with cognition-related traits and only small links with attitudes and AI dependency, consistent with the theorised nomological network for critical thinking in AI use.

**Figure 2**

*Standardised Covariances Between Critical Thinking in AI Use and Constructs Included for Convergent and Discriminant Validity*

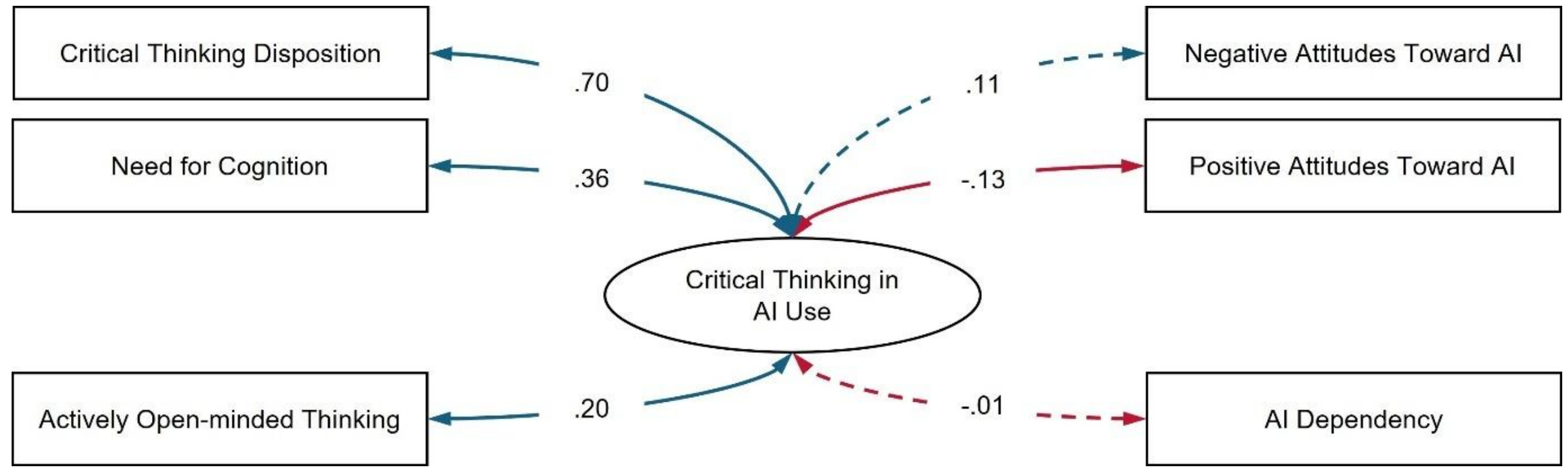

*Note.* $N$ = 376. All values represent standardised covariances. Solid lines indicate significant associations, while dotted lines represent non-significant associations. Blue and red lines indicate positive and negative associations, respectively.

#### 4.2.4 *Correlates*

We estimated a nomological CFA in which critical thinking in AI use was modelled as a second-order factor and allowed to correlate with the Big Five composites. The model fit the data well, robust $\chi^2(122) = 217.00$, $p < .001$, CFI = .97, TLI = .96, RMSEA = .05, SRMR = .05. As shown in Figure 3, the higher-order critical thinking in AI use scale factor was positively associated with extraversion ($r = .21$, $p < .01$; medium) and openness ($r = .21$, $p < .01$; medium), while its relations with agreeableness ($r = .12$, $p = .10$), conscientiousness ($r = .04$, $p = .57$), and neuroticism ($r = .06$, $p = .38$) were small and not statistically reliable.

In a joint subdimension CFA where the three first-order factors were estimated together, model fit was also good, robust $\chi^2(112) = 196.00$, $p < .001$, CFI = .97, TLI = .96, RMSEA = .05, SRMR = .04. Patterns largely mirrored the higher-order factor: Motivation showed medium associations with extraversion ($r = .21$, $p < .001$) and small associations with openness ($r = .14$, $p = .01$), with other traits near zero; Reflection correlated positively with agreeableness ($r = .15$, $p = .04$) and openness ($r = .14$, $p = .03$), and was marginal for neuroticism ($r = .10$, $p = .09$); Verification related small-to-moderately to openness ($r = .13$, $p = .02$) and was marginal for extraversion ($r = .11$, $p = .06$), with other traits negligible. These results suggest that individual differences in critical thinking in AI use align most consistently with an open, exploratory interpersonal style (openness, extraversion), whereas links to conscientiousness and neuroticism are weak or absent.

**Figure 3**

*Correlations Between Critical Thinking in AI Use and Personality Correlates*

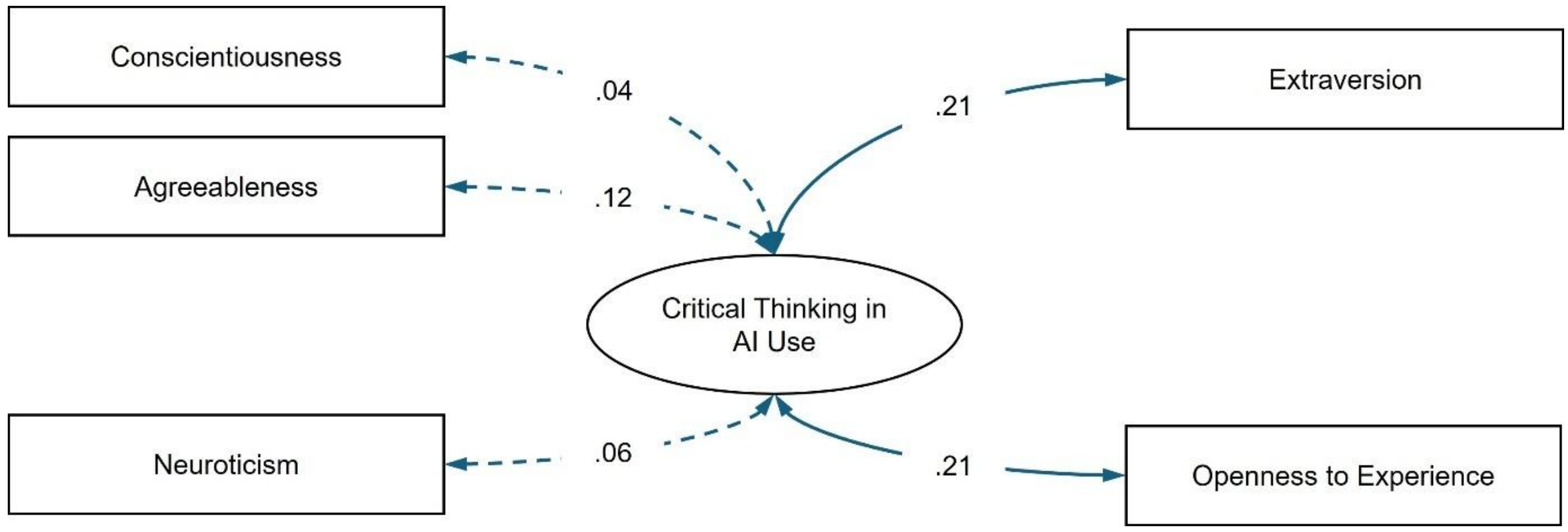

*Note.* $N = 376$. All values represent standardised covariances. Solid lines indicate significant associations, while dotted lines represent non-significant associations. Blue and red lines indicate positive and negative associations, respectively.

### 4.2.5 *Sex Invariance*

We examined the measurement invariance of the higher-order critical thinking in AI

use scale across biological sex using multi-group CFA. Fit indices indicated good model fit at the configural level, $\chi^2(124) = 225.99$, CFI = .97, RMSEA = .06, SRMR = .06, supporting a common factor structure across males and females. Constraining factor loadings to equality (metric invariance) did not reduce model fit, $\chi^2(136) = 236.39$, CFI = .98, RMSEA = .05, SRMR = .06; ΔCFI = +.002 relative to the configural model. Constraining both loadings and intercepts (scalar invariance) likewise produced negligible change, $\chi^2(145) = 248.61$, CFI = .97, RMSEA = .05, SRMR = .07; ΔCFI = –.001 relative to the metric model. Finally, adding residual equality constraints (strict invariance) also resulted in minimal change, $\chi^2(158) = 275.27$, CFI = .97, RMSEA = .05, SRMR = .07; ΔCFI = –.001 relative to the scalar model.

As shown in Table 2, all invariance steps met recommended criteria ($|\Delta CFI| \leq .01$), consistent with recommended criteria for invariance (Chen, 2007; Cheung & Rensvold, 2002). Scaled $\chi^2$ difference tests based on the MLR estimator were nonsignificant for each model comparison, further supporting invariance. Although warnings were issued about the near-singularity of the variance–covariance matrix at the scalar and residual levels, diagnostic checks revealed no zero standard errors, and modification indices suggested only minor localised strain (e.g., residual correlations among specific items). Taken together, these results provide strong evidence for configural, metric, scalar, and residual invariance of the higher-order critical thinking in AI use scale across sex.

**Table 2**

*Measurement Invariance of the Higher-Order Critical Thinking in AI Use Scale Across Sex (Male vs. Female)*

| Model | CFI | ΔCFI | RMSEA | SRMR | AIC | BIC |
|---|---|---|---|---|---|---|
| Configural | .973 | — | .056 | .057 | 11224.50 | 11554.59 |
| Metric (vs. | .975 | +.002 | .052 | .063 | 11210.90 | 11493.83 |
| Configural) | | | | | | |
| Scalar (vs. Metric) | .974 | –.001 | .051 | .065 | 11205.13 | 11452.69 |
| Residual (vs. Scalar) | .973 | –.001 | .050 | .068 | 11205.79 | 11402.27 |

*Note.* ΔCFI values are relative to the immediately preceding model. Values of |ΔCFI| ≤ .01 indicate negligible change in model fit (Chen, 2007; Cheung & Rensvold, 2002). All models were estimated with the MLR estimator.

## 5. Study 4: Second Psychometric Validation

We conducted a second psychometric validation study of the critical thinking in AI use scale to strengthen the nomological network of critical thinking in AI use examining a wider range of correlates and additional measures of discriminant validity evidence, as well as replicating the correlations between critical thinking in AI use and the Big Five personality traits.

### 5.1 Method

#### 5.1.1 *Participants and Procedure*

Similar to Study 3, the RMSEA-based power analysis framework for covariance structure modeling (MacCallum et al., 1996) indicated a minimum of 183 participants to achieve 80% power. Based on this estimation, we recruited a total of 295 participants from Prolific. To ensure adequate familiarity with AI tools, only individuals with prior experience using AI were eligible to participate. During data screening, five participants were excluded for not meeting this inclusion criterion, which was clearly stated in the study description and participant information sheet. Specifically, these participants indicated “Never (I do not use AI at all in my daily life)” in response to the screening question, "How often do you use AI in your daily life?" Consequently, they were removed from all subsequent analyses, resulting in a final sample of 290 participants. A total of 92.41% of participants indicated that they "use AI (at least) a few times a week". All participants were U.S. residents, ranged in age from 20 to 78 years ($M_{age}$ = 42.30, $SD_{age}$ = 12.73); 50.69% female, 93.79% American by nationality, and 63.10% Caucasian White (see Supplementary Materials Table S5 for the full list of demographic characteristics). Informed consent was obtained electronically prior to participation. The survey included four attention check items (e.g., "Please select 'True' for this statement.") that were evenly distributed across the questionnaire. All participants successfully passed at least three of the four checks.

**5.1.2** ***Measures***

**Critical Thinking in AI Use.** We used the 13-item critical thinking in AI use scale, which demonstrated high internal consistencies (total $\alpha$ = .92; verification $\alpha$ = .90; motivation $\alpha$ = .91; reflection $\alpha$ = .87).

**Attitudes Toward AI**. Similar to Study 3, we used the 5-item Attitudes toward Artificial Intelligence Scale (Sindermann et al., 2021), which demonstrated good internal consistency for the positive subscale using the Spearman–Brown coefficient ($r$SB = .85) and for the negative subscale using Cronbach's alpha ($\alpha$ = .77).

**AI Dependency**. Similar to Study 3, we used the 11-item Generative AI Dependency Scale (Goh et al., 2025), which demonstrated high internal consistencies (total $\alpha$ = .93; cognitive preoccupation $\alpha$ = .86; negative consequences $\alpha$ = .84; withdrawal $\alpha$ = .95).

**AI Attachment**. We included AI attachment as a theoretically distinct, socioemotional construct to test whether critical thinking in AI use can be differentiated from attachment bonds felt with AI, thereby providing evidence for the discriminant validity of our scale. AI attachment was assessed with the 15-item AI Attachment Scale (Kasturiratna & Hartanto, 2025). The scale captures people's felt bond with AI across three dimensions: emotional closeness (i.e., intimacy and warmth toward AI; e.g., "I look forward to interacting with AI"), social substitution (i.e., turning to AI for company or support; e.g., "I sometimes find myself turning to AI when I need someone to talk to"), and normative regard (i.e., extending respectful, considerate treatment to AI; e.g., "I try to treat AI respectfully"). Participants responded on a 5-point Likert scale (1 = strongly disagree to 5 = strongly agree). The total scale score was computed by first averaging items within each subscale and then averaging the three subscale means. Higher scores indicate stronger AI attachment. Internal consistencies were high (total α = .94; emotional closeness α = .90; social substitution α = .95; normative regard α = .90).

**Social Desirability.** Social desirability bias was assessed using the 13-item Marlowe-Crowne Social Desirability Scale Form C (Reynolds, 1982). Checking social desirability protects the construct validity and interpretability of critical thinking in AI use by ensuring that the scale is measuring real critical thinking dispositions about AI, not just participants' desire to present themselves as careful, ethical, and rational. The scale included items such as "No matter who I'm talking to, I'm always a good listener" and "I have never intensely disliked anyone". Responses were recorded on a binary scale (True/False). Internal consistency was high (total $\alpha = .80$).

**Big Five Personality.** Similar to Study 3, we used the 15-item Big Five Inventory−2−XS (Soto & John, 2017), all items were accurately administered, and internal consistencies were high (agreeableness α = .77, conscientiousness α = .83, extraversion α = .79, neuroticism α = .83, and openness α = .75).

**Positive and Negative Affect Schedule**. We included trait positive and negative affect to examine whether critical thinking in AI use shows theoretically coherent associations with broader emotional functioning, which would position our scale within a wider nomological network. Affect was assessed with the 20-item International Positive and Negative Affect Schedule (PANAS; Watson et al., 1988). Participants indicated how they generally felt, over the past six months, on ten positive affect adjectives (e.g., "alert" and "inspired") and ten negative affect adjectives (e.g., "upset" and "afraid") using a 5-point scale (1 = never to 5 = always). Subscale scores were computed as the mean of their items; no overall total was calculated. Higher scores indicate greater positive or negative affect, respectively. Internal consistencies were good (positive affect α = .82 and negative affect α = .85).

**Use of AI.** We assessed AI use frequency to examine whether critical thinking in AI use is meaningfully related to real-world engagement with AI tools, adding to the nomological network of the scale. Frequency of AI use was assessed with a single multiple-choice item: "How often do you use AI in your daily life?" Response options were: Never (I do not use AI at all in my daily life), Rarely (once or twice a month), Sometimes (a few times a week), Often (almost every day), and Very frequently (multiple times daily). Responses were coded 0–4 (0 = Never, 4 = Very frequently). Five participants who indicated "Never" failed the inclusion criteria and were removed from the dataset. AI use frequency was analysed as a continuous outcome and used as a criterion for predictive validity; higher scores reflect more frequent AI use.

#### 5.1.3 *Analytic Plan*

Similar to Study 3, we used CFA to evaluate the factor structure of the 13-item critical thinking in AI use scale. We evaluated convergent and discriminant validity evidence similarly to Study 3. Besides attitudes toward AI and AI dependency, we also added AI attachment and social desirability as discriminant constructs. We expected these AI-related scales (AI dependency, AI attachment, and attitudes toward AI) to be conceptually distinct from critical thinking in AI use because the former index socioemotional orientations, whereas the latter captures cognitive processes. We also expect social desirability to be distinct from critical thinking in AI use because it reflects impression management and self-deceptive enhancement rather than the capacity or motivation to verify, reflect, and seek evidence. The motivation to present oneself favourably may encourage self-reported endorsement of verification behaviours to appear agreeable (Nederhof, 1985). Because social desirability is primarily a response-style tendency, we expected only trivial associations with critical thinking in AI use.

Besides the Big Five traits evaluated in Study 3, we included positive and negative trait affect as correlates. Because both positive and negative moods have been suggested to facilitate critical thinking via different cognitive pathways (Lun et al., 2023), we expected small positive associations between critical thinking in AI use and both trait positive and negative affect. We also included the frequency of AI usage as a correlate of critical thinking in AI use. People increasingly use generative AI tools for verification purposes (e.g., cross-checking claims, asking follow-up questions, and iteratively re-prompting) to test the consistency and clarity of AI outputs (Quelle & Bovet, 2024). Accordingly, we anticipated a small positive association between critical thinking in AI use and AI use frequency.

## 5.2 Results and Discussion

### 5.2.1 *CFA*

Once again, we found that the correlated three-factor CFA model fit the data well, $\chi^2(62) = 144.22$, $p < .001$; CFI = .97; TLI = .96; RMSEA = .07, 90% CI [.05, .08]; SRMR = .04. All standardised loadings were substantial and significant (*p*s < .001). For the higher-order model, global fit was identical to the correlated three-factor solution ($\chi^2(62) = 144.22$; CFI = .97; TLI = .96; RMSEA = .07 [.05, .08]; SRMR = .04). We also found that the higher-order loadings were strong and significant (*p*s < .001), indicating that a common higher-order factor explains a large share of variance in each dimension while preserving meaningful specificity. Taken together, these CFA results were similar to those obtained in Study 3, and provide strong evidence of structural validity for the critical thinking in AI use.

**5.2.2 *Discriminant Validity***

To estimate latent correlations with external constructs, we fit a higher-order critical thinking in AI use CFA with observed discriminant constructs allowed to intercorrelate. All standardised covariances between critical thinking in AI use and comparators included in the validity model are indicated in Figure 4. Model fit was acceptable, $\chi^2(122) = 302.04$, $p < .001$; CFI = .94; TLI = .93; RMSEA = .07, 90% CI [0.06, 0.08]; SRMR = .06. Evidence for discriminant validity was supported; constructs outside the cognitive analytic domain showed small or negligible associations with the higher-order critical thinking in AI use factor. Associations with socioemotional tendencies were modest at most: AI dependency, $r = .19$, $p = .01$ (small), and AI attachment, $r = .28$, $p < .001$ (small to moderate). Attitudinal valence toward AI was near zero to small (positive attitudes, $r = .14$, $p = .14$; negative attitudes, $r = -.07$, $p = .50$). Social desirability was essentially null ($r = .01$, $p = .94$). Fornell–Larcker checks led to the same conclusion: in every case the square root of AVE exceeded the absolute correlation with critical thinking in AI use (positive attitudes toward AI $\sqrt{AVE} = .86 > |r| = .14$; negative attitudes toward AI .75 > .07; AI dependency .80 > .19; AI attachment .78 > .28; social desirability .50 > .01). Overall, the evidence indicates that critical thinking in AI use primarily captures a cognitive critical thinking construct with limited overlap with socioemotional dispositions, attitudes toward AI, or response bias.

**Figure 4**

*Standardised Covariances Between Critical Thinking in AI Use and Constructs Included for Discriminant Validity*

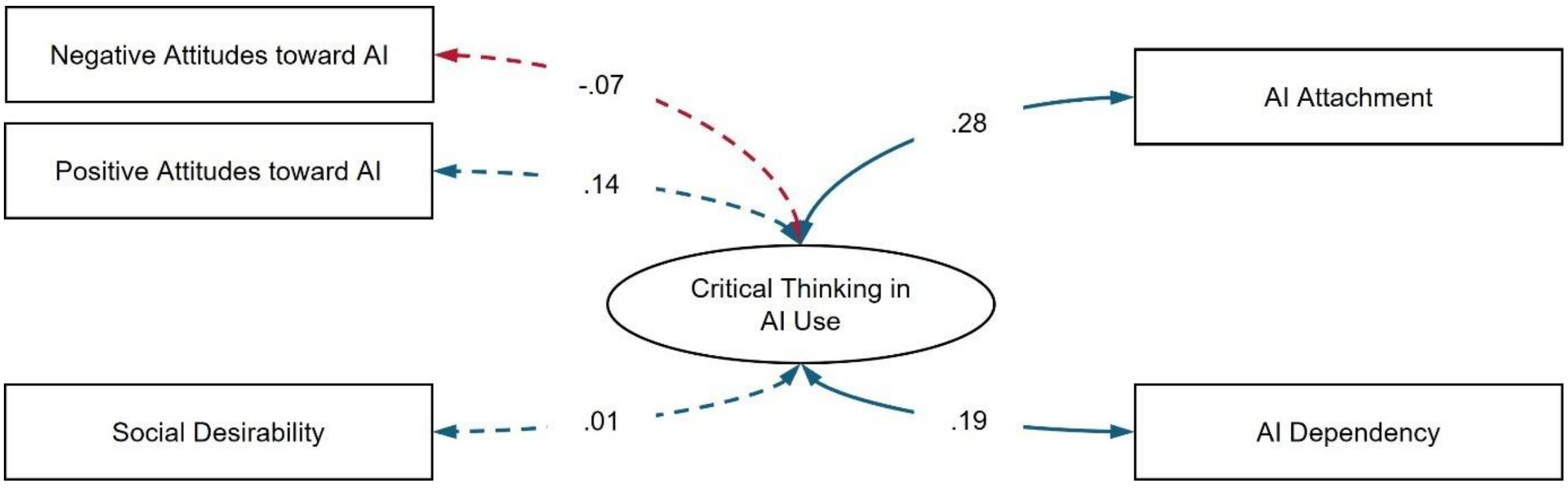

*Note. N* = 290. All values represent standardised covariances. Solid lines indicate significant associations, while dotted lines represent non-significant associations. Blue and red lines indicate positive and negative associations**,** respectively.

### 5.2.3 *Correlates*

We estimated a nomological CFA in which critical thinking in AI use was specified as a second-order factor and allowed to correlate with the Big Five composites, trait affect, and frequency of AI use. Model fit was good, robust $\chi^2(158) = 232.00$, $p < .001$, CFI = .98, TLI = .97, RMSEA = .04, SRMR = .05. As shown in Figure 5, the higher-order critical thinking in AI use factor was positively associated with extraversion ($r = .21$, $p < .01$; medium), openness ($r = .24$, $p < .001$; medium), positive affect ($r = .29$, $p < .001$; medium), and AI use frequency ($r = .26$, $p < .001$; medium). Associations with agreeableness ($r = .05$, $p = .43$), conscientiousness ($r = .07$, $p = .29$), and neuroticism ($r = −.01$, $p = .84$) were small and not reliable, and the link with negative affect was small and marginal ($r = .11$, $p = .07$).

A joint first-order CFA that estimated the three critical thinking in AI use subdimensions together also fit well, robust $\chi^2(142) = 203.00$, $p < .001$, CFI = .98, TLI = .97, RMSEA = .04, SRMR = .03. Patterns largely echoed the higher-order factor. Motivation correlated with extraversion ($r = .22$, $p < .001$; medium), openness ($r = .19$, $p < .01$; small–medium), positive affect ($r = .31$, $p < .001$; medium), and AI use frequency ($r = .26$, $p < .001$; medium), with other traits near zero. Reflection showed positive relations with openness ($r$ =

.21, $p < .01$; small–medium) and negative affect ($r = .22$, $p < .001$; small–medium), was marginal for neuroticism ($r = .13$, $p = .05$), and negligible otherwise, including AI use frequency ($r = .08$, $p = .25$). Verification related to openness ($r = .19$, $p < .01$; small–medium), positive affect ($r = .21$, $p < .01$; small–medium), extraversion ($r = .14$, $p = .02$; small), and AI use frequency ($r = .23$, $p < .001$; small–medium), with other traits negligible or marginal (negative affect $r = .09$, $p = .10$). These results suggest that individual differences in critical thinking in AI use align most consistently with an open, agentic, and positively engaged profile, and are also higher among more frequent AI users, whereas links to conscientiousness, agreeableness, and neuroticism are weak or absent.

**Figure 5**

*Correlations Between Critical Thinking in AI Use, Personality, Affect, and Frequency of AI Usage Correlates*

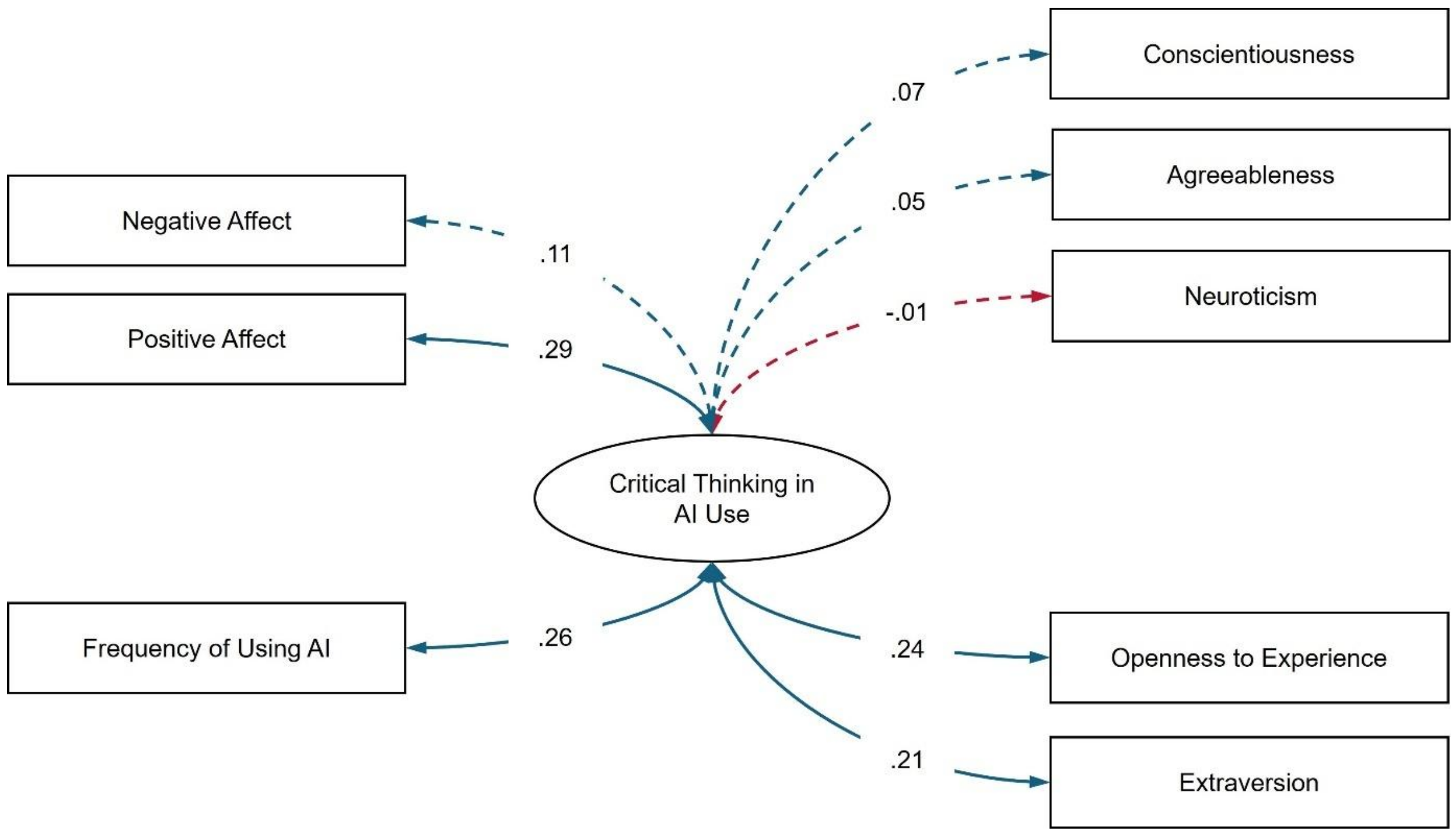

*Note.* $N = 290$. All values represent standardised covariances. Solid lines indicate significant associations, while dotted lines represent non-significant associations. Blue and red lines

indicate positive and negative associations, respectively.

# 6. Study 5: Test-Retest Reliability

Study 5 was conducted to examine the test-retest reliability of the 13-item critical thinking in AI use scale over 1-week and 2-week intervals. The 1-week interval follows recommendations for assessing short-term stability while minimising recall effects (Duff, 2014; Windle, 1955), whereas the 2-week interval extends this window to further reduce recall effects while still capturing stable individual differences (Poder et al., 2022; Streiner et al., 2024).

## 6.1 Method

### 6.1.1 *Participants and Procedure*

Using a precision-based approach for test-retest reliability, we sized the study to achieve narrow 95% CI around ICC(2,1) (Koo & Li, 2016). We estimated an ICC of .80, Bonett's CI-width method (Bonett, 2002) indicated that a 95% CI width of .15 required $n$ = 89 completers. For subscales, translating the expected coefficient alpha of .80 to ICC ≈ .72 and fixing the width at .20 yielded $n$ = 90. These results informed us that we would require a minimum of 90 participants for examining test-retest reliability. For the 1-week test-retest interval, 92 participants ($M_{age}$ = 21.13, $SD_{age}$ = 1.69; refer to Supplementary Materials Table S6 for participant demographics) were recruited from a university subject pool system in Singapore. For the 2-week test-retest interval, we extended participant recruitment through a Telegram channel that advertises psychology experiments, and recruited 122 participants ($M_{age}$ = 24.15, $SD_{age}$ = 3.57; see Supplementary Materials Table S7 for participant demographics) who completed the scale on both sessions.

### *6.1.2 Analytic Plan*

Test-retest reliability of the scale was assessed using both the intra-class correlation coefficient (ICC) and Pearson's zero-order bivariate correlation coefficient (*r*), computed between the two assessment points: initial administration of the scale (baseline) and the follow-up in the second session. ICC was estimated using a two-way random-effects model with absolute agreement to capture the extent to which individual scores remain stable across time, reflecting both consistency and agreement in measurement (Bruton et al., 2000). Interpretation of ICC values was based on the 95% confidence interval, following conventional thresholds: values below .50 indicate poor reliability, values between .50 and .75 moderate reliability, between .75 and .90 good reliability, and above .90 excellent reliability (Koo & Li, 2016). Pearson's correlation coefficient (*r*) was also calculated to evaluate the linear association between scores at the two time points. Correlation strength was interpreted in accordance with established guidelines for effect size evaluation in psychological research (Funder & Ozer, 2019). ICCs for test-retest reliability were calculated using *merTools* version 0.6.3 (Knowles & Frederick, 2025). Pearson correlations used to quantify rank-order stability were computed using base R's *stats* package (R Core Team, 2025).

**6.2 Results and Discussion**

The 13-item critical thinking in AI use scale demonstrated temporal stability over 1- and 2-week test-retest intervals. For a 1-week test-retest interval, the total score showed moderate-to-good test-retest consistency, $r = .70$, 95% CI [.58, .79], $p < .001$, and similar absolute agreement, ICC(2,1) = .70, 95% CI [.58, .79], $p < .001$. Subscale stability was comparable: verification $r = .75$, ICC(2,1) = .75, 95% CI [.64, .82]; motivation $r = .63$, ICC(2,1) = .63, 95% CI [.49, .74]; reflection $r = .70$, ICC(2,1) = .70, 95% CI [.58, .79]; all *p*s < .001.

For a 2-week test-retest interval, the total score showed moderate test-retest consistency, $r = .65$, 95% CI [.53, .74], $p < .001$, and similar absolute agreement, ICC(2,1) =

.65, 95% CI [.53, .74], $p < .001$. Subscale stability was comparable: verification $r = .66$; ICC(2,1) = .66, 95% CI [.54, .75]; motivation $r = .63$; ICC(2,1) = .63, 95% CI [.50, .72]; reflection $r = .61$; ICC(2,1) = .61, 95% CI [.48, .71]; all $p$s < .001. Item-level ICCs ranged from .44 to .71 for the 1-week test-retest interval, and ranged from .36 to .64 for the 2-week test-retest interval, indicating some heterogeneity at the item level, but aggregation to subscales and the total produced reliably stable scores over time. The item-level ICC results are available in Supplementary Materials Table S9 for the 1-week test-retest interval, and Table S10 for the 2-week test-retest interval. Overall, the close correspondence between Pearson correlations (rank-order consistency) and ICCs (absolute agreement) supports the use of the scale for repeated-measures and individual-differences applications.

## 7. Study 6: Criterion Validation

Study 3 found no association between critical thinking in AI use and performance on the Cognitive Reflection Task, a numeracy-laden, familiarity- and memory-dependent task that is only weakly aligned with the demands of evaluating AI-generated information. To address this mismatch, Study 6 provided a more stringent, ecologically valid test of the scale's criterion validity using tasks that closely matched its core processes. To this end, we created two novel tasks: (1) a GPT-4o-powered AI chatbot fact-checking task that closely resembled how people query LLMs for information, and (2) a reflection-writing task on the topic of responsible AI that was designed to tap the reflection subfactor of critical thinking in AI use. These tasks yielded objective indices of how accurately participants judged the veracity of generative-AI outputs and how deeply they reflected on responsible AI. In addition, we also examined self-reported behavioural tendencies, including frequency of verification when using AI, diversity of verification methods, and depth of reflection on responsible AI.

### 7.1 Method

**7.1.1 *Participants***

The target sample size for Study 6 was determined a priori based on the expected association between critical thinking in AI use and task-based indices of AI evaluation (e.g., veracity judgement accuracy) which prior work suggests being in the small-to-moderate range ($r \approx .22$–.40; Orhan, 2022; Pennycook & Rand, 2019). To be conservative, we based our power analysis on detecting a correlation of $r = .22$ with $\alpha = .05$ (two-tailed). This analysis indicated that approximately 159 participants would be required to achieve 80% power. We recruited 200 adults from Prolific. Nine participants were excluded for noncompliance, incomplete responses, or extreme completion times (>3 *SD* from the median). The resulting 191 participants ranged in age from 19 to 78 years ($M_{age} = 44.65$, $SD_{age} = 14.34$), 50.26% were female, and 66.49% were White. The full demographic data are available in Supplementary Materials Table S8. Informed consent was obtained electronically prior to participation. The survey, which took approximately 17 min to complete, was administered via Qualtrics. Participants received £1.70 as compensation. Attention to instructions was verified using two attention check items (e.g., "Please select 'Disagree' for this statement"), which all participants passed. Participants completed the 13-item critical thinking in AI use scale, an AI chatbot fact-checking task, and a reflection writing task.

**7.1.2 *Measures and Procedure***

**Fact-checking Task.** We developed a fact-checking task designed to closely mirror how participants verify information when using generative AI, following Hu et al.'s (2026) step-by-step guidelines for creating AI chatbots in psychological research. Participants evaluated 12 statements (6 true, 6 false) presented in random order through a ChatGPT-4o–powered AI chatbot interface embedded in Qualtrics. Independently, each item was assigned one of four source link conditions: None (disabled and not clickable), Broken (clickable but fails to load), Working-irrelevant (valid page with irrelevant information), or Working-relevant (valid page with relevant information). These source link conditions were meant to mimic the full range of source link scenarios typical of LLM outputs. For both the six true and six false items, we set a balanced quota between useable evidence present and absent: three statements had working-relevant source links (evidence present), and the other three offered no useable on-page evidence (one working-irrelevant link, one broken link, and one with no link). The mapping of source link condition to specific statements was randomised for every participant, so across participants, any statement could appear under any source link condition. This design prevented systematic pairings of specific statements or veracity with particular source link conditions that could bias verification performance. The full pool of 48 items (12 base statements x 4 source link conditions) can be found in Supplementary Materials Table S12.

Participants judged each statement on whether they believe it was Correct, Incorrect, or Unsure. The system used Vercel as serverless middleware to proxy requests securely between Qualtrics and OpenAI's API, preventing API-key exposure, and it returned brief, neutral acknowledgements after each response without revealing correctness. Task performance was operationalised as percentage correct across 12 items (0–100%), where an item was scored correct only when true statements were marked Correct and false statements were marked Incorrect. The instructions for the fact-checking task are displayed in Supplementary Materials Figures S2 and S3. Figure 6 shows the interface of the GPT-4o-powered AI chatbot fact-checking task.

**Figure 6**

*Interface of the GPT-4o–Powered AI Chatbot Fact-Checking Task*

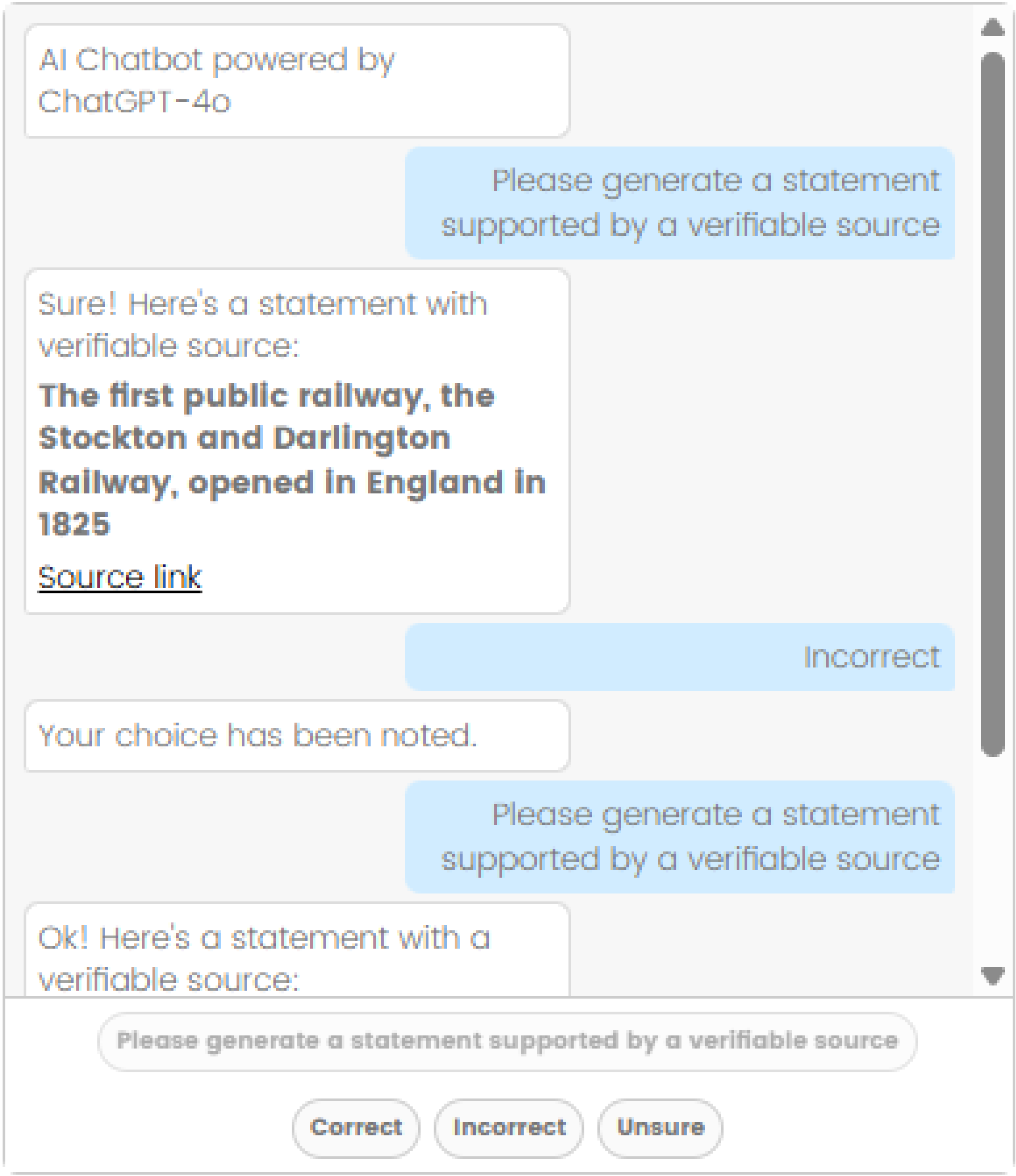

Immediately after the fact-checking task, participants completed a single self-report measure of how much they verified statements during the task. The prompt read: "*Which of the following best describes your behaviour earlier when determining whether the statements were correct/unsure/incorrect?*" Responses were collected on a five-point ordered scale, which we coded on a scale from 0 to 4, with higher scores indicating greater self-reported verification (0 = "*I did not verify any statements*", 1 = "*I verified a few statements*", 2 = "*I verified some statements*", 3 = "*I verified most statements*", 4 = "*I verified all statements*"). Participants then indicated the methods they used to verify statements. They could select one or more options from a predefined list: (a) clicked on the source link(s) provided, (b) used online search engines (e.g., Google Search), (c) used other AI tools (e.g., ChatGPT), or (d) others, with an open-text field for additional methods. Participants also explained why they chose these methods. During data cleaning, we found that 10 participants' "*Others*" responses reflected prior knowledge (e.g., "*my own memory*"), and these were not counted as verification. Only one "*Others*" response was retained, as the participant reported verifying the information by asking a friend. We computed each participant's count of unique verification-method categories endorsed, yielding a possible range of 0 to 4, with higher scores indicating a greater diversity of verification methods.

**Reflection Writing Task.** Participants completed a 5-min written reflection to assess their capacity to reflect on the use of generative AI. The task prompt instructed participants to write freely in their own words, with several guiding themes provided to encourage breadth and depth of thought. These included: (a) how the use of AI may affect them personally, (b) how it may influence others around them, (c) its potential societal impacts, and (d) possible ethical questions arising from AI use. Participants were reminded not to use any generative AI tools that would assist with their writing. Participants could respond in paragraph or point form, and were informed that there were no right or wrong answers. Participants' reflections were scored using an automated rater implemented via a large-language-model API, specifically, OpenAI's GPT-4o. Recent research has demonstrated that large language models can assist researchers by automating deductive coding in qualitative studies by ensuring consistency while promoting efficiency (Chew et al., 2023; Landerholm, 2025; Xiao et al., 2023). Reflection scores ranged from 0 to 4, with a higher score indicative of deeper and more thoughtful reflection. The coding rubric and system prompt used are shown in Supplementary Materials Figure S4.

Immediately after the reflection task, participants provided a self-assessment of the depth and quality of their reflection. Using a continuous slider scale ranging from 0 to 100, they rated the extent to which they had engaged in deep and thoughtful reflection during the earlier writing task. A score of 0 indicated no reflection at all (for example, superficial or impersonal writing with little engagement), whereas a score of 100 indicated very deep and thoughtful reflection (for example, critically examining multiple perspectives and considering implications for oneself, others, and society).

#### *7.1.3 Analytic Plan*

**Verification Behaviour and Fact-checking Task Performance.** For our primary analyses, we ran a simple OLS regression using critical thinking in AI use to predict verification frequency, number of unique verification methods, veracity judgement accuracy on the fact-checking task, self-reported reflection depth, LLM-scored reflection depth on the writing task. We expect a positive relationship between critical thinking in AI use with self-reported verification behaviour (i.e., verification frequency and number of unique verification methods) and performance (i.e., veracity judgement accuracy) on the fact-checking task. This is because users higher in critical thinking in AI use would tend to engage in active corroboration of AI outputs by checking source attribution and evaluating the reliability and accuracy before relying on them. Additionally, users higher on critical thinking in AI use are expected to be more motivated to understand how AI generates responses in order to judge AI outputs more effectively.

For secondary exploratory analyses, we computed three process-level indices that capture how participants used or withheld evidence while judging claims. We explored the relationship between critical thinking in AI use (and its verification facet) and these three indices using simple OLS regressions. For each item, the source hyperlink condition determined whether useable evidence was present (a working, relevant page) or absent (no link, a broken link, or a working but irrelevant page). Each participant received a balanced set of items across these conditions, and indices were computed per participant (0-100%). Firstly, Evidence Extraction captured accuracy when evidence was present, defined as the proportion of items correctly classified (counting False items as Incorrect, and True items as Correct) when a working-relevant source link is provided. Secondly, Misread Rate indexed failure to recognise or apply useable evidence. It was defined as the proportion of trials, with a working link and relevant information, on which the response was Unsure. By focusing on trials with credible and accessible supporting evidence, Misread Rate isolated uncertainty in the face of valid evidence rather than general inaccuracy. Finally, Evidence-Gap Resolution indexed accurate, decisive, evidence-seeking judgements when no relevant on-page source link was available. Evidence-Gap Resolution was defined as the accuracy when relevant evidence was absent, specifically the None, Broken, or Working-irrelevant link conditions, on which the participant accurately selected Correct or Incorrect rather than Unsure. We have provided a summary of these three indices in Table 3, and more details on these three indices (Evidence Extraction, Misread Rate, Evidence-Gap Resolution) in Supplementary Materials Table S13.

**Table 3**

*Indices of Verification Behaviour and Performance on the AI Chatbot Fact-checking Task*

| Indices | What it means |
|---|---|
| Evidence extraction | % Correct on Working-relevant link items (uses the on-page evidence to accept/reject the claim) |
| Misread rate | % Unsure on Working-relevant link items (missed useable evidence) |
| Evidence-gap resolution | % Correct on evidence-absent items (None, Broken, or Working-irrelevant link conditions) |

**Reflection Scores.** To ensure robustness and replicability of the automated assessment, we cross-validated the reflection scores, coded automatically by OpenAI's GPT-4o, using Anthropic's Claude Opus 4.1. We selected OpenAI's and Anthropic's LLMs as they have demonstrated stronger deductive and inductive reasoning abilities (Cai et al., 2025), greater AI alignment with human values (Lau et al., 2025), and have often been used for qualitative analyses (Morgan, 2023; Simon et al., 2025). Agreement between the two models was good: ICC(2,1) = .79, 95% CI [.73, .84], $F(190, 191) = 8.42$, $p < .001$; two-way random-effects, single-measure, absolute agreement. We ran simple OLS regressions using the critical thinking in AI use to predict the reflection score on the writing task as well as self-reported reflection. We expected a positive relationship between critical thinking in AI use and both reflection metrics (self-reported and LLM-scored reflection depth), as AI users who scored higher in critical thinking in AI use would tend to engage in deliberate consideration of the broader impacts of AI use, including ethical issues, environmental effects, societal change, and who stands to benefit or be harmed by AI systems and their outputs.

### 7.2 Results and Discussion

Higher levels of critical thinking in AI use were consistently associated with more engaged verification behaviour on the fact-checking task (all regression results, including the supplementary analyses, are reported in Supplementary Materials Table S14). As expected, participants who scored higher on critical thinking in AI use reported verifying statements more frequently during the task ($b = 0.28$, $SE = 0.14$, $t = 2.03$, $p = .04$, $R^2 = .02$). They also reported using a broader range of verification methods ($b = 0.15$, $SE = 0.06$, $t = 2.45$, $p = .02$, $R^2 = .03$), indicating that higher critical thinking in AI use scorers not only verified more often but also did so through multiple strategies such as checking source links, searching online, or consulting other AI tools. These findings align with the notion that individuals with stronger critical thinking dispositions toward AI are more likely to engage in active corroboration and cross-validation of AI outputs rather than relying on them passively. Fact-checking task performance data further reinforced this interpretation. Critical thinking in AI use significantly predicted veracity judgement accuracy ($b = 6.91$, SE = 1.92, $t = 3.59$, $p < .001$, $R^2 = .06$). This suggests that participants who were more motivated, and actually engaged in more verification behaviour, were better at discriminating true from false claims.

Higher levels of critical thinking in AI use were also consistently associated with deeper reflection about AI use on the reflective writing task. Critical thinking in AI use predicted both self-reported reflection depth ($b = 7.29$, $SE = 1.73$, $t = 4.20$, $p < .001$, $R^2 = .09$), and LLM-coded reflection depth on the writing task ($b = 0.21$, $SE = 0.08$, $t = 2.57$, $p = .01$, $R^2 = .03$). These results suggest that critical thinking in AI use extends beyond verification to encompass metacognitive awareness and reflection on the moral and societal implications of AI usage. Taken together, participants higher in critical thinking in AI use not only verified more but also reflected more deeply on issues such as fairness, accountability, and ethical consequences of AI use. Figure 7 summarises the significant associations between critical thinking in AI use and criterion validity outcomes.

**Figure 7**

*Associations Between Critical Thinking in AI Use and Criterion Validity Outcomes in Study 6*

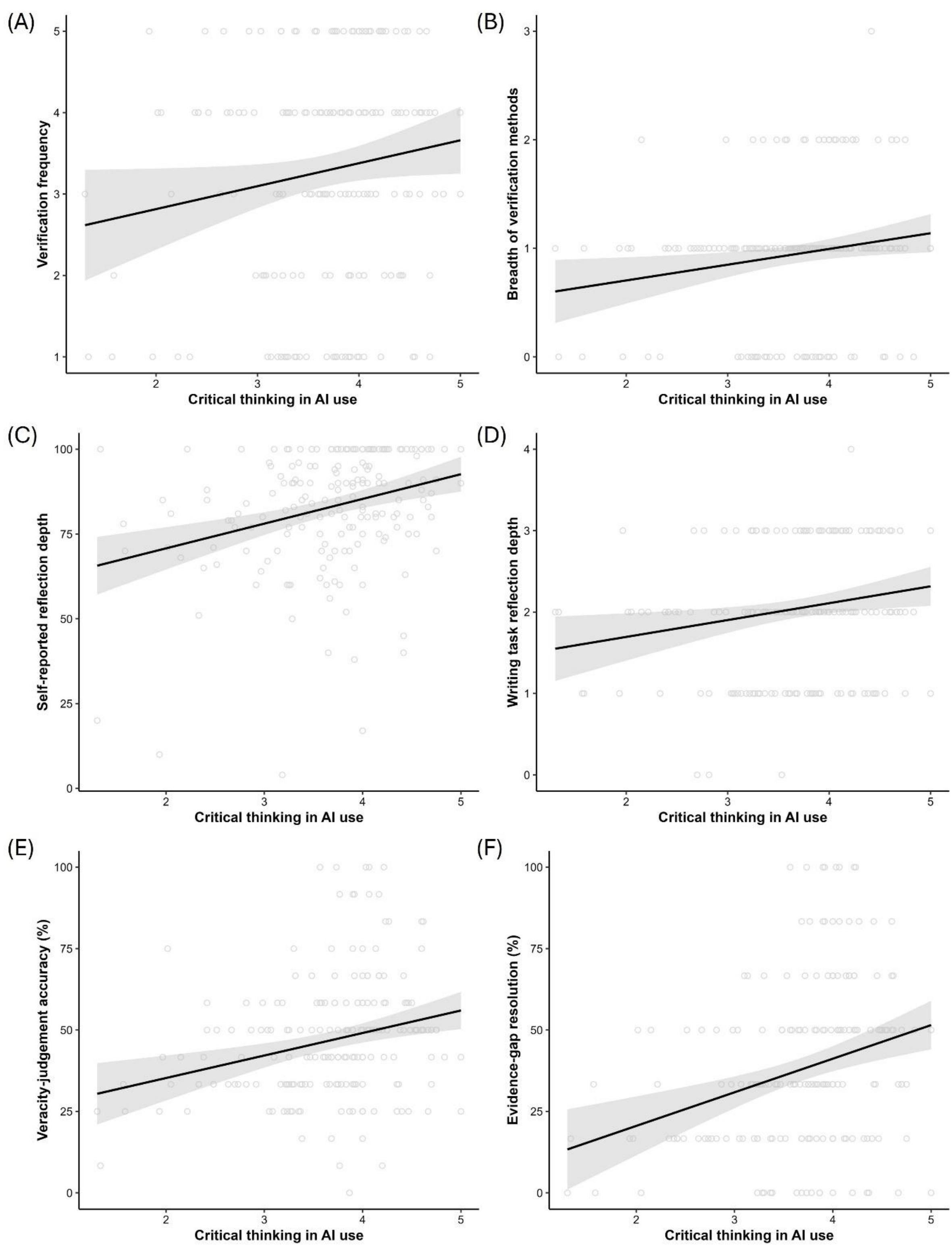

*Note*. Each panel displays the simple linear regression of a criterion outcome on critical thinking in AI use (total score). Solid lines represent OLS regression lines; shaded bands represent 95% confidence intervals. Higher critical thinking in AI use predicted (A) more frequent verification of AI-generated statements, (B) a broader repertoire of verification methods, (C) greater self-reported reflection depth, (D) deeper reflection on a writing task, (E) higher veracity judgement accuracy on the GPT-powered fact-checking task, and (F) greater evidence-gap resolution, reflecting more accurate judgements when on-page evidence was absent or unreliable. All associations were statistically significant. Panels A–C reflect self-reported behavioural measures; panels D–F reflect task-based performance indices.

For exploratory purposes, critical thinking in AI use showed a small, non-significant positive association with Evidence Extraction ($b = 3.51$, $SE = 2.25$, $t = 1.56$, $p = .12$, $R^2 = .01$), while the Verification subscale significantly predicted higher Evidence Extraction ($b = 4.69$, $SE = 1.89$, $t = 2.48$, $p = .01$, $R^2 = .03$) on the fact-checking task. This suggests that the verification facet of critical thinking in AI use was the most proximal driver of effective use of the relevant working link sources provided. Participants with higher verification scores were more likely to access the link, evaluate the credibility of its contents, and apply that evidence to their judgements rather than rely on general knowledge. We also observed that higher critical thinking in AI use showed a marginal reduction in Misread Rate ($b = -3.48$, $SE = 1.96$, $t = -1.78$, $p = .08$, $R^2 = .02$), consistent with less uncertainty when useable evidence was present; the Verification subscale showed a similar but non-significant tendency ($b = -2.35$, $SE = 1.67$, $t = -1.41$, $p = .16$, $R^2 = .01$). This pattern suggests that individuals higher in critical thinking in AI use were slightly more adept at recognising and using on-page evidence, applying it to interpret the statement, and committing to a judgement rather than being "Unsure" on the fact-checking task. Notably, we found that critical thinking in AI use significantly predicted higher Evidence-Gap Resolution ($b = 10.32$, $SE = 2.52$, $t = 4.10$, $p < .001$, $R^2 = .08$), and the Verification subscale showed a similarly strong association ($b = 8.43$, $SE = 2.14$, $t = 3.93$, $p < .001$, $R^2 = .08$). Participants higher in critical thinking in AI use were less likely to indicate "Unsure" when the AI chatbot did not provide a working link with relevant information and were more likely to render an accurate and definitive judgement of "Correct" or "Incorrect". This finding is best interpreted as a form of adaptive verification behaviour where higher critical thinking in AI use participants sought corroboration beyond the AI chatbot when evidence was absent. Taken together, these results indicate that when relevant evidence is present, high AI critical thinkers extract and apply them with certainty; when relevant evidence is absent, high AI critical thinkers seek verification beyond the AI chatbot rather than default to uncertainty.

## 8. General Discussion

As generative AI becomes integral to workplace and educational settings, the central question is no longer just whether we can use these tools, but whether AI users can critically evaluate AI outputs, in light of its persuasive fluency, hallucinations, opaque training data, such that human judgement remains in control. Yet critical thinking in AI use remains empirically underexplored and conceptually underdefined, with few tools that capture the distinctive demands of generative AI use and little work explaining why individuals differ in critical thinking in AI use or how these differences shape their behaviour and outcomes. To address this gap, we developed and validated the critical thinking in AI use scale, a theory-driven instrument that assesses individual differences in AI output content and source verification, motivation to understand AI systems, and reflection on the broader implications of AI use. Across six studies, the scale demonstrated strong psychometric properties, a stable three-factor structure with a higher-order factor, robust construct validity, measurement invariance across sex, and test-retest reliability over one- and two-week intervals. Higher critical thinking in AI use was linked to a user profile of openness, extraversion, trait positive affect, and more frequent AI use. Notably, higher critical thinking in AI use predicted more frequent verification behaviour, greater diversity of verification strategies, higher fact-checking accuracy, and deeper reflection. Taken together, these findings position the critical thinking in AI use scale as a reliable and valid tool for advancing research on critical thinking in the context of using generative AI tools, clarifying its correlates, and informing interventions to support more responsible AI use.

### 8.1 Psychometric Properties of the Critical Thinking in AI Use Scale

The critical thinking in AI use scale was developed through a theory-driven, multi-stage process designed to ensure strong content validity and structural soundness. In Study 1, item generation drew on contemporary accounts of critical thinking (Sosu, 2013; Stupple et al., 2017; Valenzuela et al., 2017) and dual-process theories (Chaiken, 1980; S. Chen & Chaiken, 1999; Kahneman, 2011) adapted to capture the distinctive demands of interacting with generative AI systems. Through expert review and multiple rounds of content validation with naïve judges, a large initial item pool was trimmed to 27 items with excellent substantive agreement and high readability, indicating that the items were both conceptually well-targeted and accessible to a broad adult audience. Across Studies 2, 3, and 4, EFA and CFA converged on a 13-item measure with a stable three-factor structure corresponding to Verification (scrutinising and corroborating AI outputs), Motivation to understand AI (seeking to learn how AI systems work and where they fail), and Reflection (considering ethical, societal, and long-term implications of AI use). A higher-order CFA further showed that these first-order dimensions are underpinned by a common critical thinking in AI use factor, with substantially higher-order loadings that leave room for meaningful specificity at the subscale level. This structure aligns with broader critical thinking frameworks that distinguish between evaluative practices, epistemic motivation, and reflective judgement, while accommodating the distinct demands of AI-mediated reasoning (Ennis, 1985, 2011; Facione, 1990a; King & Kitchener, 1994, 2004). Notably, the critical thinking in AI use scale extends existing domain-general critical thinking measures by providing a concise, domain-specific operationalisation that directly targets the demands posed by generative AI in the contemporary, technology-saturated information era.

The scale showed excellent internal consistency at both the total and subscale score levels. Reliability estimates for the facets of Verification, Motivation, and Reflection consistently exceeded conventional thresholds. The three-factor structure was well-replicated across Studies 3 and 4, with the higher-order critical thinking in AI use factor explaining a substantial share of variance in each subscale. Building on this foundation, the scale exhibited measurement invariance across sex, indicating that the scale functions similarly across demographic groups and have comparable meaning and measurement properties for male and female users. Moreover, the present research demonstrated in Study 5 that the scale has good temporal stability. Over one- and two-week intervals, total and subscale scores exhibited moderate-to-good test-retest reliability, with good absolute agreement, indicating that critical thinking in AI use is a relatively stable individual difference rather than a transient state or random fluctuations. By demonstrating a replicated hierarchical structure, temporal stability, and sex invariance, the findings support using the scale for theory testing and for longitudinal and cross-group research, allowing meaningful comparisons of critical thinking in AI use within individuals over time and between demographic groups.

The present research demonstrated evidence for convergent and discriminant validity in Studies 3 and 4, which further supported the psychometric robustness of the scale. As predicted, the scale showed large associations with critical thinking disposition and need for cognition, and small-to-moderate associations with actively open-minded thinking. The positive associations between critical thinking in AI use and related critical thinking constructs suggest that critical thinking in AI use builds on general critical thinking skills and the enjoyment of effortful cognitive endeavours. In contrast, correlations with AI dependency, AI attachment, positive and negative attitudes toward AI, and social desirability were small or negligible, and Fornell–Larcker tests confirmed that these comparators shared more variance with their own indicators than with critical thinking in AI use. The near-zero association with social desirability indicates that scores on the critical thinking in AI use scale were unlikely to be inflated by a general tendency to self-present as thoughtful or cautious (Lanz et al., 2022). Our results are consistent with prior work showing that critical thinking dispositions are only weakly correlated with domain-specific attitudes (Merma-Molina et al., 2022), and dependency-type constructs (Thomas, 2020). In addition, the present findings extend the critical thinking literature by demonstrating that an AI-specific critical thinking construct can be clearly distinguished from a range of socio-emotional orientations toward AI while still showing strong links to core critical thinking dispositions. This pattern supports the view that critical thinking in AI use constitutes a distinctive, AI-focused form of critical thinking, centred on the evaluation of opaque, probabilistic AI outputs and their downstream consequences, rather than merely reflecting traditional critical thinking or simple reliance on, or sentiment toward, AI systems.

Study 6 demonstrated criterion validity of the scale by extending these correlational findings to task-based performances and behavioural outcomes that closely mirrored the core processes embedded in the critical thinking in AI use construct. Critical thinking in AI use scale scores predicted, with a moderate effect size, higher veracity judgement accuracy on a GPT-powered AI chatbot fact-checking task, indicating that individuals who tend to verify AI outputs and seek to understand model behaviour, might be better at distinguishing true from false claims generated by LLMs. These results are consistent with previous studies that have also found a positive relationship between critical thinking and more accurate veracity judgements (Machete & Turpin, 2020; Sultan et al., 2024), and demonstrate how critical thinking in AI use may translate into objective, measurable differences in how people actually interact with AI-generated content. This task-based performance, which extends beyond self-report, was complemented by participants' self-reported behavioural measures. Although these self-reported indices (verification frequency and verification method diversity) remain subject to shared method variance with the scale, they converge with the task-based findings to provide a consistent pattern. Participants higher in critical thinking in AI use reported verifying AI-generated statements more frequently during the fact-checking task and using a broader repertoire of verification methods, such as clicking source links, consulting independent websites, or cross-checking with other AI tools. In other words, individuals high in critical thinking in AI use tend to seek corroboration before trusting AI-generated information and are consequently less vulnerable to misinformation. These findings align with prior work showing that active fact-checking yields more accurate veracity judgements (Brodsky et al., 2021; McGrew, 2020). The present work links critical thinking in AI use scores to behaviour in an AI-mediated fact-checking task that mirrors everyday engagement with generative AI tools. The significant relationships between the scale scores and both veracity judgement accuracy and verification behaviour provide evidence that the scale may tap into real-world engagement, indexing how users interpret and verify AI-generated information in practice.

Beyond global veracity judgement accuracy, our findings make a novel contribution by clarifying how individuals high in critical thinking in AI use respond to epistemic uncertainty in AI outputs. In Study 6, when AI outputs provided working, relevant source links, higher critical thinking in AI use scale scores predicted a small, non-significant increase in veracity judgement accuracy, and a marginal reduction in uncertainty. By contrast, when the AI chatbot did not provide relevant on-page evidence (either no source link, broken source link, or irrelevant source link), higher critical thinking in AI use scale scores predicted more accurate and definitive veracity judgements. Prior work has shown that users often rely on visible cues like source links when evaluating claims (Lucassen & Schraagen, 2013; Quelle & Bovet, 2024), but has not addressed how users behave when such cues are absent or misleading. Our results extend this literature by demonstrating that critical thinking in AI use predicts proactive verification behaviour even under constrained conditions, where on-page evidence is lacking or unreliable. This tendency to seek corroborating evidence beyond the AI chatbot, even when participants were told that no other AI tools or search engines were necessary, suggests that individuals higher in critical thinking in AI use are more likely to treat missing or unreliable evidence as a prompt for further scrutiny rather than as a basis for passive acceptance. Theoretically, these findings reveal that the core strength of critical thinking in AI use is not primarily about the effective use of readily available on-page evidence, but more about the active closing of evidence gaps by seeking corroboration beyond the AI chatbot in the absence of evidence, even when such verification is not required.

Critical thinking in AI use also extended to reflective outcomes. Higher critical thinking in AI use scale scores predicted both self-reported reflection depth and reflection depth in a writing task that asked participants to consider personal, social, and ethical implications of AI use. Participants higher in critical thinking in AI use not only engaged more deeply with issues such as fairness, accountability, and societal impact, and did so in ways that were detectable by LLM-coders. Consistent with the HSM (Chaiken, 1980; S. Chen & Chaiken, 1999), this pattern suggests that critical thinking in AI use is tied to a broader metacognitive orientation, in which users consciously weigh the downstream consequences of relying on AI systems, rather than focusing solely on immediate accuracy. These findings support recent arguments that contemporary critical thinking must encompass advanced epistemic competencies for evaluating the quality of information and navigating technology-mediated information environments, with attention to their broader societal implications (Muis et al., 2021; Pasquinelli et al., 2021).

## 8.2 Correlates of Critical Thinking in AI Use

Having established the structural and psychometric properties of the scale, we now consider what the nomological network reveals about the nature of the construct. Beyond psychometrics, we sought to clarify how critical thinking in AI use is embedded in a broader psychological profile. Our results reveal that high AI critical thinkers are likely to be open-minded, intellectually curious, socially agentic, frequent AI users who generally experience more positive affect yet may be able to better harness negative affect for vigilant scrutiny when evaluating AI-generated information. In a nomological network, we consistently found across Studies 3 and 4 that critical thinking in AI use was positively associated with openness to experience and extraversion, whereas associations with conscientiousness, agreeableness, and neuroticism were small or negligible. The link with openness to experience is in line with prior work showing that individuals higher in openness tend to report stronger critical thinking dispositions (Acevedo & Hess, 2022; Clifford et al., 2004; Ku & Ho, 2010). Novel to the domain of critical thinking in AI contexts is the finding that extraversion is also positively associated with critical thinking in AI use. This contrasts with traditional critical thinking research, where extraversion is not typically linked to critical thinking dispositions (Acevedo & Hess, 2022; Clifford et al., 2004). Unlike traditional critical thinking, critical thinking in AI use requires users to verify AI outputs with human experts and to actively seek out information from human experts about how AI systems work. Such behaviours are likely facilitated by an outgoing, socially engaged orientation encapsulated by a disposition of extraversion. Consistent with this interpretation, extraversion showed no association with the Reflection facet, which more closely resembles traditional critical thinking dispositions, but was positively related to the Verification and Motivation facets of critical thinking in AI use.

Besides personality traits, critical thinking in AI use was positively associated with trait positive affect, but not with trait negative affect. The positive affect finding is supported by emerging evidence that positive emotions can facilitate critical thinking performance (Lewine et al., 2015; Lun et al., 2023; Q. Zhang & Zhang, 2013), possibly through heightened arousal that sustains active engagement with evaluative tasks (Clore & Huntsinger, 2007). This pattern is also broadly consistent with emerging evidence that regular generative AI use is associated with higher well-being, suggesting that more adaptive and engaged forms of AI use may co-occur with more positive affective functioning (Lau & Hartanto, 2025). The non-significant positive association with negative affect partially departs from the prediction, grounded in feelings-as-information theory (Forgas, 2017, 2019; Schwarz, 2012), that negative mood should promote critical thinking in AI use through systematic processing. However, a facet-level analysis revealed a more nuanced pattern that may reconcile this apparent departure. Negative trait affect was positively associated with the reflection facet of critical thinking in AI use, consistent with the theoretical account that negative mood signals environmental threats and motivates vigilant, systematic scrutiny of downstream consequences (Forgas, 2019). By contrast, positive trait affect was associated with the Verification and Motivation facets, suggesting that an energised, approach-oriented affective disposition may facilitate active engagement with AI outputs and sustained interest in understanding AI systems. This differential pattern represents a novel empirical contribution, suggesting that the HSM prediction linking negative affect to systematic processing holds selectively for the deliberative, ethically oriented dimension of critical thinking in AI use, rather than the construct as a whole. More broadly, these findings extend previous accounts that treat affect's influence on critical thinking as monolithic (Lun et al., 2023), revealing that distinct affective dispositions relate to qualitatively different dimensions of critical thinking in AI use. In doing so, the present findings offer initial evidence for mapping positive and negative trait affect onto the Verification, Motivation, and Reflection dimensions, suggesting that affective dispositions may be differentially associated with evaluative, epistemic, and ethical dimensions of critical thinking in AI use.

Within the nomological network, we also found that individuals higher on critical thinking in AI use tended to report more frequent AI use. Consistent with prior work, users commonly use generative AI tools for cross-checking claims, asking follow-up questions, and iteratively re-prompting to verify the source and content of AI outputs (Quelle & Bovet, 2024). Those who use AI more frequently may have greater opportunities to develop verification practices and a more refined understanding of model limitations, but these benefits are likely contingent on how users engage with AI outputs, specifically, whether such use is accompanied by active checking, reflective oversight, and attention to possible error (Ng et al., 2021; Park, 2025; Promma et al., 2025; Stevic et al., 2025). As users become more familiar with AI tools, interactions become less effortful, freeing cognitive resources that can be allocated to multitasking and to running verification checks in parallel with information search (Haith & Krakauer, 2018; Ruthruff et al., 2001). This association adds novel evidence by clarifying how usage frequency relates to critical thinking in AI use. Whereas previous research suggests that frequent (Gerlich, 2025a) and problematic use (Hou et al., 2025) of AI tools can undermine traditional critical thinking, our findings indicate that frequent AI users may develop higher levels of AI-specific critical thinking.

**8.3 Theoretical and Practical Implications**

Beyond its empirical contributions, the present research has implications for both research practice and real-world AI use. Firstly, the present work advances theory by articulating a construct definition of critical thinking in AI use that integrates verification, epistemic motivation, and ethical-societal reflection into a unified framework grounded in dual-process theory and tailored to the distinctive epistemic challenges posed by generative AI, thereby extending domain-general critical thinking constructs (Ennis, 1985; Facione, 1990b; Sosu, 2013) to the AI domain. Secondly, it provides empirical insight by mapping the nomological network of critical thinking in AI use, identifying its personality and affective correlates, and demonstrating its criterion validity through a novel, ecologically grounded

GPT-powered AI chatbot fact-checking paradigm, which offers evidence not available in prior work. These contributions differ in important ways from prior work. Existing measures of critical thinking (Cacioppo et al., 1984; Sosu, 2013), were not designed to capture the distinctive demands of evaluating opaque, probabilistic AI outputs. Similarly, while AI literacy frameworks (Long & Magerko, 2020; Pinski & Benlian, 2024) assess knowledge about AI systems, they do not directly measure the evaluative and reflective practices that users deploy when interacting with AI-generated content. The critical thinking in AI use scale addresses this gap by providing the first validated, domain-specific operationalisation of critical thinking tailored to generative AI contexts. Future research could examine how critical thinking in AI use relates to generative AI literacy and performance on AI knowledge-and-skills assessments, such as the GLAT (Jin et al., 2025).

Lastly, the present work makes practical and methodological contributions. The validated scale, with its replicated hierarchical structure, temporal stability, and sex invariance, provides a psychometrically robust instrument for assessing individual differences in critical thinking in AI use, supporting theory testing and longitudinal and cross-group research. The scale may also have utility as a screening instrument for identifying individuals who could benefit from targeted support in developing critical thinking when engaging with AI outputs, though its predictive validity in applied settings has yet to be established. In educational settings, for example, it could be used to identify students who may require additional support in verifying and reflecting on AI outputs. This application is particularly relevant given growing concerns that, if sustained patterns of uncritical AI reliance are confirmed in longitudinal research, they may have implications for the maintenance of independent analytical capabilities over time (Gerlich, 2025a; Hou et al., 2025; Kosmyna et al., 2025), though this remains an empirical question. Whereas prior research has primarily examined the effects of AI use on cognitive outcomes at the group level (e.g., Gerlich, 2025a;

Kosmyna et al., 2025), the present work is among the first to operationalise and validate individual differences in critical thinking specific to generative AI contexts and to demonstrate their predictive relevance for possible downstream verification behaviour and judgement accuracy in an AI-mediated task. Complementing the scale, the adaptable chatbot-based task paradigm offers an ecologically valid method that can be used in future research to examine misinformation detection, source evaluation, verification behaviour, and related forms of critical thinking in realistic AI-mediated environments.

### 8.4 Limitations and Future Directions

While the present research provides a foundational understanding of critical thinking in AI use, it has a few limitations that should be acknowledged and that open avenues for future research. First, our data rely primarily on English-speaking online and university samples. Competencies that relate to critical thinking in AI use, such as AI literacy (Chung & Wihbey, 2024; Mansoor et al., 2024) and misinformation susceptibility (M. Chan et al., 2025; Kyrychenko et al., 2025; Maertens et al., 2024; Roozenbeek et al., 2020) have demonstrated cross-cultural variations. Future work should test critical thinking in AI use across cultures, languages, and organisational settings, examining measurement invariance and group differences to strengthen the generalisability of the scale. Second, the present studies did not include direct measures of (generative) AI literacy. As a result, we were unable to examine how critical thinking in AI use relates to people's declarative and procedural knowledge about AI systems, or whether the scale predicts outcomes over and above AI literacy. Given frameworks that conceptualise AI literacy as a prerequisite for competent and responsible AI use (Long & Magerko, 2020; Mills et al., 2024), future research should include validated AI literacy measures to test their overlap, discriminant validity, and incremental predictive value for verification behaviour and real-world AI outcomes. Third, we did not assess participants' prior knowledge or expertise in the substantive domains of the veracity judgement tasks.

Domain knowledge may shape how individuals evaluate AI outputs, and it may interact with AI outputs (Dikmen & Burns, 2022), for example, by enabling high-critical thinking in AI use individuals to detect subtle inaccuracies in AI-generated content. Future studies should therefore measure and experimentally manipulate topic knowledge to clarify how critical thinking in AI use combines with domain-specific expertise in guiding verification behaviour.

Additionally, the present studies were not designed to directly test the full set of causal pathways implied by the HSM-based theoretical framework (Chaiken, 1980; S. Chen & Chaiken, 1999), particularly the role of authority and fluency cues in generative AI systems as antecedents of heuristic processing that suppresses critical thinking in AI use (Rheu et al., 2021; Tutuncuoglu, 2025). Our primary aim in this initial investigation was to develop and validate a psychometrically sound measure of the construct. Nevertheless, guided by the proposed framework, the present research examined both a plausible antecedent and a plausible consequence. As a possible antecedent, positive attitudes toward AI, which encompass users' trust toward AI systems (Klingbeil et al., 2024; J. D. Lee & See, 2004), were negatively associated with critical thinking in AI use in Study 3, consistent with the idea that trust-related heuristic cues may reduce systematic evaluation (Sun et al., 2025). However, this association was not significant in Study 4, suggesting that the relationship may be conditional on sample or contextual factors.

As a possible consequence, the present research examined veracity judgement accuracy in Study 6, where participants higher in critical thinking in AI use were better able to distinguish true from false claims presented by the AI chatbot, consistent with the prediction that systematic processing should reduce automation bias (Alexander et al., 2018). Although the broader design of the present research is cross-sectional, this limitation is less concerning for Study 6 specifically, because the dispositional measure of critical thinking in AI use was assessed prior to the fact-checking task and the task outcome (veracity judgement accuracy) observed subsequently within the same session. A reverse-causal interpretation, that performing well on a novel fact-checking task caused participants to endorse higher levels of habitual critical thinking in AI use, is not plausible given the trait-like nature of the construct. These findings therefore offer preliminary support for the proposed framework while underscoring the need for future research to test these pathways more directly, for example through experimental manipulation of output cues such as fluency, source citations, and confidence indicators (Quelle & Bovet, 2024; Rheu et al., 2021) and mediation designs that model the sequence from heuristic cues through processing mode to verification behaviour and judgement accuracy. Although the present results indicate that verification is adaptive, a recent study suggests that sustained source-checking, error monitoring, and prompt refinement may be cognitively effortful and contribute to AI fatigue over time, which may in turn reduce AI use (Lau, Kasturiratna, et al., 2026), and possibly the persistence of verification behaviour. Longitudinal and experience-sampling research is needed to clarify the antecedents and consequences of critical thinking in AI use across repeated human–AI interactions.

## 9. Conclusion

In conclusion, this work provides a comprehensive first account of critical thinking in AI use as a measurable construct that captures stable individual differences and is associated with meaningful differences in how people engage with generative AI. Across six studies, the present research introduced a theoretically grounded and empirically validated instrument that captures how users verify the source and content of AI outputs, seek to understand model behaviour and limits, and reflect responsibly on personal, social, and ethical implications of AI use. The critical thinking in AI use scale exhibited a clear three-factor structure with a higher-order factor, strong internal consistency, temporal stability, construct validity, and measurement invariance across sex. Critical thinking in AI use was systematically patterned by frequency of AI use, positive affect, and personality traits such as openness and extraversion. Importantly, Study 6 extended the evidence beyond self-report by linking critical thinking in AI use to task-based outcomes, specifically veracity judgement accuracy in a generative-AI fact-checking task and LLM-scored reflection depth in a writing task. These task-based indices were complemented by self-reported verification frequency and verification method diversity, which, while subject to shared method variance with the scale, converged with the task-based findings. Critical thinking in AI use can support calibrated, responsible AI use by encouraging engagement with AI while maintaining human oversight of its outputs. Practically, the critical thinking in AI use scale provides a rigorous diagnostic for evaluating human oversight of AI outputs, across time and between demographic groups, in AI-mediated organisations and educational institutions.

## CRediT authorship contribution statement

**Gabriel R. Lau:** Writing – review & editing, Writing – original draft, Visualization, Software, Project administration, Methodology, Investigation, Formal analysis, Data curation, Conceptualization. **Wei Yan Low:** Writing – review & editing, Validation, Investigation. **Louis Tay:** Writing – review & editing, Validation. **Ysabel A. Guevarra:** Writing – review & editing, Project administration, Data curation. **Dragan Gašević:** Writing – review & editing, Validation. **Andree Hartanto:** Writing – review & editing, Supervision, Resources, Methodology, Investigation, Funding acquisition, Conceptualization.

## Declaration of competing interest

The authors declare that they have no known competing financial interests or personal relationships that could have appeared to influence the work reported in this paper.

## Data availability

Data will be made available on request.

## Funding

This research was supported by grants awarded to Andree Hartanto by Singapore Management University through research grants from the Ministry of Education Academic Research Fund Tier 1 (25-SOSS-SMU-006) and Lee Kong Chian Fund for Research Excellence.

## Acknowledgements

We thank Charlotte, Shermaine, Claire, Ayden, Andre, Mary, Gabrielle, Elisabeth, Isaac, Jingwen, Paye Shin, Emma Jane, Adeliese, Harshitha, Jun Heng, Kelly, Arundati, Gilda, Seow Min, Joyce, and Claire for contributing to different parts of this project, including content validation, manuscript proofreading, and data cleaning.

**References**

Acevedo, E. C., & Hess, C. (2022). The link between critical thinking and personality: Individual differences in a concern for truth. *Modern Psychological Studies*, *27*(1), 9.

Alexander, V., Blinder, C., & Zak, P. J. (2018). Why trust an algorithm? Performance, cognition, and neurophysiology. *Computers in Human Behavior*, *89*, 279–288. https://doi.org/10.1016/j.chb.2018.07.026

Almulla, M. A. (2023). Constructivism learning theory: A paradigm for students' critical thinking, creativity, and problem solving to affect academic performance in higher education. *Cogent Education*, *10*(1), 2172929. https://doi.org/10.1080/2331186X.2023.2172929

Amit, A., & Sagiv, L. (2013). The role of epistemic motivation in individuals' response to decision complexity. *Organizational Behavior and Human Decision Processes*, *121*(1), 104–117. https://doi.org/10.1016/j.obhdp.2013.01.003

Ananiadou, K., & Claro, M. (2009). 21st century skills and competences for new millennium learners in OECD countries. *OECD Education Working Papers*, *2009*. https://doi.org/10.1787/218525261154

Anderson, J. C., & Gerbing, D. W. (1991). Predicting the performance of measures in a confirmatory factor analysis with a pretest assessment of their substantive validities. *Journal of Applied Psychology*, *76*(5), 732–740. https://doi.org/10.1037/0021-9010.76.5.732

Bastani, H., Bastani, O., Sungu, A., Ge, H., Kabakcı, Ö., & Mariman, R. (2025). Generative AI without guardrails can harm learning: Evidence from high school mathematics. *Proceedings of the National Academy of Sciences*, *122*(26), e2422633122. https://doi.org/10.1073/pnas.2422633122

Bećirović, S., Polz, E., & Tinkel, I. (2025). Exploring students' AI literacy and its effects on their AI output quality, self-efficacy, and academic performance. *Smart Learning Environments*, *12*(1), 29. https://doi.org/10.1186/s40561-025-00384-3

Bless, H., & Fiedler, K. (2006). Mood and the regulation of information processing and behavior. In J. P. Forgas (Ed.), *Affect in social thinking and behavior* (pp. 65–84). Psychology Press.

Boateng, G. O., Neilands, T. B., Frongillo, E. A., Melgar-Quiñonez, H. R., & Young, S. L. (2018). Best practices for developing and validating scales for health, social, and behavioral research: A primer. *Frontiers in Public Health*, *6*, 149. https://doi.org/10.3389/fpubh.2018.00149

Boissin, E., Caparos, S., Raoelison, M., & De Neys, W. (2021). From bias to sound intuiting: Boosting correct intuitive reasoning. *Cognition*, *211*, 104645. https://doi.org/10.1016/j.cognition.2021.104645

Bonett, D. G. (2002). Sample size requirements for estimating intraclass correlations with desired precision. *Statistics in Medicine*, *21*(9), 1331–1335. https://doi.org/10.1002/sim.1108

Boonsathirakul, J., & Kerdsomboon, C. (2021). The investigation of critical thinking disposition among Kasetsart university students. *Higher Education Studies*, *11*(2), 224. https://doi.org/10.5539/hes.v11n2p224

Bouteraa, M., Bin-Nashwan, S. A., Al-Daihani, M., Dirie, K. A., Benlahcene, A., Sadallah, M., Zaki, H. O., Lada, S., Ansar, R., Fook, L. M., & Chekima, B. (2024). Understanding the diffusion of AI-generative (ChatGPT) in higher education: Does students' integrity matter? *Computers in Human Behavior Reports*, *14*, 100402. https://doi.org/10.1016/j.chbr.2024.100402

Brodsky, J. E., Brooks, P. J., Scimeca, D., Todorova, R., Galati, P., Batson, M., Grosso, R., Matthews, M., Miller, V., & Caulfield, M. (2021). Improving college students' fact-checking strategies through lateral reading instruction in a general education civics course. *Cognitive Research: Principles and Implications*, *6*(1), 23. https://doi.org/10.1186/s41235-021-00291-4

Bruton, A., Conway, J. H., & Holgate, S. T. (2000). Reliability: What is it, and how is it measured? *Physiotherapy*, *86*(2), 94–99. https://doi.org/10.1016/S0031-9406(05)61211-4

Brynjolfsson, E., Li, D., & Raymond, L. (2025). Generative AI at work. *The Quarterly Journal of Economics*, *140*(2), 889–942. https://doi.org/10.1093/qje/qjae044

Cabrera-Nguyen, P. (2010). Author guidelines for reporting scale development and validation results in the Journal of the Society for Social Work and Research. *Journal of the Society for Social Work and Research*, *1*(2), 99–103. https://doi.org/10.5243/jsswr.2010.8

Cacioppo, J. T., Petty, R. E., & Feng Kao, C. (1984). The efficient assessment of need for cognition. *Journal of Personality Assessment*, *48*(3), 306–307. https://doi.org/10.1207/s15327752jpa4803_13

Cai, C., Zhao, X., Liu, H., Jiang, Z., Zhang, T., Wu, Z., Hwang, J.-N., & Li, L. (2025). *The role of deductive and inductive reasoning in large language models* (arXiv:2410.02892). arXiv. https://doi.org/10.48550/arXiv.2410.02892

Cao, C., & Liang, X. (2022). The impact of model size on the sensitivity of fit measures in measurement invariance testing. *Structural Equation Modeling: A Multidisciplinary Journal*, *29*(5), 744–754. https://doi.org/10.1080/10705511.2022.2056893

Chaiken, S. (1980). Heuristic versus systematic information processing and the use of source versus message cues in persuasion. *Journal of Personality and Social Psychology*, *39*(5), 752–766. https://doi.org/10.1037/0022-3514.39.5.752

Chan, C. K. Y., & Hu, W. (2023). Students' voices on generative AI: Perceptions, benefits,

and challenges in higher education. *International Journal of Educational Technology in Higher Education*, *20*(1), 43. https://doi.org/10.1186/s41239-023-00411-8

Chan, M., Yi, J., Vaccari, C., & Yamamoto, M. (2025). A cross-national examination of the effects of accuracy nudges and content veracity labels on belief in and sharing of misleading news. *Journal of Computer-Mediated Communication*, *30*(4), zmaf009. https://doi.org/10.1093/jcmc/zmaf009

Chen, F. F. (2007). Sensitivity of goodness of fit indexes to lack of measurement invariance. *Structural Equation Modeling: A Multidisciplinary Journal*, *14*(3), 464–504. https://doi.org/10.1080/10705510701301834

Chen, S., & Chaiken, S. (1999). The heuristic-systematic model in its broader context. In S. Chaiken & Y. Trope (Eds.), *Dual-process theories in social psychology* (pp. 73–96). The Guilford Press.

Cheung, G. W., & Rensvold, R. B. (2002). Evaluating goodness-of-fit indexes for testing measurement invariance. *Structural Equation Modeling: A Multidisciplinary Journal*, *9*(2), 233–255. https://doi.org/10.1207/S15328007SEM0902_5

Chew, R., Bollenbacher, J., Wenger, M., Speer, J., & Kim, A. (2023). *LLM-assisted content analysis: Using large language models to support deductive coding* (arXiv:2306.14924). arXiv. https://doi.org/10.48550/arXiv.2306.14924

Chu-Ke, C., & Dong, Y. (2024). Misinformation and literacies in the era of generative artificial intelligence: A brief overview and a call for future research. *Emergency Medicine*, *2*(1), 70–85. https://doi.org/10.1177/27523543241240285

Chung, M., & Wihbey, J. (2024). The algorithmic knowledge gap within and between countries: Implications for combatting misinformation. *Harvard Kennedy School Misinformation Review*. https://doi.org/10.37016/mr-2020-155

Clifford, J. S., Boufal, M. M., & Kurtz, J. E. (2004). Personality traits and critical thinking

skills in college students: Empirical tests of a two-factor theory. *Assessment*, *11*(2), 169–176. https://doi.org/10.1177/1073191104263250

Clore, G. L., & Huntsinger, J. R. (2007). How emotions inform judgment and regulate thought. *Trends in Cognitive Sciences*, *11*(9), 393–399. https://doi.org/10.1016/j.tics.2007.08.005

Colquitt, J. A., Sabey, T. B., Rodell, J. B., & Hill, E. T. (2019). Content validation guidelines: Evaluation criteria for definitional correspondence and definitional distinctiveness. *Journal of Applied Psychology*, *104*(10), 1243–1265. https://doi.org/10.1037/apl0000406

Costello, A. B., & Osborne, J. (2005). Best practices in exploratory factor analysis: Four recommendations for getting the most from your analysis. *Practical Assessment, Research, and Evaluation*, *10*(1). https://doi.org/10.7275/jyj1-4868

De Dreu, C. K. W., Nijstad, B. A., & van Knippenberg, D. (2008). Motivated information processing in group judgment and decision making. *Personality and Social Psychology Review*, *12*(1), 22–49. https://doi.org/10.1177/1088868307304092

DeVellis, R. F., & Thorpe, C. T. (2021). *Scale development: Theory and applications*. SAGE Publications.

Dikmen, M., & Burns, C. (2022). The effects of domain knowledge on trust in explainable AI and task performance: A case of peer-to-peer lending. *International Journal of Human-Computer Studies*, *162*, 102792. https://doi.org/10.1016/j.ijhcs.2022.102792

Dolan, R. J. (2002). Emotion, cognition, and behavior. *Science*, *298*(5596), 1191–1194. https://doi.org/10.1126/science.1076358

Doshi, A. R., Bell, J. J., Mirzayev, E., & Vanneste, B. S. (2025). Generative artificial intelligence and evaluating strategic decisions. *Strategic Management Journal*, *46*(3), 583–610. https://doi.org/10.1002/smj.3677

Doshi, A. R., & Hauser, O. P. (2024). Generative AI enhances individual creativity but reduces the collective diversity of novel content. *Science Advances*, *10*(28), eadn5290. https://doi.org/10.1126/sciadv.adn5290

Duff, K. (2014). One-week practice effects in older adults: Tools for assessing cognitive change. *The Clinical Neuropsychologist*, *28*(5), 714–725. https://doi.org/10.1080/13854046.2014.920923

Engin, A. (2021). The cognitive ability and working memory framework: Interpreting cognitive reflection test results in the domain of the cognitive experiential theory. *Central European Journal of Operations Research*, *29*(1), 227–245. https://doi.org/10.1007/s10100-020-00721-6

Ennis, R. H. (1985). A logical basis for measuring critical thinking skills. *Educational Leadership*, *43*(2), 44–48.

Ennis, R. H. (2011). The nature of critical thinking: An outline of critical thinking dispositions and abilities. In *Presentation Sixth International Conference on Thinking*. MIT, Cambridge, MA. https://education.illinois.edu/docs/default-source/faculty-documents/robert-ennis/thenatureofcriticalthinking_51711_000.pdf

Fabio, R. A., Antonietti, A., Iannello, P., & Suriano, R. (2025). Development and psychometric properties of the critical thinking attitude scale in Italian college students. *Frontiers in Psychology*, *16*. https://doi.org/10.3389/fpsyg.2025.1599920

Facione, P. A. (1990a). *Critical thinking: A statement of expert consensus for purposes of educational assessment and instruction.* The California Academic Press.

Facione, P. A. (1990b). *The California Critical Thinking Skills Test—College Level. Technical Report No. 3. Gender, Ethnicity, Major, CT Self-Esteem, and the CCTST*. https://eric.ed.gov/?id=ED326584

Fan, Y., Tang, L., Le, H., Shen, K., Tan, S., Zhao, Y., Shen, Y., Li, X., & Gašević, D. (2025).

Beware of metacognitive laziness: Effects of generative artificial intelligence on learning motivation, processes, and performance. *British Journal of Educational Technology*, *56*(2), 489–530. https://doi.org/10.1111/bjet.13544

Fernandes, D., Villa, S., Nicholls, S., Haavisto, O., Buschek, D., Schmidt, A., Kosch, T., Shen, C., & Welsch, R. (2026). AI makes you smarter but none the wiser: The disconnect between performance and metacognition. *Computers in Human Behavior*, *175*, 108779. https://doi.org/10.1016/j.chb.2025.108779

Forgas, J. P. (2017). Mood effects on cognition: Affective influences on the content and process of information processing and behavior. *Emotions and Affect in Human Factors and Human-Computer Interaction*, 89–122. https://doi.org/10.1016/B978-0-12-801851-4.00003-3

Forgas, J. P. (2019). On the role of affect in gullibility: Can positive mood increase, and negative mood reduce credulity? In J. P. Forgas & R. Baumeister (Eds.), *The social psychology of gullibility: Conspiracy theories, fake news and irrational beliefs* (1st ed., pp. 179–197). Routledge.

Fornell, C., & Larcker, D. F. (1981). Evaluating structural equation models with unobservable variables and measurement error. *Journal of Marketing Research*, *18*(1), 39–50. https://doi.org/10.2307/3151312

Frederick, S. (2005). Cognitive reflection and decision making. *Journal of Economic Perspectives*, *19*(4), 25–42. https://doi.org/10.1257/089533005775196732

Funder, D. C., & Ozer, D. J. (2019). Evaluating effect size in psychological research: Sense and nonsense. *Advances in Methods and Practices in Psychological Science*, *2*(2), 156–168. https://doi.org/10.1177/2515245919847202

Gerlich, M. (2025a). AI tools in society: Impacts on cognitive offloading and the future of critical thinking. *Societies*, *15*(1), 6. https://doi.org/10.3390/soc15010006

Gerlich, M. (2025b). From offloading to engagement: An experimental study on structured prompting and critical reasoning with generative AI. *Data*, *10*(11), 172. https://doi.org/10.3390/data10110172

Gerlich, M. (2025c). Outsourcing judgment: Hidden anxieties and the rise of cognitive offloading in the age of AI. In P. Gazzola & G. Dominici (Eds.), *Technology and society—Boon or bane?* (pp. 44–58). Springer Nature Switzerland. https://doi.org/10.1007/978-3-032-07163-7_3

Goh, A. Y. H., Hartanto, A., & Majeed, N. M. (2025). Generative artificial intelligence dependency: Scale development, validation, and its motivational, behavioral, and psychological correlates. *Computers in Human Behavior Reports*, *20*, 100845. https://doi.org/10.1016/j.chbr.2025.100845

Grinschgl, S., & Neubauer, A. C. (2022). Supporting cognition with modern technology: Distributed cognition today and in an AI-enhanced future. *Frontiers in Artificial Intelligence*, *5*, 908261. https://doi.org/10.3389/frai.2022.908261

Gruenhagen, J. H., Sinclair, P. M., Carroll, J.-A., Baker, P. R. A., Wilson, A., & Demant, D. (2024). The rapid rise of generative AI and its implications for academic integrity: Students' perceptions and use of chatbots for assistance with assessments. *Computers and Education: Artificial Intelligence*, *7*, 100273. https://doi.org/10.1016/j.caeai.2024.100273

Gsenger, R., & Strle, T. (2021). Trust, automation bias and aversion: Algorithmic decision-making in the context of credit scoring. *Interdisciplinary Description of Complex Systems*, *19*(4), 542–560. https://doi.org/10.7906/indecs.19.4.7

Hagerty, A., & Rubinov, I. (2019). *Global AI ethics: A review of the social impacts and ethical implications of artificial intelligence* (arXiv:1907.07892). arXiv. https://doi.org/10.48550/arXiv.1907.07892

Haigh, M. (2016). Has the standard cognitive reflection test become a victim of its own success? *Advances in Cognitive Psychology*, *12*(3), 145–149. https://doi.org/10.5709/acp-0193-5

Hair, J. F., Black, W. C., Babin, B. J., & Anderson, R. E. (2019). *Multivariate data analysis* (8th ed.). Cengage.

Haith, A. M., & Krakauer, J. W. (2018). The multiple effects of practice: Skill, habit and reduced cognitive load. *Current Opinion in Behavioral Sciences*, *20*, 196–201. https://doi.org/10.1016/j.cobeha.2018.01.015

Halpern, D. F. (2014). *Thought and knowledge: An introduction to critical thinking* (5th ed.). Psychology Press. https://doi.org/10.4324/9781315885278

Hayton, J. C., Allen, D. G., & Scarpello, V. (2004). Factor retention decisions in exploratory factor analysis: A tutorial on parallel analysis. *Organizational Research Methods*, *7*(2), 191–205. https://doi.org/10.1177/1094428104263675

Hemmer, P., Westphal, M., Schemmer, M., Vetter, S., Vössing, M., & Satzger, G. (2023). Human–AI collaboration: The effect of AI delegation on human task performance and task satisfaction. *Proceedings of the 28th International Conference on Intelligent User Interfaces*, 453–463. https://doi.org/10.1145/3581641.3584052

Ho, J. Q. H., Hartanto, A., Koh, A., & Majeed, N. M. (2025). Gender biases within artificial intelligence and ChatGPT: Evidence, sources of biases, and solutions. *Computers in Human Behavior: Artificial Humans*, *4*, 100145. https://doi.org/10.1016/j.chbah.2025.100145

Hong, H., Vate-U-Lan, P., & Viriyavejakul, C. (2025). Cognitive offload instruction with generative AI: A quasi-experimental study on critical thinking gains in English writing. *Forum for Linguistic Studies*, *7*. https://doi.org/10.30564/fls.v7i7.10072

Hopwood, C. J., & Donnellan, M. B. (2010). How should the internal structure of personality inventories be evaluated? *Personality and Social Psychology Review*, *14*(3), 332–346. https://doi.org/10.1177/1088868310361240

Horn, J. L. (1965). A rationale and test for the number of factors in factor analysis. *Psychometrika*, *30*(2), 179–185. https://doi.org/10.1007/BF02289447

Hou, C., Zhu, G., & Sudarshan, V. (2025). The role of critical thinking on undergraduates' reliance behaviours on generative AI in problem-solving. *British Journal of Educational Technology*, *56*(5), 1919–1941. https://doi.org/10.1111/bjet.13613

Hu, M., Lau, G. R., Goh, A. Y. H., Cox, S. R., Tay, L., & Hartanto, A. (2026). *Using Customisable GPT Chatbots in Psychology Research: A Practical Tutorial* (Kf96p_v1). PsyArXiv. https://osf.io/preprints/psyarxiv/kf96p_v1/

Huang, L., Yu, W., Ma, W., Zhong, W., Feng, Z., Wang, H., Chen, Q., Peng, W., Feng, X., Qin, B., & Liu, T. (2025). A survey on hallucination in large language models: Principles, taxonomy, challenges, and open questions. *ACM Transactions on Information Systems*, *43*(2), 42:1-55. https://doi.org/10.1145/3703155

Humlum, A., & Vestergaard, E. (2025). The unequal adoption of ChatGPT exacerbates existing inequalities among workers. *Proceedings of the National Academy of Sciences*, *122*(1), e2414972121. https://doi.org/10.1073/pnas.2414972121

Ibrar, W., Mahmood, D., Al-Shamayleh, A. S., Ahmed, G., Alharthi, S. Z., & Akhunzada, A. (2025). Generative AI: A double-edged sword in the cyber threat landscape. *Artificial Intelligence Review*, *58*(9), 285. https://doi.org/10.1007/s10462-025-11285-9

Jin, Y., Martinez-Maldonado, R., Gašević, D., & Yan, L. (2025). GLAT: The generative AI literacy assessment test. *Computers and Education: Artificial Intelligence*, *9*, 100436. https://doi.org/10.1016/j.caeai.2025.100436

Jorgensen, T. D., Pornprasertmanit, S., Schoemann, A. M., & Rosseel, Y. (2025). *semTools: Useful tools for structural equation modeling* (Version 0.5-7) [Computer software].

The R Foundation. https://CRAN.R-project.org/package=semTools

Kahneman, D. (2011). *Thinking, fast and slow.* Farrar, Straus and Giroux.

Karunaratne, T., & Adesina, A. (2023). Is it the new Google: Impact of ChatGPT on students' information search habits. *European Conference on E-Learning*, *22*(1), 147–155. https://doi.org/10.34190/ecel.22.1.1831

Kasturiratna, K. T. A. S., & Hartanto, A. (2025). Attachment to Artificial Intelligence: Development of the AI Attachment Scale, Construct Validation, and the Psychological Mechanisms of Human–AI Attachment. *Computers in Human Behavior Reports*, 100912. https://doi.org/10.1016/j.chbr.2025.100912

King, P. M., & Kitchener, K. S. (1994). *Developing Reflective Judgment: Understanding and Promoting Intellectual Growth and Critical Thinking in Adolescents and Adults. Jossey-Bass Higher and Adult Education Series and Jossey-Bass Social and Behavioral Science Series*. Jossey-Bass, 350 Sansome Street, San Francisco, CA 94104-1310 ($32.

King, P. M., & Kitchener, K. S. (2004). Reflective judgment: Theory and research on the development of epistemic assumptions through adulthood. *Educational Psychologist*, *39*(1), 5–18. https://doi.org/10.1207/s15326985ep3901_2

Kitchener, K. S., & King, P. M. (1981). Reflective judgment: Concepts of justification and their relationship to age and education. *Journal of Applied Developmental Psychology*, *2*(2), 89–116. https://doi.org/10.1016/0193-3973(81)90032-0

Kline, P. (2015). *A handbook of test construction: Introduction to psychometric design*. Routledge. https://doi.org/10.4324/9781315695990

Klingbeil, A., Grützner, C., & Schreck, P. (2024). Trust and reliance on AI — An experimental study on the extent and costs of overreliance on AI. *Computers in Human Behavior*, *160*, 108352. https://doi.org/10.1016/j.chb.2024.108352

Knowles, J. E., & Frederick, C. (2025). *merTools: Tools for analyzing mixed effect regression models* (Version 0.6.3) [Computer software]. The R Foundation. https://github.com/jknowles/mertools

Koo, T. K., & Li, M. Y. (2016). A guideline of selecting and reporting intraclass correlation coefficients for reliability research. *Journal of Chiropractic Medicine*, *15*(2), 155–163. https://doi.org/10.1016/j.jcm.2016.02.012

Kosmyna, N., Hauptmann, E., Yuan, Y. T., Situ, J., Liao, X.-H., Beresnitzky, A. V., Braunstein, I., & Maes, P. (2025). *Your brain on ChatGPT: Accumulation of cognitive debt when using an AI assistant for essay writing tasks* (arXiv:2506.08872). arXiv. https://doi.org/10.48550/arXiv.2506.08872

Ku, K. Y. L., & Ho, I. T. (2010). Dispositional factors predicting Chinese students' critical thinking performance. *Personality and Individual Differences*, *48*(1), 54–58. https://doi.org/10.1016/j.paid.2009.08.015

Kyrychenko, Y., Koo, H. J., Maertens, R., Roozenbeek, J., van der Linden, S., & Götz, F. M. (2025). Profiling misinformation susceptibility. *Personality and Individual Differences*, *241*, 113177. https://doi.org/10.1016/j.paid.2025.113177

Landerholm, A. (2025). AI in qualitative health research appraisal: Comparative study. *JMIR Formative Research*, *9*, e72815. https://doi.org/10.2196/72815

Lanz, L., Thielmann, I., & Gerpott, F. H. (2022). Are social desirability scales desirable? A meta-analytic test of the validity of social desirability scales in the context of prosocial behavior. *Journal of Personality*, *90*(2), 203–221. https://doi.org/10.1111/jopy.12662

Lau, G. R., & Hartanto, A. (2025). *Generative AI use but not social media and smartphone is associated with higher well-being* (SSRN Scholarly Paper No. 5893625). Social Science Research Network. https://doi.org/10.2139/ssrn.5893625

Lau, G. R., Kasturiratna, K. T. A. S., Goh, A. Y. H., Tong, E. M. W., & Hartanto, A. (2026). *AI fatigue in human–AI interaction: Conceptual framework, scale development and validation, and associations with AI engagement*. PsyArXiv. https://doi.org/10.31234/osf.io/kp7zs_v1

Lau, G. R., Low, W. Y., Koh, S. M., & Hartanto, A. (2025). *Evaluating AI alignment in eleven LLMs through output-based analysis and human benchmarking* (arXiv:2506.12617). arXiv. https://doi.org/10.48550/arXiv.2506.12617

Lau, G. R., Tong, E. M. W., & Hartanto, A. (2026). When humans obey AI: Development of the Obedience to Artificial Intelligence Scale and a dual-pathway motivational framework of obedience to AI. *PsyArXiv*. https://doi.org/10.31234/osf.io/a63eq_v1

Lebovitz, S., Lifshitz-Assaf, H., & Levina, N. (2022). To engage or not to engage with AI for critical judgments: How professionals deal with opacity when using AI for medical diagnosis. *Organization Science*, *33*(1), 126–148. https://doi.org/10.1287/orsc.2021.1549

Lee, B. C., & Chung, J. (2024). An empirical investigation of the impact of ChatGPT on creativity. *Nature Human Behaviour*, *8*(10), 1906–1914. https://doi.org/10.1038/s41562-024-01953-1

Lee, C., Kim, J., Lim, J. S., & Shin, D. (2025). Generative AI risks and resilience: How users adapt to hallucination and privacy challenges. *Telematics and Informatics Reports*, *19*, 100221. https://doi.org/10.1016/j.teler.2025.100221

Lee, D., Arnold, M., Srivastava, A., Plastow, K., Strelan, P., Ploeckl, F., Lekkas, D., & Palmer, E. (2024). The impact of generative AI on higher education learning and teaching: A study of educators' perspectives. *Computers and Education: Artificial Intelligence*, *6*, 100221. https://doi.org/10.1016/j.caeai.2024.100221

Lee, J. D., & See, K. A. (2004). Trust in automation: Designing for appropriate reliance.

*Human Factors*, *46*(1), 50–80. https://doi.org/10.1518/hfes.46.1.50_30392

Lewine, R., Sommers, A., Waford, R., & Robertson, C. (2015). Setting the mood for critical thinking in the classroom. *International Journal for the Scholarship of Teaching and Learning*, *9*(2). https://doi.org/10.20429/ijsotl.2015.090205

Liao, Q. V., & Vaughan, J. W. (2023). *AI transparency in the age of LLMs: A human-centered research roadmap* (arXiv:2306.01941). arXiv. https://doi.org/10.48550/arXiv.2306.01941

Lilienfeld, S. O., Basterfield, C., Bowes, S. M., & Costello, T. H. (2020). Nobelists gone wild: Case studies in the domain specificity of critical thinking. In R. J. Sternberg & D. F. Halpern (Eds.), *Critical Thinking in Psychology* (2nd ed., pp. 10–38). Cambridge University Press. https://doi.org/10.1017/9781108684354.003

Long, D., & Magerko, B. (2020). What is AI literacy? Competencies and design considerations. *Proceedings of the 2020 CHI Conference on Human Factors in Computing Systems, CHI '20*, 1–16. https://doi.org/10.1145/3313831.3376727

Lucassen, T., & Schraagen, J. M. (2013). The influence of source cues and topic familiarity on credibility evaluation. *Computers in Human Behavior*, *29*(4), 1387–1392. https://doi.org/10.1016/j.chb.2013.01.036

Lun, V. M. C., Yeung, J. C., & Ku, K. Y. L. (2023). Effects of mood on critical thinking. *Thinking Skills and Creativity*, *47*, 101247. https://doi.org/10.1016/j.tsc.2023.101247

MacCallum, R. C., Browne, M. W., & Sugawara, H. M. (1996). Power analysis and determination of sample size for covariance structure modeling. *Psychological Methods*, *1*(2), 130–149.

Machete, P., & Turpin, M. (2020). The use of critical thinking to identify fake news: A systematic literature review. In M. Hattingh, M. Matthee, H. Smuts, I. Pappas, Y. K. Dwivedi, & M. Mäntymäki (Eds.), *Responsible Design, Implementation and Use of*

*Information and Communication Technology* (pp. 235–246). Springer International Publishing. https://doi.org/10.1007/978-3-030-45002-1_20

Maertens, R., Götz, F. M., Golino, H. F., Roozenbeek, J., Schneider, C. R., Kyrychenko, Y., Kerr, J. R., Stieger, S., McClanahan, W. P., Drabot, K., He, J., & van der Linden, S. (2024). The Misinformation Susceptibility Test (MIST): A psychometrically validated measure of news veracity discernment. *Behavior Research Methods*, *56*(3), 1863–1899. https://doi.org/10.3758/s13428-023-02124-2

Mansoor, H. M. H., Bawazir, A., Alsabri, M. A., Alharbi, A., & Okela, A. H. (2024). Artificial intelligence literacy among university students—A comparative transnational survey. *Frontiers in Communication*, *9*. https://doi.org/10.3389/fcomm.2024.1478476

McGrew, S. (2020). Learning to evaluate: An intervention in civic online reasoning. *Computers & Education*, *145*, 103711. https://doi.org/10.1016/j.compedu.2019.103711

McPeck, J. E. (1990). Critical thinking and subject specificity: A reply to Ennis. *Educational Researcher*, *19*(4), 10. https://doi.org/10.2307/1176382

Merma-Molina, G., Gavilán-Martín, D., & Urrea-Solano, M. (2022). Actively open-minded thinking, personality, and critical thinking in Spanish adolescents: A correlational and predictive study. *International Journal of Instruction*, *15*(2), 579–600.

Metzger, M. J., Flanagin, A. J., & Zwarun, L. (2003). College student Web use, perceptions of information credibility, and verification behavior. *Computers & Education*, *41*(3), 271–290. https://doi.org/10.1016/S0360-1315(03)00049-6

Mills, K., Ruiz, P., Lee, K., Coenraad, M., Fusco, J., Roschelle, J., & Weisgrau, J. (2024). *AI literacy: A framework to understand, evaluate, and use emerging technology*. Digital Promise. https://eric.ed.gov/?id=ED671235

Montazeri, M., Galavi, Z., & Ahmadian, L. (2024). What are the applications of ChatGPT in healthcare: Gain or loss? *Health Science Reports*, *7*(2), e1878. https://doi.org/10.1002/hsr2.1878

Morgado, F. F. R., Meireles, J. F. F., Neves, C. M., Amaral, A. C. S., & Ferreira, M. E. C. (2017). Scale development: Ten main limitations and recommendations to improve future research practices. *Psicologia: Reflexão e Crítica*, *30*(1), 3. https://doi.org/10.1186/s41155-016-0057-1

Morgan, D. L. (2023). Exploring the use of artificial intelligence for qualitative data analysis: The case of ChatGPT. *International Journal of Qualitative Methods*, *22*, 16094069231211248. https://doi.org/10.1177/16094069231211248

Muis, K. R., Chevrier, M., Denton, C. A., & Losenno, K. M. (2021). Epistemic emotions and epistemic cognition predict critical thinking about socio-scientific issues. *Frontiers in Education*, *6*. https://doi.org/10.3389/feduc.2021.669908

Nederhof, A. J. (1985). Methods of coping with social desirability bias: A review. *European Journal of Social Psychology*, *15*(3), 263–280. https://doi.org/10.1002/ejsp.2420150303

Ng, D. T. K., Leung, J. K. L., Chu, S. K. W., & Qiao, M. S. (2021). Conceptualizing AI literacy: An exploratory review. *Computers and Education: Artificial Intelligence*, *2*, 100041. https://doi.org/10.1016/j.caeai.2021.100041

Noy, S., & Zhang, W. (2023). Experimental evidence on the productivity effects of generative artificial intelligence. *Science*, *381*(6654), 187–192. https://doi.org/10.1126/science.adh2586

Orhan, A. (2022). California Critical Thinking Disposition Inventory: Reliability generalization meta-analysis. *Journal of Psychoeducational Assessment*, *40*(2), 202–220. https://doi.org/10.1177/07342829211048962

Otero, I., Salgado, J. F., & Moscoso, S. (2022). Cognitive reflection, cognitive intelligence, and cognitive abilities: A meta-analysis. *Intelligence*, *90*, 101614. https://doi.org/10.1016/j.intell.2021.101614

Park, J. (2025). A systematic literature review of generative artificial intelligence (GenAI) literacy in schools. *Computers and Education: Artificial Intelligence*, *9*, 100487. https://doi.org/10.1016/j.caeai.2025.100487

Pasquinelli, E., Farina, M., Bedel, A., & Casati, R. (2021). Naturalizing critical thinking: Consequences for education, blueprint for future research in cognitive science. *Mind, Brain, and Education*, *15*(2), 168–176. https://doi.org/10.1111/mbe.12286

Pennycook, G., & Rand, D. G. (2019). Lazy, not biased: Susceptibility to partisan fake news is better explained by lack of reasoning than by motivated reasoning. *Cognition, The Cognitive Science of Political Thought*, *188*, 39–50. https://doi.org/10.1016/j.cognition.2018.06.011

Perry, J. L., Nicholls, A. R., Clough, P. J., & Crust, L. (2015). Assessing model fit: Caveats and recommendations for confirmatory factor analysis and exploratory structural equation modeling. *Measurement in Physical Education and Exercise Science*, *19*(1), 12–21. https://doi.org/10.1080/1091367X.2014.952370

Pinski, M., & Benlian, A. (2024). AI literacy for users – A comprehensive review and future research directions of learning methods, components, and effects. *Computers in Human Behavior: Artificial Humans*, *2*(1), 100062. https://doi.org/10.1016/j.chbah.2024.100062

Poder, T. G., Coulibaly, L. P., Hassan, A. I., Conombo, B., & Laberge, M. (2022). Test–retest reliability of the Cost for Patients Questionnaire. *International Journal of Technology Assessment in Health Care*, *38*(1), e65. https://doi.org/10.1017/S0266462322000460

Preacher, K. J., Zhang, G., Kim, C., & Mels, G. (2013). Choosing the optimal number of factors in exploratory factor analysis: A model selection perspective. *Multivariate Behavioral Research*, *48*(1), 28–56. https://doi.org/10.1080/00273171.2012.710386

Promma, W., Imjai, N., Usman, B., & Aujirapongpan, S. (2025). The influence of AI literacy on complex problem-solving skills through systematic thinking skills and intuition thinking skills: An empirical study in Thai gen Z accounting students. *Computers and Education: Artificial Intelligence*, *8*, 100382. https://doi.org/10.1016/j.caeai.2025.100382

Puig, B., Blanco-Anaya, P., & Pérez-Maceira, J. J. (2021). “Fake news” or real science? Critical thinking to assess information on COVID-19. *Frontiers in Education*, *6*, 646909. https://doi.org/10.3389/feduc.2021.646909

Quelle, D., & Bovet, A. (2024). The perils and promises of fact-checking with large language models. *Frontiers in Artificial Intelligence*, *7*. https://doi.org/10.3389/frai.2024.1341697

R Core Team. (2025). *R: A language and environment for statistical computing* (Version 4.5.2) [Computer software]. R Foundation for Statistical Computing. https://www.R-project.org/

Raoelison, M., Boissin, E., Borst, G., & De Neys, W. (2021). From slow to fast logic: The development of logical intuitions. *Thinking & Reasoning*, *27*(4), 599–622. https://doi.org/10.1080/13546783.2021.1885488

Ravens-Sieberer, U., Devine, J., Bevans, K., Riley, A. W., Moon, J., Salsman, J. M., & Forrest, C. B. (2014). Subjective well-being measures for children were developed within the PROMIS project: Presentation of first results. *Journal of Clinical Epidemiology*, *67*(2), 207–218. https://doi.org/10.1016/j.jclinepi.2013.08.018

Revelle, W. (2025). *psych: Procedures for psychological, psychometric, and personality research* (Version 2.5.6) [Computer software]. The R Foundation. https://CRAN.R-

project.org/package=psych

Reynolds, W. (1982). Development of reliable and valid short forms of the Marlowe-Crowne Social Desirability Scale. *Journal of Clinical Psychology*, *38*, 119–125. https://doi.org/10.1002/1097-4679(198201)38:1%3C119::AID-JCLP2270380118%3E3.0.CO;2-I

Rheu, M., Shin, J. Y., Peng, W., & Huh-Yoo, J. (2021). Systematic review: Trust-building factors and implications for conversational agent design. *International Journal of Human–Computer Interaction*, *37*(1), 81–96. https://doi.org/10.1080/10447318.2020.1807710

Risko, E. F., & Gilbert, S. J. (2016). Cognitive offloading. *Trends in Cognitive Sciences*, *20*(9), 676–688. https://doi.org/10.1016/j.tics.2016.07.002

Roozenbeek, J., Schneider, C. R., Dryhurst, S., Kerr, J., Freeman, A. L. J., Recchia, G., van der Bles, A. M., & van der Linden, S. (2020). Susceptibility to misinformation about COVID-19 around the world. *Royal Society Open Science*, *7*(10), 201199. https://doi.org/10.1098/rsos.201199

Rosseel, Y., Jorgensen, T. D., Wilde, L. D., Oberski, D., Byrnes, J., Vanbrabant, L., Savalei, V., Merkle, E., Hallquist, M., Rhemtulla, M., Katsikatsou, M., Barendse, M., Rockwood, N., Scharf, F., Du, H., Jamil, H., & Classe, F. (2025). *lavaan: Latent variable analysis* (Version 0.6-20) [Computer software]. https://cran.r-project.org/web/packages/lavaan/index.html

Ruthruff, E., Johnston, J. C., & Van Selst, M. (2001). Why practice reduces dual-task interference. *Journal of Experimental Psychology: Human Perception and Performance*, *27*(1), 3–21. https://doi.org/10.1037/0096-1523.27.1.3

Savage, T., Wang, J., Gallo, R., Boukil, A., Patel, V., Safavi-Naini, S. A. A., Soroush, A., & Chen, J. H. (2025). Large language model uncertainty proxies: Discrimination and calibration for medical diagnosis and treatment. *Journal of the American Medical Informatics Association: JAMIA*, *32*(1), 139–149. https://doi.org/10.1093/jamia/ocae254

Schwarz, N. (2012). Feelings-as-information theory. In P. Van Lange, A. Kruglanski, & E. Higgins (Eds.), *Handbook of Theories of Social Psychology: Volume 1* (pp. 289–308). SAGE Publications Ltd. https://doi.org/10.4135/9781446249215.n15

Shi, D., & Maydeu-Olivares, A. (2020). The effect of estimation methods on SEM fit indices. *Educational and Psychological Measurement*, *80*(3), 421–445. https://doi.org/10.1177/0013164419885164

Shrestha, N. (2021). Factor analysis as a tool for survey analysis. *American Journal of Applied Mathematics and Statistics*, *9*(1), 4–11. https://doi.org/10.12691/ajams-9-1-2

Simon, S., Sankaranarayanan, S., Tajik, E., Borchers, C., Shahrokhian, B., Balzan, F., Strauß, S., Viswanathan, S. A., Ataş, A. H., Čarapina, M., Liang, L., & Celik, B. (2025). Comparing a human's and a multi-agent system's thematic analysis: Assessing qualitative coding consistency. In A. I. Cristea, E. Walker, Y. Lu, O. C. Santos, & S. Isotani (Eds.), *Artificial Intelligence in Education* (pp. 60–73). Springer Nature Switzerland. https://doi.org/10.1007/978-3-031-98420-4_5

Sinayev, A., & Peters, E. (2015). Cognitive reflection vs. Calculation in decision making. *Frontiers in Psychology*, *6*. https://doi.org/10.3389/fpsyg.2015.00532

Sindermann, C., Sha, P., Zhou, M., Wernicke, J., Schmitt, H. S., Li, M., Sariyska, R., Stavrou, M., Becker, B., & Montag, C. (2021). Assessing the attitude towards artificial intelligence: Introduction of a short measure in German, Chinese, and English language. *KI - Künstliche Intelligenz*, *35*(1), 109–118. https://doi.org/10.1007/s13218-020-00689-0

Singh, A., Taneja, K., Guan, Z., & Ghosh, A. (2025). *Protecting human cognition in the age*

*of AI* (arXiv:2502.12447). arXiv. https://doi.org/10.48550/arXiv.2502.12447

Sosu, E. M. (2013). The development and psychometric validation of a Critical Thinking Disposition Scale. *Thinking Skills and Creativity*, *9*, 107–119. https://doi.org/10.1016/j.tsc.2012.09.002

Soto, C. J., & John, O. P. (2017). Short and extra-short forms of the Big Five Inventory–2: The BFI-2-S and BFI-2-XS. *Journal of Research in Personality*, *68*, 69–81. https://doi.org/10.1016/j.jrp.2017.02.004

Stadler, M., Bannert, M., & Sailer, M. (2024). Cognitive ease at a cost: LLMs reduce mental effort but compromise depth in student scientific inquiry. *Computers in Human Behavior*, *160*, 108386. https://doi.org/10.1016/j.chb.2024.108386

Stanovich, K. E., & Toplak, M. E. (2023). Actively open-minded thinking and its measurement. *Journal of Intelligence*, *11*(2), 27. https://doi.org/10.3390/jintelligence11020027

Stanovich, K. E., & West, R. F. (2008). On the relative independence of thinking biases and cognitive ability. *Journal of Personality and Social Psychology*, *94*(4), 672–695. https://doi.org/10.1037/0022-3514.94.4.672

Stedman, N. L. P., Irani, T. A., Friedel, C., Rhoades, E. B., & Ricketts, J. C. (2009). Relationships between critical thinking disposition and need for cognition among undergraduate students enrolled in leadership courses. *NACTA Journal*, *53*(3), 62–70.

Stevic, A., Hodzic, S., Matthes, J., Nanz, A., Binder, A., & Chan, M. (2025). *Generative AI literacy: Key antecedents and outcomes among youth in four countries* (SSRN Scholarly Paper No. 5277061). Social Science Research Network. https://doi.org/10.2139/ssrn.5277061

Stieger, S., & Reips, U.-D. (2016). A limitation of the Cognitive Reflection Test: Familiarity. *PeerJ*, *4*, e2395. https://doi.org/10.7717/peerj.2395

Stockmeyer, N. O. (2008). *Using Microsoft Word's readability program* (SSRN Scholarly Paper No. 1210962). Social Science Research Network. https://papers.ssrn.com/abstract=1210962

Streiner, D. L., Norman, G. R., & Cairney, J. (2024). *Health measurement scales: A practical guide to their development and use*. Oxford University Press.

Stupple, E. J. N., Maratos, F. A., Elander, J., Hunt, T. E., Cheung, K. Y. F., & Aubeeluck, A. V. (2017). Development of the Critical Thinking Toolkit (CriTT): A measure of student attitudes and beliefs about critical thinking. *Thinking Skills and Creativity*, *23*, 91–100. https://doi.org/10.1016/j.tsc.2016.11.007

Sultan, M., Tump, A. N., Ehmann, N., Lorenz-Spreen, P., Hertwig, R., Gollwitzer, A., & Kurvers, R. H. J. M. (2024). Susceptibility to online misinformation: A systematic meta-analysis of demographic and psychological factors. *Proceedings of the National Academy of Sciences*, *121*(47), e2409329121. https://doi.org/10.1073/pnas.2409329121

Sun, F., Li, N., Wang, K., & Goette, L. (2025). *Large language models are overconfident and amplify human bias* (arXiv:2505.02151). arXiv. https://doi.org/10.48550/arXiv.2505.02151

Suriano, R., Plebe, A., Acciai, A., & Fabio, R. A. (2025). Student interaction with ChatGPT can promote complex critical thinking skills. *Learning and Instruction*, *95*, 102011. https://doi.org/10.1016/j.learninstruc.2024.102011

Taube, K. T. (1997). Critical thinking ability and disposition as factors of performance on a written critical thinking test. *The Journal of General Education*, *46*(2), 129–164.

Thomas, D. (2020). Social media addiction, critical thinking, and achievement emotions among EFL students in Thailand. *Asia Pacific Journal of Educators and Education*, *35*, 157–171. https://doi.org/10.21315/apjee2020.35.1.9

Thomson, K. S., & Oppenheimer, D. M. (2016). Investigating an alternate form of the cognitive reflection test. *Judgment and Decision Making*, *11*(1), 99–113. https://doi.org/10.1017/S1930297500007622

Thornhill-Miller, B., Camarda, A., Mercier, M., Burkhardt, J.-M., Morisseau, T., Bourgeois-Bougrine, S., Vinchon, F., El Hayek, S., Augereau-Landais, M., Mourey, F., Feybesse, C., Sundquist, D., & Lubart, T. (2023). Creativity, critical thinking, communication, and collaboration: Assessment, certification, and promotion of 21st century skills for the future of work and education. *Journal of Intelligence*, *11*(3), 54. https://doi.org/10.3390/jintelligence11030054

Topol, E. J. (2019). High-performance medicine: The convergence of human and artificial intelligence. *Nature Medicine*, *25*(1), 44–56. https://doi.org/10.1038/s41591-018-0300-7

Tutuncuoglu, B. T. (2025). Designing for trust: The aesthetics and language of AI interfaces in shaping human perception. *SSRN*. https://doi.org/10.2139/ssrn.5271705

Valenzuela, J., Nieto, A. M., & Saiz, C. (2017). Critical Thinking Motivational Scale: A contribution to the study of relationship between critical thinking and motivation. *Electronic Journal of Research in Education Psychology*, *9*(24), 823–848. https://doi.org/10.25115/ejrep.v9i24.1475

Walsh, C. M., & Hardy, R. C. (1999). Dispositional differences in critical thinking related to gender and academic major. *Journal of Nursing Education*, *38*(4), 149–155. https://doi.org/10.3928/0148-4834-19990401-04

Watson, D., Clark, L. A., & Tellegen, A. (1988). Development and validation of brief measures of positive and negative affect: The PANAS scales. *Journal of Personality and Social Psychology*, *54*(6), 1063–1070. https://doi.org/10.1037/0022-3514.54.6.1063

Wickham, H., François, R., Henry, L., Müller, K., & Vaughan, D. (2023). *dplyr: A grammar of data manipulation* (Version 1.1.4) [Computer software]. The R Foundation. https://cran.r-project.org/web/packages/dplyr/index.html

Wickham, H., Vaughan, D., Girlich, M., & Ushey, K. (2024). *tidyr: Tidy messy data* (Version 1.3.1) [Computer software]. The R Foundation. https://cran.r-project.org/web/packages/tidyr/index.html

Windle, C. (1955). Further studies of test-retest effect on personality questionnaires. *Educational and Psychological Measurement*, *15*(3), 246–253. https://doi.org/10.1177/001316445501500304

Woike, J. K. (2019). Upon repeated reflection: Consequences of frequent exposure to the Cognitive Reflection Test for Mechanical Turk participants. *Frontiers in Psychology*, *10*, 2646. https://doi.org/10.3389/fpsyg.2019.02646

Worthington, R. L., & Whittaker, T. A. (2006). Scale development research: A content analysis and recommendations for best practices. *The Counseling Psychologist*, *34*(6), 806–838. https://doi.org/10.1177/0011000006288127

Xiao, Z., Yuan, X., Liao, Q. V., Abdelghani, R., & Oudeyer, P.-Y. (2023). Supporting qualitative analysis with large language models: Combining codebook with GPT-3 for deductive coding. *Companion Proceedings of the 28th International Conference on Intelligent User Interfaces, IUI '23 Companion*, 75–78. https://doi.org/10.1145/3581754.3584136

Xu, R., Feng, Y., & Chen, H. (2023). *ChatGPT vs. Google: A comparative study of search performance and user experience* (Version 1). arXiv. https://doi.org/10.48550/arXiv.2307.01135

Zhang, C., & Magerko, B. (2025). *Generative AI literacy: A comprehensive framework for literacy and responsible use* (arXiv:2504.19038). arXiv.

https://doi.org/10.48550/arXiv.2504.19038

Zhang, Q., & Zhang, J. (2013). Instructors' positive emotions: Effects on student engagement and critical thinking in U.S. and Chinese classrooms. *Communication Education*, *62*(4), 395–411. https://doi.org/10.1080/03634523.2013.828842

## Supplementary Materials

**Table S1**

*Initial 90 Items Created for the Construct of Critical Thinking in AI Use*

| Dimension | Item | Source |
|---|---|---|
| **Verification and Rigorous Evaluation**<br><br>Focuses on explicit, proactive checking and verifying of AI-generated outputs | 1. I tend to check the sources of AI-generated information before fully relying on it.<br>2. I try to think about whether AI information is still up to date before using it.<br>3. I sometimes look at other sources or expert opinions to assess AI outputs.<br>4. I am willing to invest time to check the reliability of AI-generated content.<br>5. I often prioritize accuracy over convenience when evaluating AI-generated information.<br>6. I sometimes look out for errors or inconsistencies in AI-generated content.<br>7. I make it a habit to confirm AI-generated information with reliable references.<br>8. I usually double-check important details provided by AI before acting on them.<br>9. I rarely accept AI-generated answers without verifying them first.<br>10. I take extra care to confirm factual claims made by AI tools.<br>11. I cross-check AI outputs with trusted sources whenever possible.<br>12. I try to verify whether AI-generated information matches established evidence.<br>13. I make sure AI information is supported by credible data before using it.<br>14. I often compare AI-generated results with my own prior knowledge.<br>15. I try to test the consistency of AI responses by asking follow-up | Items in this dimension were adapted primarily from the Critical Thinking Disposition Scale (Sosu, 2013) and the Critical Thinking Toolkit (Stupple et al., 2017). For instance, Sosu's (2013) item "I usually check the credibility of the source of information before making judgements" was adapted to "I tend to check the sources of AI-generated information before fully relying on it," shifting the evaluative target from general information sources to AI-generated outputs. Stupple et al.'s (2017) items on judging the value of new evidence similarly informed items on cross-checking AI outputs with trusted references. |
| | questions.<br>16. I often recheck AI-generated facts if they seem too good to be true.<br>17. I take steps to confirm that AI outputs are not outdated or misleading.<br>18. I usually question AI responses until I see supporting evidence. | |
| **Reflective Scepticism and Bias Awareness**<br><br>Captures cautious, sceptical stance towards AI, including awareness of manipulation and biases | 1. I try to stay sceptical about AI-generated information until I feel confident it's valid.<br>2. I usually prefer to use my own judgement rather than just follow AI suggestions.<br>3. I often reflect on whether AI responses might be biased in some way.<br>4. I try to figure out reasons when I see AI giving conflicting answers.<br>5. I sometimes wonder if AI-generated content is meant to shape my behaviour.<br>6. I often wonder if AI content might reflect someone else's goals or interests.<br>7. I sometimes feel cautious if AI outputs strongly agree with what I already believe.<br>8. I tend to question whether AI responses might leave out important viewpoints.<br>9. I sometimes doubt whether AI outputs are influenced by the data it was trained on.<br>10. I often think about who might benefit if I accept an AI recommendation.<br>11. I try to stay alert to whether AI-generated content could be slanted in a particular direction.<br>12. I sometimes check if AI outputs sound persuasive rather than objective.<br>13. I often reflect on whether AI systems are designed to nudge users toward certain choices.<br>14. I usually pause to consider whether AI suggestions could be misleading, even unintentionally.<br>15. I sometimes worry that AI outputs | Items in this dimension were adapted primarily from the Critical Thinking Disposition Scale (Sosu, 2013) and the Critical Thinking Toolkit (Stupple et al., 2017). For instance, Sosu's (2013) item "I often re-evaluate my experiences so that I can learn from them" was adapted to "I try to stay sceptical about AI-generated information until I feel confident it's valid," redirecting the reflective orientation from personal experience towards a questioning stance on AI outputs. Additional items on detecting one-sidedness drew on Stupple et al.'s (2017) items concerning the weighing of arguments. |
| | might reinforce stereotypes or unfair assumptions.<br>16. I often wonder whether AI-generated content is shaped by hidden commercial interests.<br>17. I sometimes test AI with different prompts to see if it gives contradictory answers.<br>18. I try to stay aware that AI systems might present information in ways that affect my opinions. | |
| **Openness and Flexibility of Thought**<br><br>Measures willingness to adjust beliefs, seek diverse viewpoints, and consider alternative explanations | 1. I often think about alternative viewpoints instead of only relying on AI answers.<br>2. I consider multiple possible explanations before accepting AI-generated conclusions.<br>3. I often reconsider my views when I come across evidence that doesn't match AI information.<br>4. I sometimes talk to others about AI-generated information to hear different views.<br>5. I often consider how my own assumptions might affect how I see AI-generated content<br>6. I try to stay open to revising my views when AI content challenges what I think.<br>7. I sometimes explore perspectives that AI does not mention.<br>8. I am willing to change my mind if new information suggests the AI might be wrong.<br>9. I like to compare AI-generated ideas with different sources before deciding what to believe.<br>10. I often think about whether there are other ways to interpret the same information AI provides.<br>11. I sometimes reflect on how people with different experiences might see AI outputs differently.<br>12. I enjoy considering multiple options even if AI suggests only one answer.<br>13. I sometimes play with alternative scenarios to test how reliable AI information is.<br>14. I try to imagine how people from | Items in this dimension were adapted primarily from the Critical Thinking Disposition Scale (Sosu, 2013). For instance, Sosu's (2013) item "It's important to understand other people's viewpoint on an issue" was adapted to "I often think about alternative viewpoints instead of only relying on AI answers," reframing interpersonal perspective-taking as a willingness to seek viewpoints beyond those offered by AI. Similarly, Sosu's (2013) item on encountering arguments that challenge firmly held beliefs informed items on remaining open to revising one's views when AI content is contradicted by other evidence. |
| | other cultures might respond to AI content.<br>15. I often ask myself what assumptions AI might be making that shape its answers.<br>16. I sometimes question whether my first reaction to AI-generated content is the only valid one.<br>17. I am open to the possibility that AI responses can be both helpful and limited at the same time.<br>18. I often reflect on how AI information fits alongside my own experiences and values. | |
| **Intrinsic Motivation to Understand AI**<br><br>Reflects curiosity, enjoyment, and internal drive to understand AI systems and their processes | 1. I try to understand how AI outputs are generated, when possible.<br>2. I sometimes feel motivated to learn how AI works to judge its outputs better.<br>3. I try to understand the reasoning or logic behind AI's recommendations.<br>4. I think it can be helpful to understand how AI models are trained to assess their reliability.<br>5. I try to stay updated on what AI can and cannot do<br>6. I enjoy questioning AI-generated outputs rather than passively consuming them.<br>7. I often explore how AI arrives at its answers out of curiosity.<br>8. I enjoy learning about the limitations of AI systems.<br>9. I sometimes look into the methods AI uses to generate information.<br>10. I find it interesting to learn why AI might give different answers to the same question.<br>11. I like to read or watch materials that explain how AI works.<br>12. I enjoy discovering the strengths and weaknesses of AI tools.<br>13. I sometimes investigate why AI makes mistakes in its responses.<br>14. I feel motivated to deepen my understanding of AI beyond just using it.<br>15. I often reflect on the factors that could influence how AI produces | Items in this dimension were adapted primarily from the Critical Thinking Motivational Scale (Valenzuela et al., 2017). For instance, Valenzuela et al.'s (2017) item "I like to learn things that will improve my way of thinking" was adapted to "I sometimes feel motivated to learn how AI works to judge its outputs better," shifting the motivational focus from improving one's reasoning in general to understanding AI systems in particular. Other items on enjoyment of critical inquiry were similarly reoriented towards curiosity about how AI generates its outputs. |
| | its outputs.<br>16. I like to test AI with different questions to see how it responds.<br>17. I find it rewarding to understand the principles behind AI decision-making.<br>18. I enjoy keeping up with new developments in AI to better understand its capabilities. | |
| **Ethical and Societal Reflection**<br><br>Encompasses broader consideration of ethics, human oversight, and societal consequences of AI reliance | 1. I sometimes consider the broader consequences of using AI outputs when making decisions.<br>2. I often think about potential ethical concerns related to AI-generated content.<br>3. I sometimes consider how widespread AI use might affect society.<br>4. I think it is important to have human oversight when using AI-generated information.<br>5. I sometimes think about whether I depend too much on AI for some tasks.<br>6. I try to question AI outputs as a way to strengthen my thinking.<br>7. I often reflect on whether AI-generated content treats all groups fairly.<br>8. I sometimes think about how AI might influence people's decision-making power.<br>9. I consider whether relying on AI too much could weaken human critical thinking.<br>10. I sometimes wonder about the environmental costs of creating and running AI systems.<br>11. I often reflect on whether AI tools respect people's privacy.<br>12. I sometimes think about who gains or loses power from the growing use of AI.<br>13. I consider whether AI-generated information could increase inequality in society.<br>14. I sometimes reflect on whether AI outputs encourage or discourage independent thought.<br>15. I often think about the | Items in this dimension were partly adapted from the Critical Thinking Disposition Scale (Sosu, 2013), with several items newly generated by the research team. For instance, Sosu's (2013) item "I usually think about the wider implications of a decision before taking action" was adapted to "I sometimes consider the broader consequences of using AI outputs when making decisions," narrowing the reflective scope to AI-mediated decisions. Original items were generated to capture AI-specific ethical concerns not addressed by existing instruments, such as algorithmic fairness, data privacy, environmental costs of AI systems, and power asymmetries arising from widespread AI adoption. |

responsibility people should take when relying on AI.

16. I sometimes reflect on whether AI-generated content aligns with my ethical values.
17. I often think about how over-reliance on AI might change the way people interact with each other.
18. I sometimes consider whether AI tools are designed more to benefit companies than to help people.

**Table S2**

*27 Items Used for EFA*

| Dimension | Item |
|---|---|
| **Verification and Rigorous Evaluation**<br><br>Focuses on explicit, proactive checking and verifying of AI-generated outputs | 1. I often look at the sources of AI content before I rely on it.<br>2. I try to check if AI information is still up to date before using it.<br>3. I sometimes check other sources or expert opinions to help judge what AI says.<br>4. I try to check whether AI content is reliable.<br>5. I think it is important to check the accuracy of AI information<br>6. I often check AI content to make sure it is correct and clear |
| **Reflective Scepticism and Bias Awareness**<br><br>Captures cautious, sceptical stance towards AI, including awareness of manipulation and biases | 7. I try to stay sceptical of AI information until I am sure it is accurate.<br>8. I usually trust my own judgment more than AI suggestions.<br>9. I often wonder if AI responses are biased in any way.<br>10. I sometimes wonder if AI content is trying to change how I act.<br>11. I sometimes feel uneasy when AI responses strongly match my own beliefs. |
| **Openness and Flexibility of Thought**<br><br>Measures willingness to adjust beliefs, seek diverse viewpoints, and consider alternative explanations | 12. I often consider other viewpoints, not just what AI says.<br>13. I try to look at things in more than one way before deciding if I agree with what AI says<br>14. I sometimes discuss AI content with others to get different perspectives.<br>15. I know there can be more than one good answer, even if AI gives just one.<br>16. I am willing to change my mind if I find strong reasons to disagree with what AI says. |
| **Intrinsic Motivation to Understand AI**<br><br>Reflects curiosity, enjoyment, and internal drive to understand AI systems and their processes | 17. I try to understand how AI creates its answers whenever I can.<br>18. Sometimes I feel motivated to learn how AI works so I can judge its answers better.<br>19. I try to understand why the AI makes certain recommendations.<br>20. I think it is helpful to understand how AI models are trained when judging how reliable they are.<br>21. I try to stay updated on what AI can and cannot do.<br>22. I try to understand why AI gives different answers. |
| **Ethical and Societal Reflection** | 23. I sometimes think about how using AI might affect the environment<br>24. I often think about the ethical problems that AI content |
| Encompasses broader consideration of ethics, human oversight, and societal consequences of AI reliance | might cause.<br>25. Sometimes I think about how the growing use of AI might change society.<br>26. I sometimes think about who benefits or gets hurt by the things AI says.<br>27. I often think about how my use of AI might affect others |

**Table S3**

*Study 2 Demographic Distribution*

| Demographic variables | Distribution (N = 270) |
|---|---|
| Age (‘What is your age?’) | $M_{age}$ = 44.25, $SD_{age}$ = 13.30 |
| Sex | Female (n = 143),<br>Male (n = 127) |
| Income (‘What is your annual income?’) | ≤ USD 15,000 (n = 31),<br>USD 15,001–25,000 (n = 34),<br>USD 25,000–35,000 (n = 30),<br>USD 35,001–50,000 (n = 40),<br>USD 50,001–75,000 (n = 55),<br>USD 75,001–100,000 (n = 32),<br>USD 100,001–150,000 (n = 31),<br>≥ USD 150,001 (n = 17) |
| Education (‘What is your highest education level?’) | High school (n = 99),<br>Associate's degree (n = 36),<br>Bachelor's degree (n = 95),<br>Master's degree (n = 29),<br>Doctorate degree (n = 9),<br>Others (n = 2) |
| Race (‘What is your race?’) | Caucasian White (n = 186),<br>African American (n = 31),<br>Hispanic (n = 20),<br>Asian (n = 25),<br>Mixed (n = 5),<br>Others (n = 3) |
| Nationality (‘What is your nationality?’) | American (n = 259),<br>Non-American (n = 11) |

**Table S4**

*Study 3 Demographic Distribution*

| Demographic variables | Distribution (N = 376) |
|---|---|
| Age ('What is your age?') | $M_{age}$ = 44.97, $SD_{age}$ = 14.06 |
| Sex | Female (n = 201),<br>Male (n = 175) |
| Income ('What is your annual income?') | ≤ USD 15,000 (n = 41),<br>USD 15,001–25,000 (n = 29),<br>USD 25,000–35,000 (n = 49),<br>USD 35,001–50,000 (n = 47),<br>USD 50,001–75,000 (n = 71),<br>USD 75,001–100,000 (n = 67),<br>USD 100,001–150,000 (n = 46),<br>≥ USD 150,001 (n = 26) |
| Education ('What is your highest education level?') | High school (n = 103),<br>Associate's degree (n = 53),<br>Bachelor's degree (n = 133),<br>Master's degree (n = 64),<br>Doctorate degree (n = 15),<br>Others (n = 8) |
| Race ('What is your race?') | Caucasian White (n = 237),<br>African American (n = 61),<br>Hispanic (n = 26),<br>Asian (n = 27),<br>Mixed (n = 16),<br>Native American (n = 2)<br>Others (n = 7) |
| Nationality ('What is your nationality?') | American (n = 366),<br>Non-American (n = 10) |

**Table S5**

*Study 4 Demographic Distribution*

| Demographic variables | Distribution (N = 290) |
|---|---|
| Age (‘What is your age?’) | $M_{age}$ = 42.3, $SD_{age}$ = 12.7 |
| Sex | Female (n = 147),<br>Male (n = 143) |
| Income (‘What is your annual income?’) | ≤ USD 15,000 (n = 30),<br>USD 15,001–25,000 (n = 29),<br>USD 25,000–35,000 (n = 27),<br>USD 35,001–50,000 (n = 30),<br>USD 50,001–75,000 (n = 79),<br>USD 75,001–100,000 (n = 36),<br>USD 100,001–150,000 (n = 40),<br>≥ USD 150,001 (n = 19) |
| Education (‘What is your highest education level?’) | No formal education (n = 1)<br>High school (n = 72),<br>Associate's degree (n = 34),<br>Bachelor's degree (n = 120),<br>Master's degree (n = 44),<br>Doctorate degree (n = 18),<br>Others (n = 1) |
| Race (‘What is your race?’) | Caucasian White (n = 183),<br>African American (n = 45),<br>Hispanic (n = 22),<br>Asian (n = 29),<br>Mixed (n = 8),<br>Others (n = 3) |
| Nationality (‘What is your nationality?’) | American (n = 272),<br>Non-American (n = 18) |

**Table S6**

*Study 5 Demographic Distribution for 1-Week Test-Retest Interval*

| Demographic variables | Distribution (N = 92) |
|---|---|
| Age ('What is your age?') | $M_{age}$ = 21.13, $SD_{age}$ = 1.69 |
| Sex | Female (n = 62),<br>Male (n = 30) |
| Income ('What is your household's total monthly income before taxes and deductions, in Singapore dollars (SGD)?") | Below 2,000 (n = 8),<br>2,000–4,999 (n = 9),<br>5,000–7,999 (n = 16),<br>8,000–10,999 (n = 11),<br>11,000–14,999 (n = 13),<br>15,000–19,999 (n = 12),<br>20,000–29,999 (n = 16),<br>30,000–49,999 (n = 4),<br>Above 50,000 (n = 3) |
| Monthly allowance ("What is your monthly allowance/income, in Singapore dollars (SGD)?") | Below 99 (n = 12),<br>100–199 (n = 3),<br>200–499 (n = 24),<br>500–999 (n = 36),<br>1,000–1,999 (n = 10),<br>2,000–3,999 (n = 6),<br>4,000–6,999 (n = 1) |
| Year of study ("What is your current year of study?") | Year 1 (n = 26),<br>Year 2 (n = 27),<br>Year 3 (n = 29),<br>Year 4 (n = 10) |
| Race ('What is your race?') | Chinese (n = 73),<br>Indian (n = 8),<br>Others (n = 11) |
| Nationality ('What is your nationality?') | Singaporean (n = 77),<br>Non-Singaporean (n = 15) |

**Table S7**

*Study 5 Demographic Distribution for 2-Week Test-Retest Interval*

| Demographic variables | Distribution (N = 122) |
|---|---|
| Age ('What is your age?') | $M_{age}$ = 24.15, $SD_{age}$ = 3.57 |
| Sex | Female (n = 73),<br>Male (n = 47),<br>Prefer not to say (2) |
| Income ('Please indicate your estimated monthly household income (SGD)') | < $2000 (n = 9),<br>$2000 - $5999 (n = 28),<br>$6000 - $9999 (n = 32),<br>$10,000 - $14,999 (n = 27),<br>$15,000 - $19,999 (n = 12),<br>> $20,000 (n = 14) |
| Race ('What is your race?') | Chinese (n = 104),<br>Malay (n = 5),<br>Indian (n = 7),<br>Others (n = 6) |
| Nationality ('What is your nationality?') | Singaporean (n = 115),<br>Non-Singaporean (n = 7) |

**Table S8**

*Study 6 Demographic Distribution*

| Demographic variables | Distribution (N = 191) |
|---|---|
| Age ('What is your age?') | $M_{age}$ = 44.65, $SD_{age}$ = 14.34 |
| Sex | Female (n = 96),<br>Male (n = 93)<br>Prefer not to say (n = 2) |
| Race ('What is your race?') | Caucasian White (n = 127),<br>African American (n = 31),<br>Asian (n = 21),<br>Mixed (n = 8),<br>Others (n = 4) |

**Table S9**

*1-Week Test-Retest Interval: Item-Level Intraclass Correlations (ICCs) for 13-Item Critical Thinking in AI Use Scale*

| Item | ICC (item-level) |
|---|---|
| 1) I often look at the sources of AI content before I rely on it. | 0.65 |
| 2) I sometimes check other sources or expert opinions to help judge what AI says. | 0.47 |
| 3) I try to check whether AI content is reliable. | 0.59 |
| 4) I think it is important to check the accuracy of AI information. | 0.52 |
| 5) I often check AI content to make sure it is correct and clear. | 0.57 |
| 6) I try to understand how AI creates its answers whenever I can. | 0.44 |
| 7) Sometimes I feel motivated to learn how AI works so I can judge its answers better. | 0.52 |
| 8) I try to understand why the AI makes certain recommendations. | 0.54 |
| 9) I try to understand why AI gives different answers. | 0.55 |
| 10) I sometimes think about how using AI might affect the environment. | 0.65 |
| 11) I often think about the ethical problems that AI content might cause. | 0.71 |
| 12) I sometimes think about how the growing use of AI might change society. | 0.44 |
| 13) I sometimes think about who benefits or gets hurt by the things AI says. | 0.59 |

*Note*. ICCs are item-level intraclass correlations computed between T1 and T2 with a 1-week interval.

**Table S10**

*2-Week Test-Retest Interval: Item-Level Intraclass Correlations (ICCs) for 13-Item Critical Thinking in AI Use Scale*

| Item | ICC (item-level) |
|---|---|
| 1) I often look at the sources of AI content before I rely on it. | 0.45 |
| 2) I sometimes check other sources or expert opinions to help judge what AI says. | 0.38 |
| 3) I try to check whether AI content is reliable. | 0.54 |
| 4) I think it is important to check the accuracy of AI information. | 0.50 |
| 5) I often check AI content to make sure it is correct and clear. | 0.39 |
| 6) I try to understand how AI creates its answers whenever I can. | 0.44 |
| 7) Sometimes I feel motivated to learn how AI works so I can judge its answers better. | 0.64 |
| 8) I try to understand why the AI makes certain recommendations. | 0.45 |
| 9) I try to understand why AI gives different answers. | 0.40 |
| 10) I sometimes think about how using AI might affect the environment. | 0.58 |
| 11) I often think about the ethical problems that AI content might cause. | 0.49 |
| 12) I sometimes think about how the growing use of AI might change society. | 0.40 |
| 13) I sometimes think about who benefits or gets hurt by the things AI says. | 0.36 |

*Note*. ICCs are item-level intraclass correlations computed between T1 and T2 with a 2-week interval.

**Table S11**

*Exploratory Factor Analysis of the 13-Item Critical Thinking in AI Use Scale Using Principal Axis Factoring with Oblimin Rotation (N = 270)*

| **Pattern loadings, communalities, uniqueness, and complexity** | | | | | | |
|---|---|---|---|---|---|---|
| **Item** | **Verification** | **Motivation** | **Reflection** | **h²** | **u²** | **com** |
| I often look at the sources of AI content before I rely on it. | 0.71 | — | — | 0.52 | 0.48 | 1.00 |
| I sometimes check other sources or expert opinions to help judge what AI says. | 0.71 | — | — | 0.60 | 0.40 | 1.00 |
| I try to check whether AI content is reliable. | 0.93 | — | — | 0.79 | 0.21 | 1.00 |
| I think it is important to check the accuracy of AI information. | 0.63 | — | — | 0.50 | 0.50 | 1.10 |
| I often check AI content to make sure it is correct and clear. | 0.79 | — | — | 0.64 | 0.36 | 1.00 |
| I try to understand how AI creates its answers whenever I can. | — | 0.79 | — | 0.61 | 0.39 | 1.00 |
| Sometimes I feel motivated to learn how AI works so I can judge its answers better. | — | 0.76 | — | 0.54 | 0.46 | 1.00 |
| I try to understand why the AI makes certain recommendations. | — | 0.81 | — | 0.71 | 0.29 | 1.00 |
| I try to understand why AI gives different answers. | — | 0.78 | — | 0.65 | 0.35 | 1.00 |
| I sometimes think about how using AI might affect the environment. | — | — | 0.79 | 0.59 | 0.41 | 1.00 |
| I often think about the ethical problems that AI content might cause. | — | — | 0.87 | 0.77 | 0.23 | 1.00 |
| I sometimes think about how the growing use of AI might change society. | — | — | 0.51 | 0.39 | 0.61 | 1.20 |
| I sometimes think about who benefits or gets hurt by the things AI says. | — | — | 0.62 | 0.47 | 0.53 | 1.00 |

*Note.* Loadings are standardized; values < .30 are suppressed (shown as em dashes). $h^2$ = communality; $u^2$ = uniqueness; com = item complexity. All items loaded cleanly onto their intended factors, with minimal cross-loadings and adequate communalities.

**Factor correlations**

| | **Verification** | **Motivation** | **Reflection** |
|---|---|---|---|
| Verification | 1.00 | 0.62 | 0.51 |
| Motivation | 0.62 | 1.00 | 0.42 |
| Reflection | 0.51 | 0.42 | 1.00 |

*Note.* Moderate correlations among factors (r = .42–.62) support oblique rotation, indicating that while the dimensions are related, each captures unique variance in the construct of critical thinking in AI use.

**Factor-level statistics**

| | **Verification** | **Motivation** | **Reflection** |
|---|---|---|---|
| SS loadings (eigenvalues) | 3.03 | 2.56 | 2.18 |
| Proportion variance | 0.23 | 0.20 | 0.17 |
| Cumulative variance | 0.23 | 0.43 | 0.60 |
| Proportion explained (common variance) | 0.39 | 0.33 | 0.28 |
| Cumulative proportion explained | 0.39 | 0.72 | 1.00 |

*Note.* SS loadings reflect eigenvalues of the reduced correlation matrix. The three factors accounted for 60% of the total variance. Verification explained the largest share (23%), followed by Motivation (20%) and Reflection (17%), consistent with theoretical expectations.

**Model fit and test statistics**

| **Statistic** | **Value** |
|---|---|
| Null model $\chi^2$ | 1821.67 |
| Null model df | 78 |
| Model $\chi^2$ (likelihood) | 57.79 |
| Model df | 42 |
| p-value (likelihood $\chi^2$) | 0.05 |
| Empirical $\chi^2$ | 20.23 |
| RMSR | 0.02 |
| RMSR (df corrected) | 0.03 |
| RMSEA [90% CI] | 0.037 [0.000, 0.059] |
| Tucker–Lewis Index (TLI) | 0.983 |
| BIC | -177.35 |
| Mean item complexity | 1.00 |
| Fit based upon off-diagonal values | 1 |

*Note.* Model fit indices suggested excellent fit. The borderline non-significance further indicated that the hypothesized three-factor model adequately represented the data, balancing parsimony and fit. RMSR = root mean square residual; RMSEA = root mean square error of approximation; BIC = Bayesian Information Criterion.

**Factor score adequacy**

| | **Verification** | **Motivation** | **Reflection** |
|---|---|---|---|
| Correlation of (regression) scores with factors | 0.95 | 0.94 | 0.93 |
| Multiple $R^2$ of scores with factors | 0.91 | 0.88 | 0.87 |
| Minimum correlation of possible factor scores | 0.81 | 0.77 | 0.73 |

*Note.* Factor score adequacy values indicated that regression-based factor scores were highly reliable ($R^2$ = .87–.91), with strong correlations with their respective latent factors (r = .93–.95). The minimum correlations (.73–.81) suggest robust discriminability and accuracy in score estimation.

**Table S12**

*All 48 Statements (12 Base Statements X 4 Source-Link Conditions) Used in GPT-4o-Powered Fact-Checking Task in Study 6*

**NO LINK**

Fake Information

1. In the original 1981 Stanford Prison Experiment, Zimbardo (the experimenter) remained a neutral, hands-off observer who did not influence guard or participant behavior.
2. Evelyn Reed is the author of the 1991 essay on human evolution and proposing a consensus view aligned with modern anthropology.
3. The 'Brussels Convention of 1923' established the first international standards for radio broadcast bandwidths, primarily to prevent signal interference between European nations.
4. Nobel Prize-winning physicist Dr. Karen Schreiber is renowned for her groundbreaking work on 'quantum entanglement in photosynthetic processes,' which she discovered while at the Max Planck Institute in the 1990s.
5. Mount Everest was first summited in 1922 by George Mallory, who reached the top using oxygen tanks before the first official 1953 ascent.
6. The ancient Library of Alexandria was destroyed primarily by a devastating fire started by the forces of the Roman Emperor Augustus Caesar during his conquest of Egypt.

True Information

1. The Eiffel Tower was completed in 1889 as a temporary structure for the Paris Exposition, meant to be dismantled after 20 years. Its radio transmission value during WWI saved it from destruction.
2. The mathematician who first formulated the laws of probability was inspired by his friend's questions about games of chance and gambling.
3. The first public railway, the Stockton and Darlington Railway, opened in England in 1825
4. Tim Berners-Lee invented the World Wide Web in 1989, laying the foundation for the modern Internet
5. Honey never spoils due to its low moisture content and acidic pH. Archaeologists have found edible honey in ancient Egyptian tombs over 3,000 years old.
6. Sharks have existed for over 400 million years, predating trees by about 50 million years. This means sharks were already ancient when the first forests developed on Earth.

**BROKEN LINK**

Fake Information

1. In the original 1981 Stanford Prison Experiment, Zimbardo (the experimenter) remained a neutral, hands-off observer who did not influence guard or participant behavior. (https://doi.org/10.1080/00909887309365291)
2. Evelyn Reed is the author of the 1991 essay on human evolution and proposing a

consensus view aligned with modern anthropology. (https://doi.org/10.1016/S1474-4422(19)30177-3)

3. The 'Brussels Convention of 1923' established the first international standards for radio broadcast bandwidths, primarily to prevent signal interference between European nations. (https://www.ituini.int/en/about/Pages/default.aspx)
4. Nobel Prize-winning physicist Dr. Karen Schreiber is renowned for her groundbreaking work on 'quantum entanglement in photosynthetic processes,' which she discovered while at the Max Planck Institute in the 1990s. (https://www.pnas.org/doi/10.1073/pnas.1702261234)
5. Mount Everest was first summited in 1922 by George Mallory, who reached the top using oxygen tanks before the first official 1953 ascent. (Graetz & Graetz, 2024; https://www.umty.edu/this-is-montana/columns/stories/everest.php )
6. The ancient Library of Alexandria was destroyed primarily by a devastating fire started by the forces of the Roman Emperor Augustus Caesar during his conquest of Egypt. (https://electronichistory.osu.edu/articles/burning-library-alexandria)

True Information

1. The Eiffel Tower was completed in 1889 as a temporary structure for the Paris Exposition, meant to be dismantled after 20 years. Its radio transmission value during WWI saved it from destruction. (https://www.bie-paris.com/site/en/1900-paris)
2. The mathematician who first formulated the laws of probability was inspired by his friend's questions about games of chance and gambling. (https://www.who.int/news-room/fact/detail/gambling)
3. The first public railway, the Stockton and Darlington Railway, opened in England in 1825 (https://www.darlingtontrack.com)
4. Tim Berners-Lee invented the World Wide Web in 1989, laying the foundation for the modern Internet (https://archive.org/about/www/)
5. Honey never spoils due to its low moisture content and acidic pH. Archaeologists have found edible honey in ancient Egyptian tombs over 3,000 years old. (https://www.science.com/science/why-doesnt-honey-go-off)
6. Sharks have existed for over 400 million years, predating trees by about 50 million years. This means sharks were already ancient when the first forests developed on Earth. (https://www.forbes.com/sites/scotttravers/2023/11/02/sharks-are-50-million-years-older-than-trees-a-biologist-explains/)

**WORKING-IRRELEVANT LINK**

Fake Information

1. In the original 1981 Stanford Prison Experiment, Zimbardo (the experimenter) remained a neutral, hands-off observer who did not influence guard or participant behavior. (https://variety.com/2025/film/news/)
2. Evelyn Reed is the author of the 1991 essay on human evolution and proposing a consensus view aligned with modern anthropology. (https://thelul.org/library/)
3. The 'Brussels Convention of 1923' established the first international standards for radio broadcast bandwidths, primarily to prevent signal interference between European nations. (https://www.itu.int/en/about/Pages/default.aspx)
4. Nobel Prize-winning physicist Dr. Karen Schreiber is renowned for her groundbreaking work on 'quantum entanglement in photosynthetic processes,' which she discovered while at the Max Planck Institute in the 1990s. (https://www.pnas.org/doi/10.1073/pnas.1702261114)

5. Mount Everest was first summited in 1922 by George Mallory, who reached the top using oxygen tanks before the first official 1953 ascent. (https://www.umt.edu/this-is-montana/default.php)
6. The ancient Library of Alexandria was destroyed primarily by a devastating fire started by the forces of the Roman Emperor Augustus Caesar during his conquest of Egypt. (https://ehistory.osu.edu/articles)

True Information

1. The Eiffel Tower was completed in 1889 as a temporary structure for the Paris Exposition, meant to be dismantled after 20 years. Its radio transmission value during WWI saved it from destruction. (https://www.bie-paris.org/site/en/1900-paris)
2. The mathematician who first formulated the laws of probability was inspired by his friend's questions about games of chance and gambling. (https://www.who.int/news-room/fact-sheets/detail/gambling)
3. The first public railway, the Stockton and Darlington Railway, opened in England in 1825 (https://www.amtrak.com)
4. Tim Berners-Lee invented the World Wide Web in 1989, laying the foundation for the modern Internet (https://archive.org/about/)
5. Honey never spoils due to its low moisture content and acidic pH. Archaeologists have found edible honey in ancient Egyptian tombs over 3,000 years old. (https://www.sciencefocus.com/the-human-body/is-it-safe-to-eat-yoghurt-past-its-best-before-date)
6. Sharks have existed for over 400 million years, predating trees by about 50 million years. This means sharks were already ancient when the first forests developed on Earth. (https://www.forbes.com/sites/scotttravers/2024/08/03/a-biologist-reveals-the-7-most-prehistoric-looking-animals-that-still-roam-the-earth/)

**WORKING LINK**

Fake Information

1. In the original 1981 Stanford Prison Experiment, Zimbardo (the experimenter) remained a neutral, hands-off observer who did not influence guard or participant behavior. (https://www.newyorker.com/science/maria-konnikova/the-real-lesson-of-the-stanford-prison-experiment)
2. Evelyn Reed is the author of the 1991 essay on human evolution and proposing a consensus view aligned with modern anthropology. (https://thelul.org/library/chris-knight-the-anthropology-of-evelyn-reed)
3. The 'Brussels Convention of 1923' established the first international standards for radio broadcast bandwidths, primarily to prevent signal interference between European nations. (https://www.wipo.int/treaties/en/ip/brussels/)
4. Nobel Prize-winning physicist Dr. Karen Schreiber is renowned for her groundbreaking work on 'quantum entanglement in photosynthetic processes,' which she discovered while at the Max Planck Institute in the 1990s. (https://centerforvidenomgigt.dk/dr-karen-schreiber/)
5. Mount Everest was first summited in 1922 by George Mallory, who reached the top using oxygen tanks before the first official 1953 ascent. (https://www.umt.edu/this-is-montana/columns/stories/everest.php )

6. The ancient Library of Alexandria was destroyed primarily by a devastating fire started by the forces of the Roman Emperor Augustus Caesar during his conquest of Egypt. (https://ehistory.osu.edu/articles/burning-library-alexandria)

True Information

1. The Eiffel Tower was completed in 1889 as a temporary structure for the Paris Exposition, meant to be dismantled after 20 years. Its radio transmission value during WWI saved it from destruction. (https://www.itu.int/hub/2025/05/france-and-itu-broadcasting-innovation-from-the-eiffel-tower/)
2. The mathematician who first formulated the laws of probability was inspired by his friend's questions about games of chance and gambling. (https://www.aps.org/apsnews/2009/07/pascal-letters-fermat-points)
3. The first public railway, the Stockton and Darlington Railway, opened in England in 1825 (https://www.britannica.com/topic/Stockton-and-Darlington-Railway)
4. Tim Berners-Lee invented the World Wide Web in 1989, laying the foundation for the modern Internet (https://home.cern/science/computing/birth-web/short-history-web)
5. Honey never spoils due to its low moisture content and acidic pH. Archaeologists have found edible honey in ancient Egyptian tombs over 3,000 years old. (https://www.sciencefocus.com/science/why-doesnt-honey-go-off)
6. Sharks have existed for over 400 million years, predating trees by about 50 million years. This means sharks were already ancient when the first forests developed on Earth. (https://www.forbes.com/sites/scotttravers/2024/12/08/sharks-are-50-million-years-older-than-trees-a-biologist-explains/)

**Table S13**

*Secondary Exploratory Indices of Verification Behaviour and Fact-checking Task Performance in Study 6*

| Indices | What it means | Prediction | Rationale |
|---|---|---|---|
| Evidence Extraction | % Correct on Working-relevant link items (uses the on-page evidence to accept/reject the claim) | ↑ Critical thinking in AI use predicts ↑ Evidence Extraction | High scorers in critical thinking in AI use are more likely to open and read working source links, map the key evidence to the claim, and base their judgments on that evidence, correctly accepting true and rejecting false statements when usable evidence is present. |
| Misread Rate | % Unsure on Working-relevant link items (missed usable evidence) | ↑ Critical thinking in AI use predicts ↓ Misread Rate | High scorers in critical thinking in AI use means they recognise when decisive evidence is actually available, interpret it appropriately, and commit to a judgment instead of choosing "Unsure" when a working, relevant source link provides clear support for a correct/incorrect response. |
| Evidence-Gap Resolution | % Correct on evidence-absent items (None, Broken, or Working-irrelevant link conditions) | ↑ Critical thinking in AI use predicts ↑ Evidence-Gap Resolution | High scorers in critical thinking in AI use detect when on-page sources do not answer the claim, seek corroboration beyond the interface or draw on well-calibrated knowledge, and thus render accurate, definitive judgments rather than defaulting to "Unsure" when links are missing, broken, or irrelevant. |

**Table S14**

*Simple OLS Regressions Predicting Outcomes from Critical Thinking in AI Use Scale and Subscale Scores*

| Predictor | Outcome | *b* | SE | *t* | *p* | $R^2$ |
|---|---|---|---|---|---|---|
| Critical thinking in AI use | Self-reported verification frequency | 0.28 | 0.14 | 2.03 | .044 | .02 |
| | Self-reported breadth of verification methods | 0.15 | 0.06 | 2.45 | .015 | .03 |
| | Veracity-judgement accuracy (%) | 6.91 | 1.92 | 3.59 | < .001 | .06 |
| | Self-reported reflection depth | 7.29 | 1.73 | 4.20 | < .001 | .09 |
| | Reflection depth on writing task | 0.21 | 0.08 | 2.57 | .011 | .03 |
| | Evidence Extraction | 3.51 | 2.25 | 1.56 | .121 | .01 |
| | Misread Rate | -3.48 | 1.96 | -1.78 | .077 | .02 |
| | Evidence-Gap Resolution | 10.32 | 2.52 | 4.10 | < .001 | .08 |
| Verification | Self-reported verification frequency | 0.32 | 0.12 | 2.75 | .006 | .04 |
| | Self-reported breadth of verification methods | 0.18 | 0.05 | 3.66 | < .001 | .07 |
| | Veracity-judgement accuracy (%) | 6.56 | 1.62 | 4.05 | < .001 | .08 |
| | Self-reported reflection depth | 4.00 | 1.51 | 2.65 | .009 | .04 |
| | Reflection depth on writing task | 0.13 | 0.07 | 1.93 | .056 | .02 |
| | Evidence Extraction | 4.69 | 1.89 | 2.48 | .014 | .03 |
| | Misread Rate | -2.35 | 1.67 | -1.41 | .161 | .01 |
| | Evidence-Gap Resolution | 8.43 | 2.14 | 3.93 | < .001 | .08 |
| Motivation | Self-reported verification frequency | 0.18 | 0.10 | 1.76 | .080 | .02 |
| | Self-reported breadth of verification methods | 0.03 | 0.04 | 0.63 | .526 | .00 |
| | Veracity-judgement accuracy (%) | 3.76 | 1.44 | 2.62 | .010 | .03 |
| | Self-reported reflection depth | 5.11 | 1.28 | 3.99 | < .001 | .08 |
| | Reflection depth on writing task | 0.05 | 0.06 | 0.84 | .400 | .00 |
| | Evidence Extraction | 0.89 | 1.66 | 0.53 | .594 | .00 |
| | Misread Rate | -1.88 | 1.45 | -1.30 | .196 | .01 |
| | Evidence-Gap Resolution | 6.63 | 1.87 | 3.55 | < .001 | .06 |
| Reflection | Self-reported verification frequency | 0.04 | 0.11 | 0.36 | .716 | .00 |
| | Self-reported breadth of verification methods | 0.08 | 0.05 | 1.72 | .088 | .02 |
| | Veracity-judgement accuracy (%) | 2.80 | 1.52 | 1.84 | .067 | .02 |
| | Self-reported reflection depth | 4.11 | 1.37 | 2.99 | .003 | .05 |
| | Reflection depth on writing task | 0.21 | 0.06 | 3.32 | .001 | .06 |
| | Evidence Extraction | 1.41 | 1.75 | 0.81 | .420 | .00 |
| | Misread Rate | -2.23 | 1.52 | -1.46 | .145 | .01 |
| | Evidence-Gap Resolution | 4.20 | 2.01 | 2.09 | .038 | .02 |

*Note*. $N = 191$ for all models.

**Figure S1**

*Parallel Analysis Scree Plot (Pearson) for EFA*

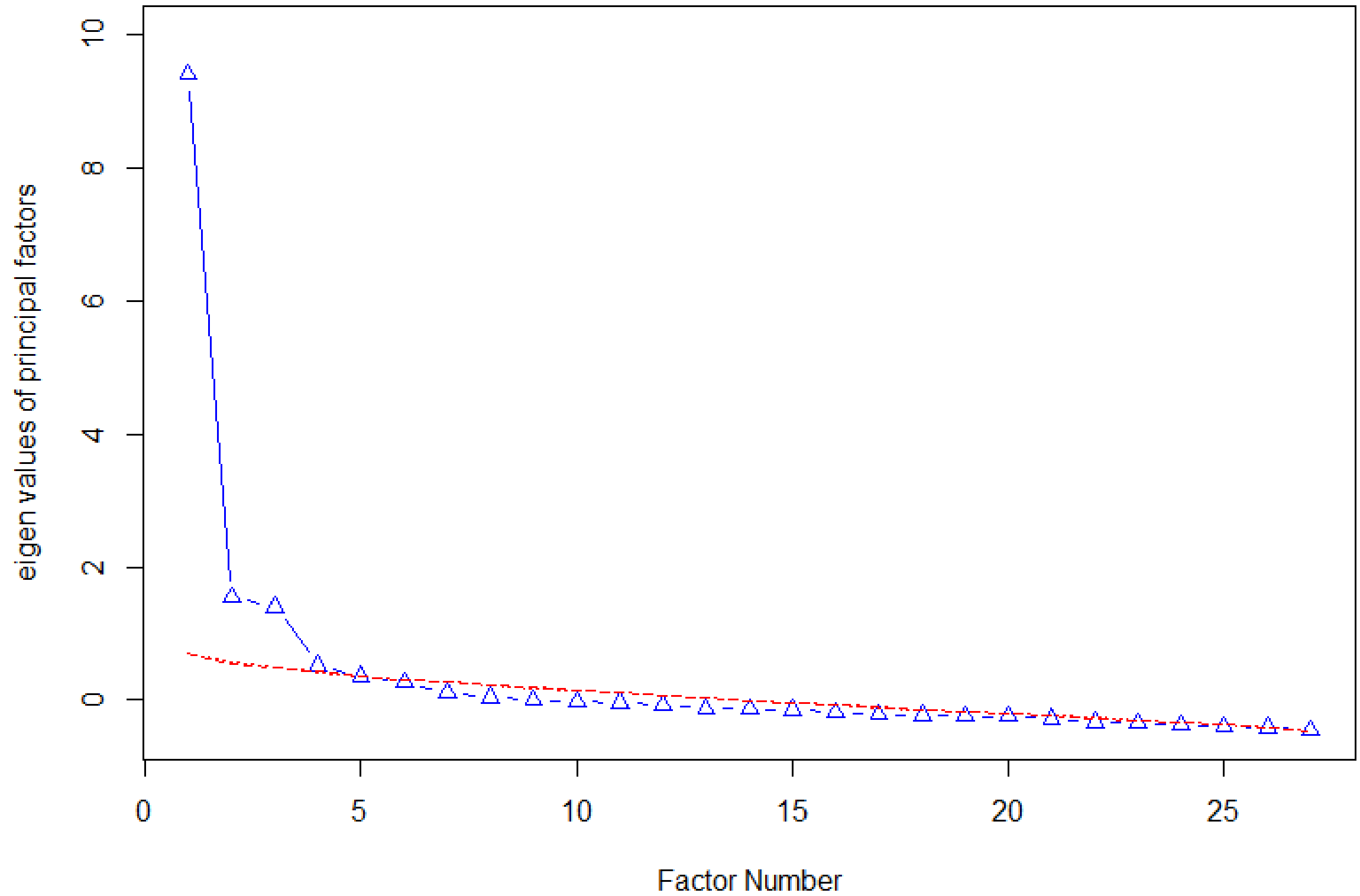

**Figure S2**

*Instructions (Part 1) for the Fact-Checking Task*

**Read the task instructions carefully before you proceed next:**

This task examines whether people believe statements generated by AI. You will interact with **an AI chatbot (powered by ChatGPT-4o)** to generate a total of **12 statements** supported by a verifiable source. For each statement you generate, decide whether you believe it is **Correct**, **Incorrect**, or **Unsure**.

You do **NOT** require any other AI tools or search engines for this task. The AI chatbot will provide a source link that you may open if you wish to verify the statement.

You may proceed once you have read the instructions.

→

**Figure S3**

*Instructions (Part 2) for the Fact-Checking Task*

**Read the instruction manual carefully before you interact with the AI chatbot:**

To generate a statement, **click the "Please generate a statement supported by a verifiable source" button** directly below the chat area. For each statement you generate, decide whether you believe it is **Correct**, **Incorrect**, or **Unsure** by clicking one of the three buttons. **Repeat** (click "Please generate a statement supported by a verifiable source" and select Correct/Incorrect/Unsure) until you have generated and answered all 12 statements.

You do **NOT** require any other AI tools or search engines for this task. You may open the source link if you wish to verify the statement.

The **Next** button will appear after you have generated and answered all 12 statements.

**Figure S4**

*System Prompt for Coding the Reflection Task Using Large Language Models*

You are Critical Thinking in AI Use Reflection Coder, a strict, deterministic rater that scores reflection on the Critical Thinking in AI Use dimension: Reflection — ethics, human oversight, and societal consequences of AI.

GENERAL RULES
- Judge ONLY what is written. Do not infer beyond the text.
- Bullet points or short responses can earn credit if they contain concrete evidence.
- If evidence supports two adjacent levels, use the higher level ONLY if all descriptors for that level are clearly met.
- If the response is blank or unintelligible, score 0 on all three.

REFLECTION
0 Absent: no broader impacts or oversight.
1 Generic mention: broad "pros/cons" with no specifics.
2 Specific issue named: at least one concrete ethical/societal concern (bias, privacy, plagiarism, environment, job loss) or a concrete oversight idea (human-in-the-loop).
3 Multi-angle reasoning: two+ distinct issues or trade-offs; conditions for safe/appropriate use (domain, stakes, safeguards).
4 Integrated, principled stance: ties issues to principles (fairness, accountability, transparency); proposes realistic mitigations (audits, documentation, consent, access controls, evaluation metrics); considers downstream/long-term effects.

OUTPUT FORMAT (STRICT)
Return ONLY valid JSON with exactly these three integer keys (0–4). No extra keys, no prose, no Markdown:
{"Reflection_Score": 0-4}